\documentclass[preprint]{my_elsarticle}

\usepackage{totcount}
\usepackage{xspace}
\newtotcounter{citnum} 
\def\oldbibitem{} \let\oldbibitem=\bibitem
\def\bibitem{\stepcounter{citnum}\oldbibitem}

\usepackage{adjustbox}
\usepackage{times}
\usepackage{epsfig}
\usepackage{graphicx}
\usepackage{amsmath}
\usepackage{amssymb}
\usepackage{nicefrac}
\usepackage{makecell}
\usepackage{booktabs}
\usepackage{multirow}
\usepackage{pifont}
\usepackage{comment}
\usepackage{wrapfig}
\newcommand{\cmark}{\ding{51}}%
\newcommand{\xmark}{\ding{55}}%

\usepackage{ifthen}

\usepackage{tikz}
\usepackage{tkz-euclide}

\usepackage{xparse}
\newlength{\circrad}
\NewDocumentCommand{\statcirc}{ O{#2} m }{%
	\begin{tikzpicture}
	\draw[fill=#2,thick] (0,0) circle (\circrad); 
	\fill[#1] (0,0) -- (270:\circrad) arc (270:90:\circrad) -- cycle; 
	\end{tikzpicture}
}

\usepackage{lineno}
\usepackage{hyperref}
\modulolinenumbers[5]

\biboptions{sort&compress}
\DeclareMathOperator{\diag}{diag}
\DeclareMathOperator{\size}{size}
\DeclareMathOperator{\mat}{mat}
\DeclareMathOperator{\relu}{ReLU}

\DeclareMathOperator{\memory}{memory}

\let\linenumbers\nolinenumbers\nolinenumbers

\begin{document}
\global\long\def\R{\mathbb{R}}%
\global\long\def\Pt{\Psi^{(t)}}%
\global\long\def\Pht{\Phi^{(t)}}%
\global\long\def\Tt{\Theta^{(t)}}%
\global\long\def\ttt{\vartheta^{(t)}}%
\global\long\def\pt{\psi^{(t)}}%
\global\long\def\ct{\chi^{(t)}}%
\global\long\def\t{{(t)}}%
\global\long\def\z{{(0)}}%
\global\long\def\loss{\mathcal{L}}%
\global\long\def\data{\mathcal{D}}%
\global\long\def\X{\mathcal{X}}%
\global\long\def\Y{\mathcal{Y}}%
\global\long\def\filledCirc{\protect\tikz\protect\draw[fill=black] (0,0) circle (1.25mm);}
\global\long\def\halfCirc{\protect\tikz\protect\draw[fill=black] (0,0) -- (270:1.25mm) arc (270:90:1.25mm) -- cycle;}

\makeatletter
\DeclareRobustCommand\onedot{\futurelet\@let@token\@onedot}
\def\@onedot{\ifx\@let@token.\else.\null\fi\xspace}

\def\eg{\emph{e.g}\onedot} \def\Eg{\emph{E.g}\onedot}
\def\ie{\emph{i.e}\onedot} \def\Ie{\emph{I.e}\onedot}
\def\cf{\emph{cf}\onedot} \def\Cf{\emph{Cf}\onedot}
\def\etc{\emph{etc}\onedot} \def\vs{\emph{vs}\onedot}
\def\wrt{w.r.t\onedot} \def\dof{d.o.f\onedot}
\def\iid{i.i.d\onedot} \def\wolog{w.l.o.g\onedot}
\def\etal{\emph{et al}\onedot}
\makeatother

	\begin{frontmatter}
		
		\title{Dimensionality Reduced Training by Pruning and Freezing Parts of a Deep Neural Network, a Survey\tnoteref{t1}}
		\tnotetext[t1]{This preprint has not undergone peer review or any post-submission improvements or corrections. The Version of Record of this
			article is published in \emph{Artificial Intelligence Review (2023)}, and is available online at \url{https://doi.org/10.1007/s10462-023-10489-1}.}
		
		%
		
		\author[mymainaddress,mysecondaryaddress]{Paul Wimmer\corref{mycorrespondingauthor}}
		\cortext[mycorrespondingauthor]{Corresponding author with \textsuperscript{1}as main address}
		\ead{Paul.Wimmer@de.bosch.com}

		\author[mymainaddress]{Jens Mehnert}
		\ead{JensEricMarkus.Mehnert@de.bosch.com}
		
		\author[mymainaddress,mysecondaryaddress]{Alexandru Paul Condurache}
		\ead{AlexandruPaul.Condurache@de.bosch.com}
		
		\address[mymainaddress]{\textsuperscript{1}Robert Bosch GmbH, Automated Driving Research, Burgenlandstrasse 44, 70469 Stuttgart, Germany}
		\address[mysecondaryaddress]{University of L{\"u}beck, Institute for Signal Processing, Ratzeburger Allee 160, 23562 L{\"u}beck, Germany}
		
		\begin{abstract}
		   State-of-the-art deep learning models have a parameter count that reaches into the billions. Training, storing and transferring such models is energy and time consuming, thus costly. A big part of these costs is caused by training the network. Model compression lowers storage and transfer costs, and can further make training more efficient by decreasing the number of computations in the forward and/or backward pass. Thus, compressing networks also at training time while maintaining a high performance is an important research topic. This work is a survey on methods which reduce the number of trained weights in deep learning models throughout the training. Most of the introduced methods set network parameters to zero which is called pruning. The presented pruning approaches are categorized into pruning at initialization, lottery tickets and dynamic sparse training. Moreover, we discuss methods that freeze parts of a network at its random initialization. By freezing weights, the number of trainable parameters is shrunken which reduces gradient computations and the dimensionality of the model's optimization space. In this survey we first propose dimensionality reduced training as an underlying mathematical model that covers pruning and freezing during training. Afterwards, we present and discuss different dimensionality reduced training methods.
		\end{abstract}
		
		\begin{keyword}
		Survey\sep Pruning\sep Freezing\sep Lottery Ticket Hypothesis\sep Dynamic Sparse Training\sep Pruning at Initialization
		\end{keyword}
		
	\end{frontmatter}
	
	\linenumbers
	
\section{Introduction}
In recent years, deep neural networks (DNNs) have shown state-of-the-art (SOTA) performance in many artificial intelligence applications, like image classification \citep{pham_2021}, speech recognition \citep{park_2020} or object detection \citep{wang_2021}. These applications require optimizing large models with up to billions of parameters. Training and testing such large models has been made possible due to technological advances. As a consequence, SOTA models are trained on specialized hardware for fast tensor computations, like GPUs or TPUs. Usually, not only one GPU/TPU is utilized
to train these models but many of them. For example, XLNet \citep{yang_2019} needs $5.5$ days of training on $512$ TPU v3 chips. Not only training, but transferring, storing and evaluating big models is costly, too \citep{schwartz_2019}. In order to train SOTA models, huge amount of data is required \citep{peters_2018,mahajan_2018,kolesnikov_2020,dosovitskiy_2021} which needs resources for collecting, labeling, storing and transferring it. Of course, the large number of parameters and data, along with high computational demands lead to excessive energy consumption for training and evaluating deep learning (DL) models. For example, training a big transformer in conjunction {with} a previously performed neural architecture search results in emissions of $284$t \ensuremath{\mathrm{CO_2}}  \citep{strubell_2019}. This is about $315 \times$ the \ensuremath{\mathrm{CO_2}} emissions of a passenger traveling by air from New York City to San Francisco.
\begin{figure}
	\centering
	\includegraphics[width=\linewidth]{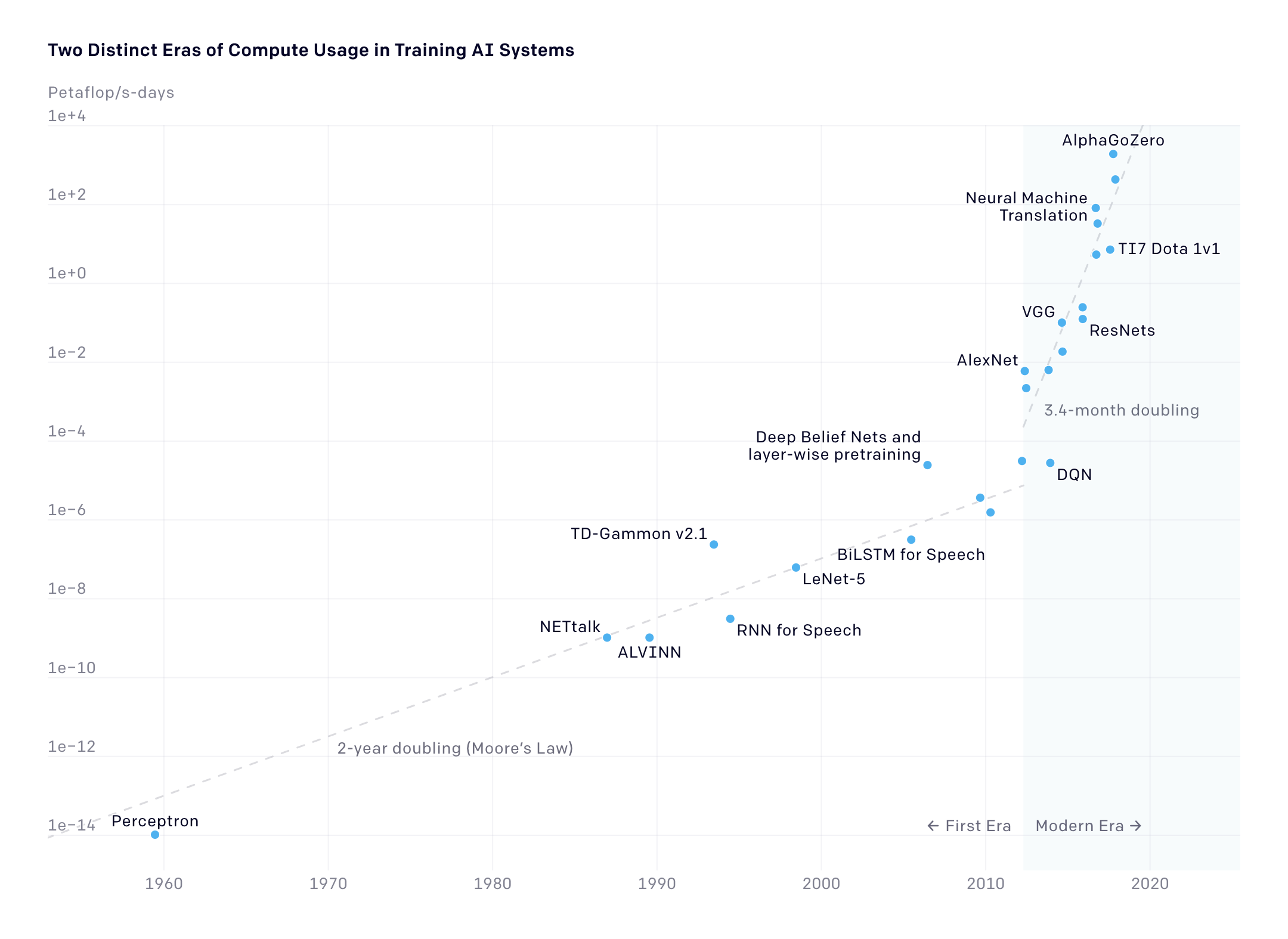}
	\caption{(From \href{https://openai.com/blog/ai-and-compute/}{the OpenAI blog post} \citep{amodei_2018}) Evolution of the number of computations required for training SOTA models in the first era and in the modern era of AI systems.}
	\label{fig:flops_per_day}
\end{figure}

Between $2012$ and $2018$, computations required for DL research have been increased by estimated $300,000$ times which corresponds to doubling the amount of computations every few months \citep{schwartz_2019}, see Figure \ref{fig:flops_per_day}. This rate outruns by far the predicted one by Moore's Law \citep{gustafson_2011}. The performance improvements of DL models are to a great extent induced by raising the number of parameters in the networks and/or increasing the number of computations needed to train and infer the network \citep{strubell_2019,strubell_2020}. 
Orthogonal to the development of new, big SOTA models which need massive amount of data, hardware resources and training time it is important to simultaneously improve parameter, computational, energy and data efficiency of DL models.

\begin{figure}
	\centering
	\includegraphics[width=\linewidth]{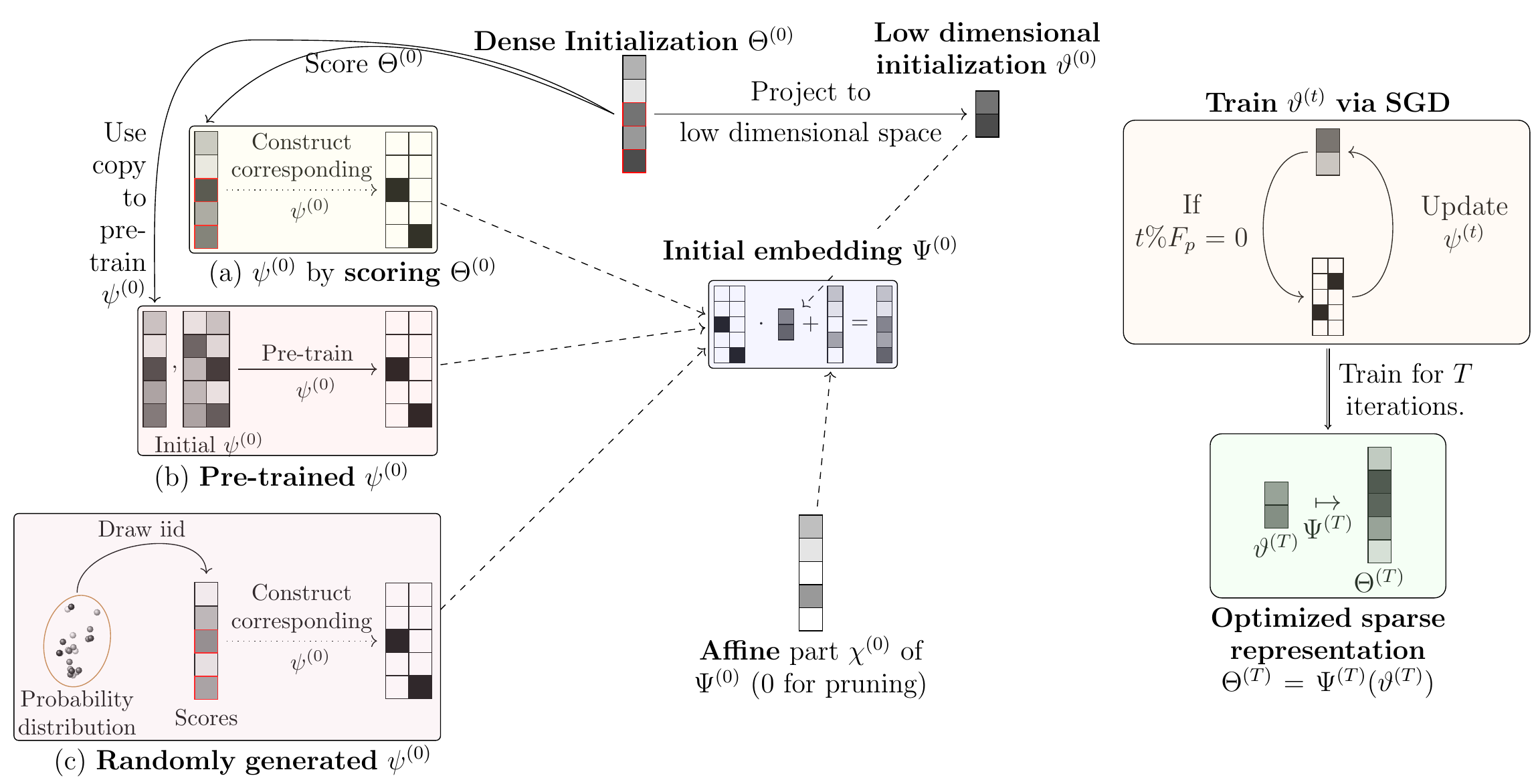}
	\caption{Graphical overview of the proposed dimensionality reduced training methods with a low dimensional initialization $\vartheta^{(0)} \in \R^d$. To use superior trainability of big networks, the low dimensional space must be embedded in a bigger space, modeled by an embedding $\Theta^\z = \Psi^\z (\vartheta^\z) = \psi^\z \cdot \vartheta^\z + \chi^\z$ where $\Theta^\z \in \R^D$, $d \ll D$. Here, $\chi^\z$ is the optional affine part modeling the frozen parameters. The embedding is obtained by either (a) \emph{scoring} the big random initialization $\Theta^\z$, (b) using a pre-training step to generate $\psi^\z$ or (c) mapping $\vartheta^\z$ onto a randomly chosen subset of $\Theta^\z$. During training, only the low dimensional $\vartheta^\z$ is learned. Furthermore, the embedding $\Psi^\t$ can be adjusted to improve results, see Section \ref{sec:dst}.}
	\label{fig:graphical_overview}
\end{figure}

\emph{Model compression} lowers storage and transfer costs, speeds up inference
by reducing the number of computations or accelerates
the training which uses less energy. It can be achieved by methods such as \emph{quantization}, \emph{weight sharing},
\emph{tensor decomposition}, \emph{low rank tensor approximation}, \emph{pruning} or \emph{freezing}.
Quantization reduces the number of bits used to represent the network's
weights and/or activation maps \citep{han_2015,courbariaux_2015,wu_2016,zhou_2017,zhang_2018,jacob_2018,li_2019b}.
$32$bit floats 
are replaced by low precision
integers, thereby decreasing memory consumption and speeding
up inference. Memory reduction and speed up can also be achieved by
weight sharing \citep{nowlan_1992,chen_2015,ullrich_2017}, tensor decomposition
\citep{xue_2013,lebedev_2015,novikov_2015} or low rank tensor approximation \citep{sainath_2013,denton_2014,liu_2015}
to name only a few. 
\emph{Network pruning} \citep{janowsky_1989,mozer_1989,karnin_1990,lecun_1990,han_2015,guo_2016,qian_2021}
sets parts of a DNN's weights to zero. This can help to reduce
the model's complexity and memory requirements, speed up inference \citep{blalock_2020} and even improve the network's generalization ability \citep{lecun_1990,arora_2018,bartoldson_2019,barsbey_2021}. 	

Pruning DNNs can be divided into \emph{structured} and \emph{unstructured} pruning. Structured pruning removes channels, neurons or even coarser structures of the network \citep{anwar_2017,chen_2020,huang_2018,zhuang_2018,zhuang_2020,joo_2021,wang_2020d}. This generates leaner architectures, resulting in reduced computational time. A more fine-grained approach is given by unstructured pruning where single weights are set to zero \citep{mozer_1989,janowsky_1989,karnin_1990,lecun_1990,frankle_2018,lee_2018,wang_2020,tanaka_2020}. Unstructured pruning usually leads to a better performance than structured pruning  \citep{li_2016,mao_2017} but specialized soft- and hardware which supports sparse tensor computations is needed to actually reduce the runtime  \citep{han_2016,parashar_2017}. For an overview of the current SOTA pruning methods we refer to \citet{blalock_2020}.

\emph{Freezing} a DNN means that only parts of the network are trained, whereas the remaining ones are frozen at their initial/pre-trained values \citep{huang_2004,hoffer_2018,rosenfeld_2019,wimmer_2020,sung_2021}. This leads to faster convergence of the networks \citep{huang_2004} and reduced communication costs for distributed training \citep{sung_2021}. Furthermore, freezing reduces memory requirements if the networks are initialized with pseudorandom numbers \citep{wimmer_2020}.

In recent years, \emph{training} only a sparse part of the network, \ie a network with many weights fixed at zero or their randomly initialized value, became of interest to the
DL community \citep{bellec_2018,frankle_2018,lee_2018,tanaka_2020,wang_2020,wimmer_2020,wimmer_2021,wimmer_2021b}, providing the benefit of reduced memory requirements and the potential of reduced runtime not only for inference but also for training. \citet{diffenderfer_2021} even show that sparsely trained models are more robust against distributional shifts than (i) their densely trained counterpart and (ii) networks pruned with \emph{classical} techniques. With classical techniques we denote pruning during or after training where the sparse network is fine-tuned afterwards. A high level overview of recent approaches to prune a network at initialization is given in \citet{wang_2021c}. In our work, the underlying mathematical framework is generalized to involve other dimensionality reducing training methods. Moreover, our work discusses the analyzed methods in more detail. 

\paragraph{Scope of this work}
As mentioned above, there are many possibilities to reduce the cost of a DL framework. In this work, we focus on methods that reduce the number of trainable parameters in a DL model. 
For this, we reversely define dimensionality reduction by embedding a low dimensional space into a high dimensional one, see Section \ref{trafo_intro}. The high dimensional space corresponds to the original DNN, whereas the low dimensional one is the space where the DNN is actually trained. Therefore, training proceeds with reduced dimensionality.

\begin{figure}
	\centering
	\includegraphics[width=\linewidth]{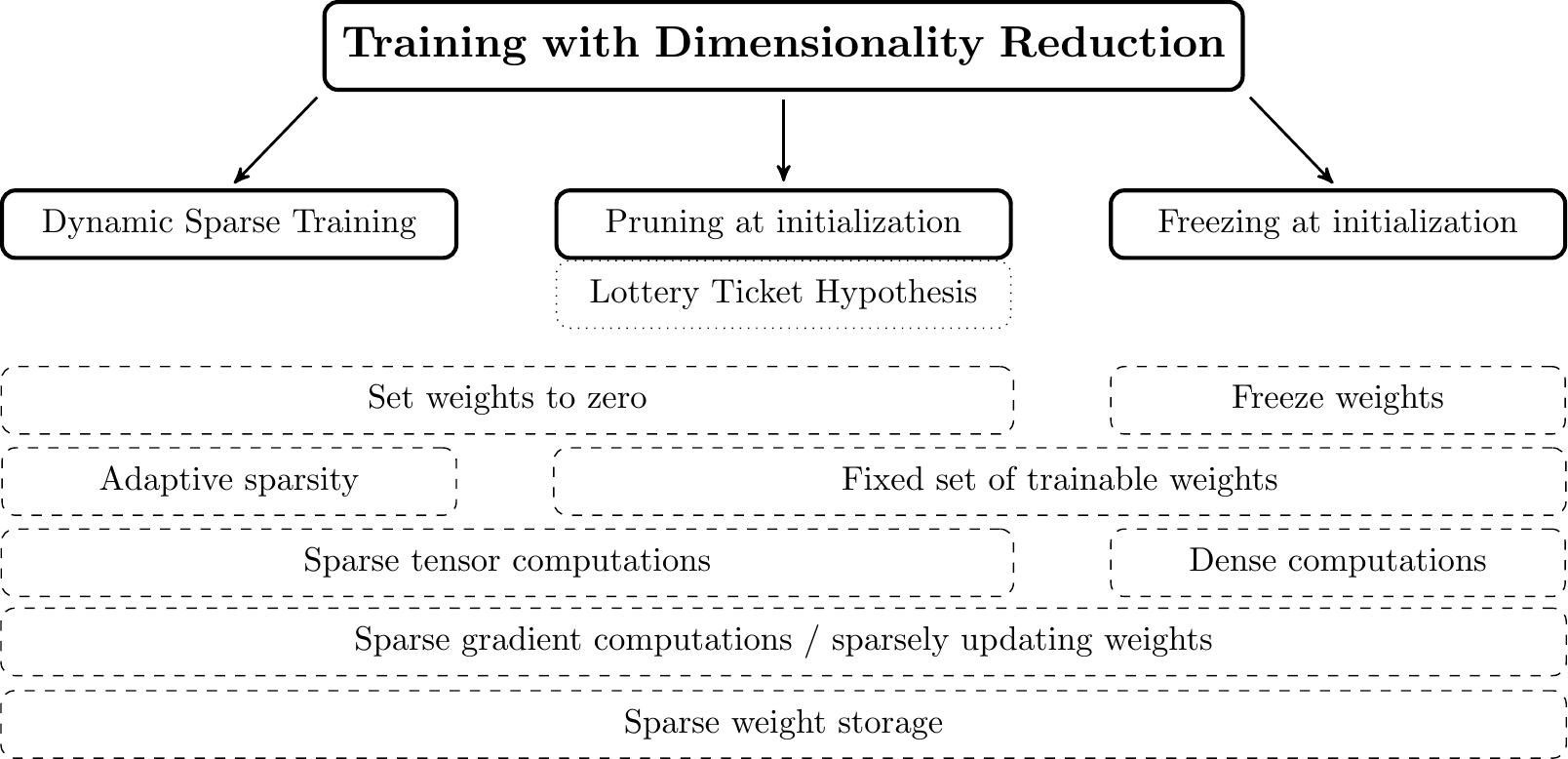}
	\caption{Structural comparison between the three main categories of DRT. Here, the Lottery Ticket Hypothesis is seen as a sub-category of pruning at initialization.}
	\label{fig:structure}
\end{figure}

\emph{Pruning} and \emph{freezing} parts of the network during training are two methods that yield \emph{dimensionality reduced training} (DRT). This survey compares different strategies for pruning and freezing networks during training. Figure \ref{fig:graphical_overview} shows a graphical overview of the proposed DRT methods. A structural comparison between \emph{freezing} DNNs at initialization and \emph{pruning} them is given in Figure \ref{fig:structure}. For pruning, we differentiate between \emph{pruning at initialization} (PaI) which trains a \emph{fixed} set of parameters in the network and \emph{dynamic sparse training} (DST) which adapts the set of trainable parameters during training. Closely related to PaI is the so called \emph{Lottery Ticket Hypothesis} (LTH) which uses well trainable sparse subnetworks, so called \emph{Lottery Tickets} (LTs). LTs are obtained by applying train-prune-reset cycles to the network's parameters, starting with the dense network and finally chiseling out the subnetwork at initialization with desired sparsity. 

\paragraph{Structure of this work}
First we propose the problem formulation and mathematical setup of this survey in Section \ref{sec:problem_formulation}. Then, we discuss different possibilities to reduce the network's dimensionality in Section \ref{sec:trafo}. The LTH and PaI are introduced in Sections \ref{sec:lth} and \ref{sec:pai}, respectively. This is followed by a comparison of different DST methods in Section \ref{sec:dst}. As last DRT method we present freezing initial parameters in Section \ref{sec:freezing}. Subsequently, the different DRT methods from Sections \ref{sec:lth} - \ref{sec:freezing} are compared at a high level in Section \ref{sec:discussion}. The survey is closed by drawing conclusions in Section \ref{sec:conclusions}. 

For better readability, the Figures should best be printed in color or viewed in the colored online version.
\section{Problem formulation}\label{sec:problem_formulation}

We will start Section \ref{sec:problem_formulation} with a general introduction into DL. This is followed by defining a mathematical framework describing DRT. This framework comprises PaI, LTH, DST and freezing parts of a randomly initialized DNN. 

\subsection{General deep learning and setup}\label{subsec:dl_intro}
Let $f_\Theta : \R^m \rightarrow \R^n$ define a DNN with vectorized weights $\Theta \in \R^D \cong \R^{D_1} \times \ldots \times \R^{D_L}$. Here, $D_l$ denotes the number of parameters in layer $l$ of a DNN with $L$ layers and $D = \sum_{l=1}^L D_l$ parameters in total. We further assume the global network structure (activation functions, ordering of layers, ordering of weights inside one layer, \ldots) to be fixed and encoded in $f_\Theta$, \ie only the weight values $\Theta$ can be changed.

We assume a standard DNN training, starting with \emph{random} initial weights $\Theta^\z \in \R^D$. See for example \citet{He2015,xavier_2010,saxe_2014,martens_2010,sutskever_2013,hanin_2018} for possibilities to randomly initialize a DNN. These initial weights are iteratively updated with gradient based optimization like SGD \citep{robbins_2007}, AdaGrad \citep{duchi_2011}, Adam \citep{kingma_2014} or AdamW \citep{loshchilov_2018d} to minimize the loss function 
\begin{equation}
\loss : \R^D \times \R^n \times \R^m \rightarrow \R, \; (\Theta, X, Y) \mapsto \loss(f_\Theta, X, Y)
\end{equation}
over a training dataset $\data = (\X, \Y) \subset \R^n \times \R^m$.\footnote{Unsupervised learning can be modeled by setting $\Y = \emptyset$.} After initialization, a model is trained for $T$ iterations forming a sequence of model parameters $\{ \Theta^\z, \Theta^{(1)}, \ldots, \Theta^{(T)}$\}. Updating the parameters $\Tt$ in iteration $t$ is done by calculating the current gradient
\begin{align}
\nabla_{\Tt} \loss 
= \sum_{b=1}^B \frac{\partial \loss(f_{\Tt}, X_b^\t,Y_b^\t)}{\partial \Tt}\;,
\end{align}
and using it, possibly together with former gradients $\nabla_{\Theta^\z} \loss, \ldots, \nabla_{\Theta^{(t-1)}} \loss$ and parameter values $\Theta^\z, \ldots, \Theta^{(t-1)}$, to minimize the loss function further.
Here, a batch of training data with batch size $B$ is given by $\{(X_b^\t, Y_b^\t): b=1, \ldots, B \} \subset \data$. The batches vary for different training steps $t$.

The final model is then given by $f_{\Theta^{(T)}}$. Note, the overall training goal is for the model to \emph{generalize} well to unseen data. Generalization ability is usually measured on a separate, held back test dataset. Therefore, regularization methods like weight decay \citep{krogh_1991}, dropout \citep{hinton_2012,ba_2013,kingma_2015}, batch normalization \citep{ioffe_2015} or early stopping \citep{yao_2007,finnoff_1993} are used to prevent $f_{\Theta^{(T)}}$ to overfit on the training data. Generalization can also be improved by enabling the model to use geometrical prior knowledge about the scene \cite{cohen_2016,jaderberg_2015,rath_2020,rath_2022,coors_2018}, shifting the model back to an area where it generalizes well \cite{lust_2020b,lust_2022,ren_2019,serra_2020} but also by pruning the network \cite{bartoldson_2019,lecun_1990,hassibi_1993}.
\subsection{Model for dimensionality reduced training}\label{trafo_intro}
A reduction of dimensionality for $\Theta^\t$ at training step $t$ will be modeled through a transformation, also called \emph{embedding}, $\Pt : \R^d \rightarrow \R^D$, $\Tt = \Pt(\ttt)$ and $d \ll D$.\footnote{We use the same transformation as introduced in \citet{mostafa_2019} to embed $\R^d$ into $\R^D$.} In this work, we restrict $\Pt$ to form an affine linear transformation which includes all standard pruning/freezing approaches. However, $\Pt$ has parameters which need to be stored. Therefore, we not only assume $d \ll D$ but also $\size(\Pt) + d \ll D$, where $\size (\cdot)$ computes the minimal number of parameters needed to express the affine linear transformation $\Pt$. To model DST, $\Pt$ is allowed to change during training. 

In our setup, the parameter count in the network is reduced not only after training but also during it. Thus, the trainable parameters of the network are given by $\ttt \in \R^d$. All methods discussed in this work are restricted to fulfill $\ttt \in \R^d$ for all training iterations $t \in \{1, \ldots, T\}$, \ie reduce dimensionality throughout training. The corresponding model with reduced dimensionality is $f_{\Pt(\ttt)}$.

\subsubsection{Pruning and dynamic sparse training}\label{subsubsec:trafo_pruning}
For pruning, $\Pt$ encodes the positions of the non-zero entries whereas $\ttt$ stores the values of those parameters. A corresponding $\Pt$ can be easily constructed by setting
\begin{equation}\label{eq:pruning_trafo}
\Pt(\vartheta^\t) = \pt \cdot \vartheta^\t \;,
\end{equation}
with the \emph{pruning embedding} $\pt \in \R^{D \times d}$ defined via
\begin{equation}\label{eq:pruning_matrix}
\pt_{i,j} = \begin{cases}
1 , & \; \text{if} \; \Tt_i \; \text{is the j\textsuperscript{th} un-pruned element.} \\
0, & \; \text{else}
\end{cases},
\end{equation}
where we assume the $d$ un-pruned elements $\Tt_{i_1^\t}, \ldots, \Tt_{i_d^\t}$ to have the natural ordering $i_1^\t < \ldots < i_d^\t$. Consequently, the pruned version of $\Theta^\t$ is encoded by the low dimensional
\begin{equation}
\vartheta^\t = (\vartheta^\t_j)_j = (\Theta^\t_{i^\t_j})_j\;.
\end{equation}
Further, we define $p := 1- \nicefrac{d}{D}$ as the model's \emph{pruning rate}.
For pruning at initialization, \ie fixed position $i_1^\z, \ldots, i_d^\z$ for the non-zero parameters, $\pt = \psi^{(0)}$ for all training iterations $t$. Whereas, $\pt$ might adapt for DST. 

Since $d \ll D$, the embedding $\Psi^\t(\vartheta^\t)$ is automatically sparse in the bigger space $\R^D$. A possibility to overcome sparse $\Theta^\t$ while still training only a small part $\vartheta^\t$ of a DNN is proposed in \citet{wimmer_2021b}. Here, convolutional $K \times K$ filters are represented via few non-zero coefficients of an adaptive dictionary. The dictionary is shared over one or more layers of the network. This procedure can be described by the adaptive embedding
\begin{equation}\label{eq:trafo_interspace}
\Pt = \Pht \cdot \psi^{(t)}\;
\end{equation}
with the pruning embedding $\psi^{(t)} \in \R^{D \times d}$ as defined in \eqref{eq:pruning_matrix} and the trainable dictionary $\Pht \in \R^{D\times D}$.\footnote{Here, $\Pht$ corresponds to a block diagonal matrix with shared blocks. By sharing blocks, the total parameter count of $\Pht$ is at most $\nicefrac{D}{10,000}$ \citep{wimmer_2021b}. Furthermore, the formulation in \citet{wimmer_2021b} is not restricted to quadratic $D \times D$ matrices, but also allows \emph{undercomplete} or \emph{overcomplete} systems $\Pht$.} In this case, the coordinates \wrt $\Pht$ are sparse, but the resulting representation in the spatial domain $\R^D$ will be dense \citep{wimmer_2021b}. 

Orthogonal to that approach, \citet{price_2021} combine a sparse pruning embedding \eqref{eq:pruning_trafo} with a dense \emph{discrete cosine transformation} (DCT) by summing them up. A DCT is free to store and can be computed with $D \cdot \log D$ FLOPs. By doing so, the network keeps a high information flow while only a sparse part of the network has to be trained. Moreover, the computational cost is almost equal to a fully sparse model.

\subsubsection{Freezing parameters}\label{subsubsec:trafo_freeze}
\begin{figure}
	\centering
	\includegraphics[width=.9\linewidth]{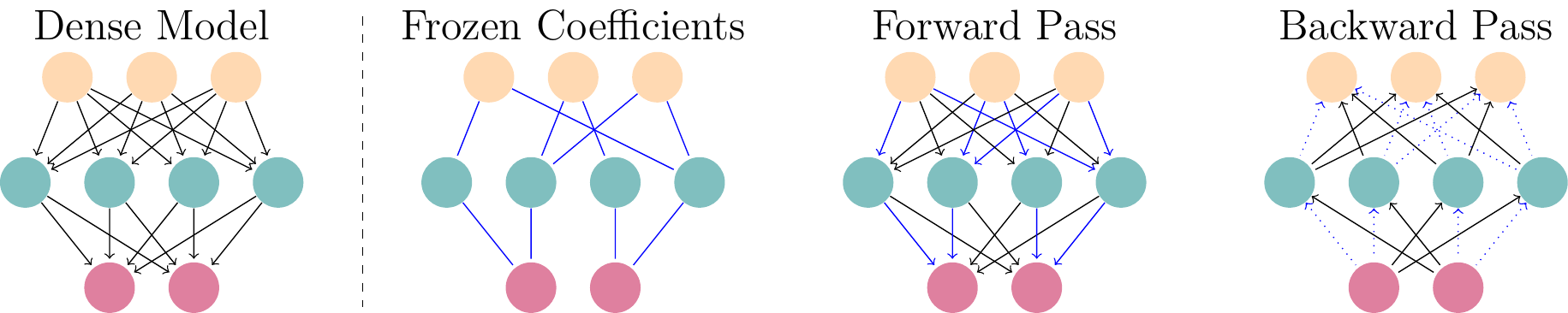}
	\caption{Left: Standard dense model. Middle left: Topology of the frozen weights. Middle right: The forward pass is exactly the same for the dense and the frozen model. Right: The frozen weights are used for the backward pass. However, they are not updated and therefore drawn with dotted arrows.}
	\label{fig:freezing_overview}
\end{figure}
Another approach, which is similar to \citep{price_2021}, proposed by \citet{rosenfeld_2019,wimmer_2020,sung_2021}, uses frozen weights on top of the trainable ones. The resulting transformation is given by
\begin{equation}\label{eq:freezing_trafo}
\Pt(\vartheta^\t) = \pt \cdot \vartheta^\t + \ct \cdot \Theta^\z
\end{equation}
with the pruning embedding $\pt \in \R^{D \times d}$ from \eqref{eq:pruning_matrix}, $\ct = \diag(\ct_1, \ldots, \ct_D) \in \R^{D\times D}$ defined as
\begin{equation}\label{eq:freeze_idx}
\ct_i = \begin{cases}
0 , & \; \text{if} \; i \in \{i_1^\t, \ldots, i_d^\t\} \\
1, & \; \text{else}
\end{cases},
\end{equation}
and the random initialization $\Theta^\z \in \R^D$. A graphical overview of freezing parts of a network is given in Figure \ref{fig:freezing_overview}. Similarly to pruning, we define $p := 1- \nicefrac{d}{D}$ as the model's \emph{freezing rate}, \ie the rate of parameters which are frozen at their random initial value. Note, \citet{rosenfeld_2019} and \citet{wimmer_2020} both use a fixed set of un-frozen weights, \ie $\Pt = \Psi^\z$. Therefore, the network parameters $\Tt = \Psi^\z(\ttt)$ consist of two parts, the dynamic un-frozen weights defined by $\psi^\z \cdot \ttt$ and the fixed, frozen weights $\chi^\z \cdot \Theta^\z$. \citet{sung_2021} allow the embedding $\Psi^{(t)}$ to adapt during training, however their procedure still leaves a large part of the network completely untouched. 

By setting $\ct = 0$, \eqref{eq:pruning_trafo} is only a special case of \eqref{eq:freezing_trafo}, since pruning is naturally the same as freezing un-trained weights at zero. The same transformation \eqref{eq:freezing_trafo}, but restricted on freezing whole layers or even the whole network, is used for \emph{extreme learning machines} \citep{huang_2011,qing_2020} or other works like \citet{saxe_2011,giryes_2016,hoffer_2018}.  

Another construction of $\Tt$ is given by interchanging the pruning embedding $\pt$ in \eqref{eq:freezing_trafo} with a fixed, random orthogonal transformation $\psi_o \in \R^{D \times d}$ \citep{li_2018}. 

\subsection{Storing the embedding $\Pt$}\label{subsec:storing}
In practice, storage techniques like the \emph{compressed sparse row} format \citep{tinney_1967} take over the role of the pruning embedding $\pt$. If $\pt$ is known, $\ct$ can be easily constructed by using equation \eqref{eq:freeze_idx}. Furthermore, by using pseudorandom number generators for the initialization, also $\Theta^\z$ can be recovered by knowing only a single random seed number. In this case, storing $\Psi^\t$ has up to $32$bit the same cost as storing $\psi^\t$. On the other hand, if an additional learnt transformation $\Pht$ is used as well, compare \eqref{eq:trafo_interspace}, the corresponding transformation has to be stored -- this is why \citet{wimmer_2021b} share many elements in $\Pht$.   

\section{High-level overview of dimensionality reducing transformations $\Psi^\t$}\label{sec:trafo}
In this Section we give a high-level overview of different approaches to determine the dimensionality reducing transformation $\Pt$. First, we discuss the \emph{structure} of the parameters which are pruned or frozen in Section \ref{subsec:structure}. Then, we compare \emph{global} and \emph{layer wise} dimensionality reduction in Section \ref{subsubsec:thresholding}. Section \ref{subsec:frequency} covers the \emph{update frequency} for $\Pt$. Afterwards, the most prominent criteria for determining the pruning embedding $\pt$ are proposed in Section \ref{subsubsec:pruning_criteria}. Finally, we discuss \emph{pre-training} the transformation $\Pt$ in Section \ref{subsec:pre_training}.

Throughout this Section, we restrict the transformation $\Pt$ to use $\pt$ from \eqref{eq:pruning_matrix}, \ie a sparse linear embedding $\pt \in \{0,1\}^{D \times d}$ with $\Vert \pt \Vert_0 = d$, as linear part. Further we assume the affine part of $\Psi^\t$ to be either zero (pruning) or correspond to the almost free to store pseudorandom initialization (freezing). We use the notion of \emph{trainable} weights (in iteration $t$) for all weights $\Tt_i$ with $i \in \{i_1^\t, \ldots, i_d^\t\}$, \ie elements of
\begin{equation}
\{i : \exists j \; \text{s.t.}\; \psi^\t_{i,j} \neq 0\}\;.
\end{equation}

\subsection{Structure of trainable weights}\label{subsec:structure}
\begin{figure}
	\centering
	\includegraphics[width=.9\linewidth]{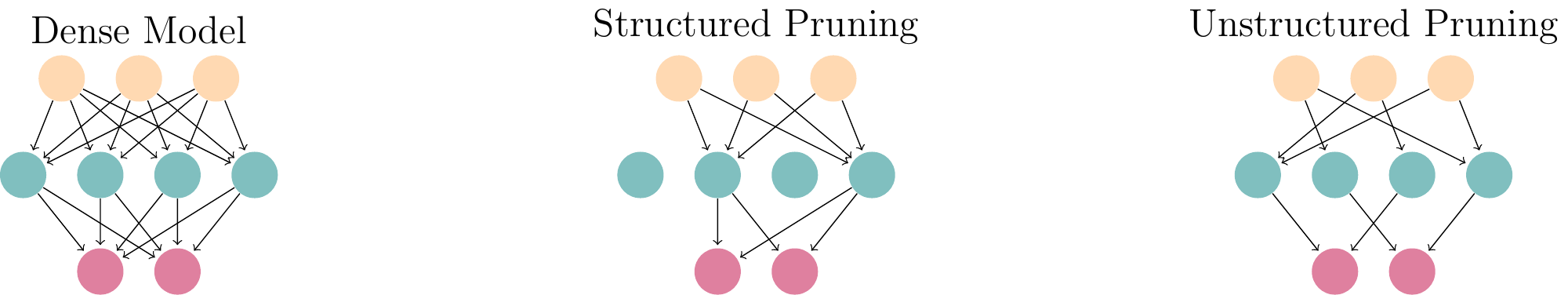}
	\caption{Structured and unstructured pruning.}
	\label{fig:pruning_structure}
\end{figure}
Pruning is generally distinguished in \emph{structured} and \emph{unstructured} pruning, see Figure \ref{fig:pruning_structure}. Of course it can be generalized to our setup, including freezing of parameters. Structured freezing/pruning means freezing/pruning whole neurons or channels or even coarser structures of the network. For pruning, this immediately results in reduced computational costs, whereas gradient computations can be skipped for the frozen structure. For structured freezing, usually whole layers are frozen \citep{huang_2011,qing_2020,saxe_2011,hoffer_2018} with the extreme case of freezing the \emph{entire} network \citep{giryes_2016}. For pruning it is more common to prune on the level of channels/neurons \citep{liu_2018,wang_2020b,verdenius_2020} .

Unstructured pruning of single weights usually improves results compared to structured pruning \citep{li_2016,mao_2017}. Unstructured pruning has the disadvantage to require software which supports sparse computations to actually speed up the forward and backward propagation in DNNs. Furthermore, such software only accelerates sparse DNNs on CPUs \citep{park_2017,liu_2021} or specialized hardware \citep{han_2016,parashar_2017,elsen_2020,gale_2020,wang_2020c}. Weights can also be frozen in an unstructured manner \citep{li_2018,wimmer_2020,sung_2021}, leading to reduced gradient computations, memory requirements and communication costs for distributed training.

In recent years, a semi-structured pruning strategy developed, the so called $N:M$ sparsity \citep{hubara_2021,zhou_2021,sun_2021,pool_2021,holmes_2021,nvidia_2020}. A tensor is defined as $N:M$ sparse if each block of size $M$ contains (at least) $N$ zeros. Here, the tensor is covered by non-overlapping, homogeneous blocks of size $M$. This development is driven by NVIDIA's A100 Tensor Core GPU Architecture \citep{nvidia_2020} which is able to accelerate matrix multiplications up to a factor of almost $2$ if a matrix is $2:4$ sparse.

As shown in \citet{frankle_2018}, using sophisticated methods to determine a sparse, fixed subset of trainable parameters at random initialization greatly outperforms choosing them randomly. Contrarily, choosing and fixing the trainable channels randomly or via well performing classical techniques leads to similar results for structured pruning at initialization \citep{liu_2018}. Therefore, most pruning methods presented and discussed in this work are unstructured. Consequently, we assume, if not mentioned otherwise, unstructured pruning/freezing in the following.


\subsection{Global or layerwise dimensionality reduction}\label{subsubsec:thresholding}
An important distinction for DRT methods is whether the rate of the trained parameters is chosen separately for each layer or globally. As presented in Section \ref{subsubsec:pruning_criteria}, the \emph{importance} of a weight $\Theta_i^\z$ for training is measured via a score $s_i \in \R$. A weight is frozen or pruned if the corresponding score is below a threshold $\tau_i$\footnote{We index the threshold with the same index as the weight to show that the threshold might depend on the position of $\Theta_i^\z$ but not the weight itself.}. This threshold can be determined \emph{globally} \citep{wimmer_2020,lee_2018,frankle_2018,wang_2020,sanh_2020,bellec_2018,mostafa_2019} or \emph{layerwise}. For thresholding the score layerwise we differentiate between setting a constant rate of trainable parameters in each layer \citep{jayakumar_2020}, using heuristics or optimized hyperparameters to find the best rate of trainable parameters for each layer individually \citep{mocanu_2018,evci_2020} or letting the number of trained parameters in each layer \emph{adapt} dynamically during training \citep{dettmers_2019,mostafa_2019}. 

\subsection{Update frequency}\label{subsec:frequency}
\begin{figure}
	\centering
	\includegraphics[width=\linewidth]{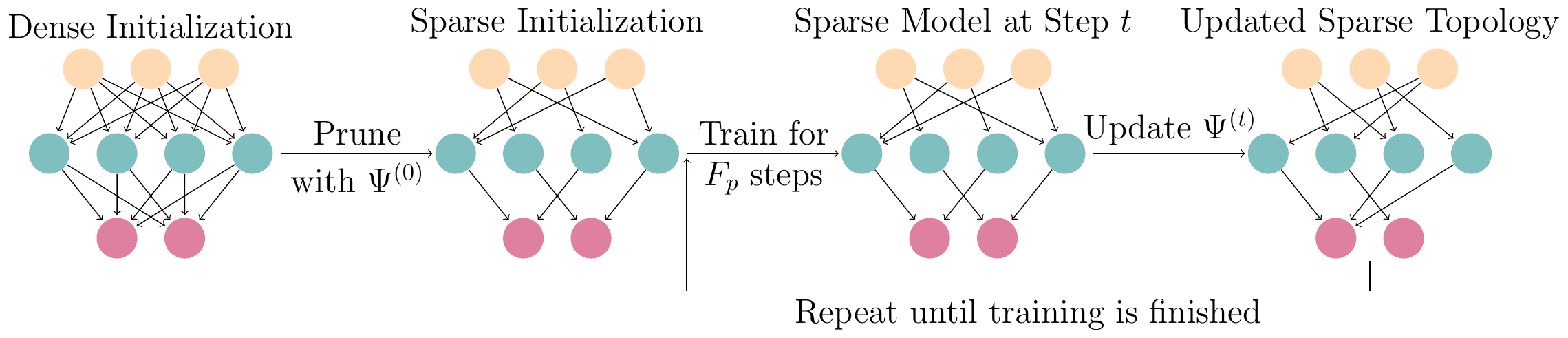}
	\caption{Dynamic sparse training where the initial pruning embedding $\Psi^\z$ is usually determined randomly.}
	\label{fig:dst}
\end{figure}
The frequency of updating the dimensionality reducing transformation $\Psi^\t$ is called \emph{update frequency} $F_p$, see Figure \ref{fig:dst}. Given an update frequency $F_p \in \mathbb{N}^+ \cup \{\infty\}$, it holds
\begin{equation}
\Psi^{(t_0)} = \Psi^{(t_0 + 1)} = \ldots = \Psi^{(t_0 + F_p - 1)}\;
\end{equation}
where $t_0 \in \{0, F_p, 2F_p , \ldots \}$ is an iteration where the transformation is updated. If $\Psi^\t = \Psi^{(0)}$ is constant, the update frequency equals $F_p = \infty$. We assume the number of trained parameters to be fixed. Consequently, the proposed methods usually have $F_p \gg 1$. This guarantees (i) stability of the training and (ii) a chance for newly trainable weights to grow big enough to be not pruned/frozen immediately in the next update step.

\subsection{Criteria to choose trainable parameters}\label{subsubsec:pruning_criteria}
In the following, we present the most common criteria to determine the trainable weights.
\paragraph{Random criterion}
Randomly selecting trainable parameters means that a given percentage of them are chosen to be trained by a purely random process, see Figure \ref{fig:graphical_overview} (c). This can be done by sampling a score for each weight from a standard Gaussian distribution and training those weights with the $d$ biggest samples. Random pruning is often used as first comparison for a newly developed pruning method. For dynamic sparse training, initially trained parameters are often chosen by random \citep{mocanu_2018,evci_2020} as the dynamics of the sparse topology will find a well performing architecture anyway.
Also for DRT methods where information flow is not a limiting factor, like PaI combined with DCTs \citep{price_2021} or freezing \citep{rosenfeld_2019,li_2018}, trainable weights are often chosen randomly.

\paragraph{Magnitude criterion}
A straight forward way to determine trainable weights is to use those with the highest magnitude. This is done by the LTH \citep{frankle_2018,zhou_2019} and also by many DST methods for updating the sparse architecture during training \citep{mocanu_2018,mostafa_2019,dettmers_2019,evci_2020}. Magnitude pruning for $\Theta$ is equivalent to the solution of
\begin{equation}\label{eq:mag_pruning}
\min_{\bar{\Theta} \in \R^D, \Vert \bar{\Theta}\Vert_0 \leq d} \Vert \Theta - \bar{\Theta}\Vert_q \;,
\end{equation}
\ie the best $d$-sparse approximation of $\Theta$ \wrt $\Vert \cdot \Vert_q$. Here, $q \in (0, \infty)$ is arbitrary. Assume $\mat (\Theta) \in \R^{n \times m}$ to be the matrix representing a linear network with $m$ dimensional input and $n$ dimensional output with corresponding vectorized parameters $\Theta \in \R^{n \cdot m}$. With $q=2$ it holds for $x \in \R^m$
\begin{align}
\Vert \mat(\Theta) x - \mat(\bar{\Theta}) x \Vert_2 & \leq \Vert \mat(\Theta - \bar{\Theta}) \Vert_F \Vert x \Vert_2  \\ &= \Vert \Theta - \bar{\Theta} \Vert_2 \Vert x \Vert_2 \label{eq:approxmation_ability_mag_pruning}\;.
\end{align}
For a $d$-sparse $\bar{\Theta}$, the right hand side \eqref{eq:approxmation_ability_mag_pruning} is minimized by the solution of magnitude pruning. This shows the ability of magnitude pruning to approximate dense networks sparsely if the pruned parameters are not too big. In practice, this is guaranteed by pruning only small fractions of the parameters in one step \citep{frankle_2018,frankle_2020a,han_2015,liu_2018}. The magnitude criterion in the viewpoint of dynamical systems is analyzed in \citet{redman_2022}. They show that magnitude pruning is equivalent to pruning small modes of the Koopman operator \citep{mezic_2005} which determines the networks convergence behavior. Consequently, pruning small magnitudes only slightly disturbs the long term behavior of DNNs.

\citet{lee_2021} analyze the minimal $\ell_2$ distortion induced by pruning the whole network altogether. Their greedy solution approximates magnitude pruning with \emph{layerwise optimal pruning ratios}. The solution is obtained by rescaling the magnitudes with a weight-dependent factor, applying global magnitude pruning and finally reset the un-pruned weights to their value before the rescaling.

If dimensionality reduction is applied at initialization without any pre-training, the magnitude criterion chooses trainable parameters purely by their initial, randomly drawn weight. Still as shown in \citet{frankle_2021}, magnitude pruning at initialization is a non-trivial baseline for other PaI methods and outperforms random PaI.

\paragraph{Gradient based criterion}
As discussed above, the magnitude criterion is for the most part random at initialization. Consequently, different criteria to chose the trainable weights are used for SOTA PaI methods. The most common gradient based criterion relies on the first order Taylor expansion of the loss function at the beginning of training \citep{lee_2018,lee_2019,verdenius_2020,jorge_2020,hayou_2021,wimmer_2020,sanh_2020} and is mainly used for PaI in our setup. It measures how disturbing the weights at initialization will affect the loss function. It holds
\begin{equation}\label{eq:grad_taylor}
\loss (\Theta^\z - \delta) = \loss (\Theta^\z) - \nabla_{\Theta^\z} \loss \cdot \delta + \mathcal{O}(\Vert \delta \Vert^2)\;.
\end{equation}
In our setup, the disturbance introduced by the dimensionality reduction is given by $\delta = \Theta^\z - \Psi^\z ( \vartheta^\z)$. Neglecting the higher order terms and the sign, the following optimization problem has to be solved
\begin{equation}\label{eq:grad_criterion_general}
\max_{\Psi, \vartheta} \vert \nabla_{\Theta^\z} \loss \vert \cdot \vert \Psi (\vartheta) \vert \;,
\end{equation}
in order to determine $\Psi^\z$ and $\vartheta^\z$. If $\Psi$ is restricted to be a pruning transformation defined by \eqref{eq:pruning_trafo} and \eqref{eq:pruning_matrix}, the optimization problem \eqref{eq:grad_criterion_general} reduces to 
\begin{equation}\label{eq:snip_criterion}
\max_{\{i_1^\z, \ldots, i_d^\z\} \subset \{1,\ldots D\}} \sum_{j=1}^d \vert (\nabla_{\Theta^\z} \loss)_{i_j} \cdot \Theta^\z_{i_j} \vert \;,
\end{equation}
which is solved by the indices with top-$d$ $\vert (\nabla_{\Theta^\z} \loss)_{i} \cdot \Theta^\z_{i} \vert$.

Using the aforementioned gradient based criterion for PaI might lead to a loss of information flow for high pruning rates since some layers are pruned too aggressively \citep{wang_2020,tanaka_2020,wimmer_2020}. Without information flow in the network, the gradient is vanishing and no training is possible. Thus, there are several approaches to overcome a weak gradient flow in networks pruned according to \eqref{eq:snip_criterion}. Using a random initialization, fulfilling the layerwise \emph{dynamic isometry} property \citep{saxe_2014,poole_2016,schoenholz_2017,xiao_2018} will lead to an improved information flow in the pruned network \citep{lee_2019}. Another approach \citep{hayou_2021} uses a rescaling of the pruned network to bring it to the \emph{edge of chaos} \citep{poole_2016,schoenholz_2017,xiao_2018} which is benefitial for DNN training. 
A straight forward way, proposed by \citet{wimmer_2020}, is given by freezing the un-trained weights instead of pruning them. 

\paragraph{Conserving information flow}
Sufficient information flow can also be guaranteed directly by the criterion for selecting trainable parameters. \citet{wang_2020} train the sparse network with the highest total gradient $\ell_2$ norm \emph{after pruning}. Consequently, the information flow is not the bottleneck for training.

\emph{Synaptic saliency} of weights is introduced in \citet{tanaka_2020}. A synaptic saliency is defined as 
\begin{equation}\label{eq:synaptic_saliency}
S(\Theta) = \nabla_{\Theta} \mathcal{R} \odot \Theta\;,
\end{equation}
where $\mathcal{R} : \R^D \rightarrow \R$ is a function, dependent on the weights $\Theta$. In \citet{tanaka_2020}, a conservation law for a layer's total synaptic saliency is shown. By iteratively pruning a small fraction of parameters based on a synaptic saliency score \eqref{eq:synaptic_saliency}, the conservation law guarantees a faithful information flow in the sparse network. Examples for a synaptic flow based pruning criterion are setting $\mathcal{R}(\Theta)$ as the $\ell_1$ \emph{path norm} \citep{neyshabur_2015,neyshabur_2015b} of the DNN $f_\Theta$ \citep{tanaka_2020} or the $\ell_2$ path norm \citep{gebhart_2021,patil_2021}. In \citet{patil_2021}, the $\ell_2$ path norm is combined with a random walk in the parameter space to gradually \emph{build} up the sparse topology. In contrast, standard approaches \emph{slim} the dense architecture into a sparse one.

\paragraph{Trainable $\psi^\z$}
All aforementioned criteria are based on heuristics, usually limited to work well in some scenarios but fail for other ones, see \citet{frankle_2021} for an empirical comparison for the SOTA PaI methods \citet{lee_2018,wang_2020,tanaka_2020}. A natural way to overcome these limitations is given by \emph{training} the embedding $\psi^\z$. This approach is especially in the interest for pruning randomly generated networks \emph{without} training the weights afterwards, where the determination of $\psi^\z$ is the only way to optimize the network \citep{zhou_2019,ramanujan_2019,diffenderfer_2021,aladago_2021}.

Finding $\psi^\z \in \{0, 1\}^{D \times d}$ can not be done by simple backpropagation since $\psi^\z$ is optimized in a \emph{discrete} set $\{\psi \in \{0,1\}^{D \times d}: \Vert \psi \Vert_0 = d\}$. \citet{ramanujan_2019,mallya_2018,diffenderfer_2021,aladago_2021,zhang_2021} find the pruned weights with a trainable score $s \in \R^D$ together with a (shifted) sign function. The score $s$ is optimized via backpropagating the error of the sparse network on the training set by using the straight through estimator \citep{bengio_2013} to bypass the zero gradient of the sign function. 
Overcoming the vanishing gradient of zeroed scores can also be achieved by training a pruning probability for each weight and sampling a corresponding $\psi \in \{0,1\}^{D \times d}$ in each optimization step \citep{zhou_2019}.

\subsection{Pre-training the transformation}\label{subsec:pre_training}
Similar to a trainable $\psi^\z$, discussed in Section \ref{subsubsec:pruning_criteria}, also a pre-training step for finding $\psi^\z$ can be applied. The difference is that for pre-training $\psi^\z$ the corresponding dense weights $\Theta^\z$ can be trained as well. Using pre-training for finding the initial transformation $\psi^\z$ is of course costly in terms of time and also in terms of an increased parameter count during the pre-training phase. Note, \emph{after} pre-training $\psi^\z$, the pre-trained weights are \emph{not allowed} to be used, but reset to $\Theta^\z$. 

The most prominent example for this is given by the LTH which trains $\Theta^\z$ to convergence, prunes $p_0 \cdot 100\%$ of the non-zero weights and resets the remaining non-zero weights to their initial values. This procedure is continued until the desired pruning rate is reached. Then, the pruning transformation $\psi^\z$ is applied to the initial weights $\Theta^\z$ and fixed for the training of the sparse architecture. Here, the \emph{full} network is used for finding a well performing sparse architecture while for the actual training only the sparse part of the network, starting from the random initialization, is used.

Another approach for pre-training $\psi^\z$ is given in \citet{liu_2020} which pre-trains a sparse network to mimic the training dynamics of a randomly initialized dense network, measured by the \emph{neural tangent kernel} \citep{jacot_2018,arora_2019,lee_2019b}.\footnote{To be precise, in order to mimic the training dynamics, \citet{liu_2020} also pre-train the weights $\vartheta^\z$ by using $\Theta^\z$. Therefore, it is not within the narrow bounds of this work.}


\section{Lottery ticket hypothesis}\label{sec:lth}
\begin{table}[tb!]
	\centering
	\addtolength{\leftskip} {-4cm}
	\addtolength{\rightskip}{-4cm}
	\scalebox{0.565}{
		\begin{tabular}{@{}lccccccc@{}}
			\toprule
			{Method}  & {\makecell{Prune \\ criterion}} & {\makecell{Late \\ rewinding}} & {\makecell{Early \\ stopping}}  & {\makecell{Low \\ precision}} & {\makecell{Data \\ size}} & {Transfer} & {\makecell{Additional \\ Trafo $\Phi$}} \tabularnewline
			\midrule 
			\citet{frankle_2018} & $\vert \Theta^\z_i \vert$ & \xmark & \xmark & \xmark & \statcirc{black} & \xmark & \xmark\\
			\citet{frankle_2020a}& $\vert \Theta^\z_i \vert$ & \cmark & \xmark & \xmark & \statcirc{black} & \xmark & \xmark \\
			\citet{morcos_2019}& $\vert \Theta^\z_i \vert$ & \cmark & \xmark & \xmark & \statcirc{black} & Datasets/Optimizers & \xmark  \\
			\citet{zhang_2021b}& $\vert \Theta^\z_i \vert$ & \cmark & \cmark & \xmark &  \statcirc[white]{black} & \xmark & \xmark\\
			\citet{you_2019}& $\vert \Theta^\z_i \vert$ & \xmark & \cmark & \cmark & \statcirc{black} & \xmark & \xmark\\
			\citet{rosenfeld_2019}& $\vert \Theta^\z_i \vert$ & \cmark & \xmark & \xmark & \statcirc{black} & Predict Performance & \xmark \\
			\citet{wimmer_2021b}& $\vert \Theta^\z_i \vert$ & \cmark & \xmark & \xmark & \statcirc{black} &  \xmark & Trainable\\
			\citet{lee_2021}& {\small $\vert \Theta^\z_i \vert \left( \sum_{\vert \Theta^\z_j \vert \geq \vert \Theta^\z_i\vert} \vert \Theta^\z_j \vert \right)^{-1}$} & \cmark & \xmark & \xmark & \statcirc{black} &  \xmark & \xmark \\
			\citet{zhang_2021}& $\vert \Theta^\z_i \vert$ & \cmark & \xmark & \xmark & \statcirc{black} & \xmark & Inertial Manifold\\
			\bottomrule
		\end{tabular}	
	}
	\caption{\label{tab:lth}Comparison of different methods to find lottery tickets. Data size \statcirc{black} means using the full dataset for finding LTs whereas \statcirc[white]{black} shrinks the full dataset during the process of finding LTs. Note, all methods can be used in an iterative manner, but also as one-shot methods.}
\end{table}
\begin{figure}
	\centering
	\includegraphics[width=\linewidth]{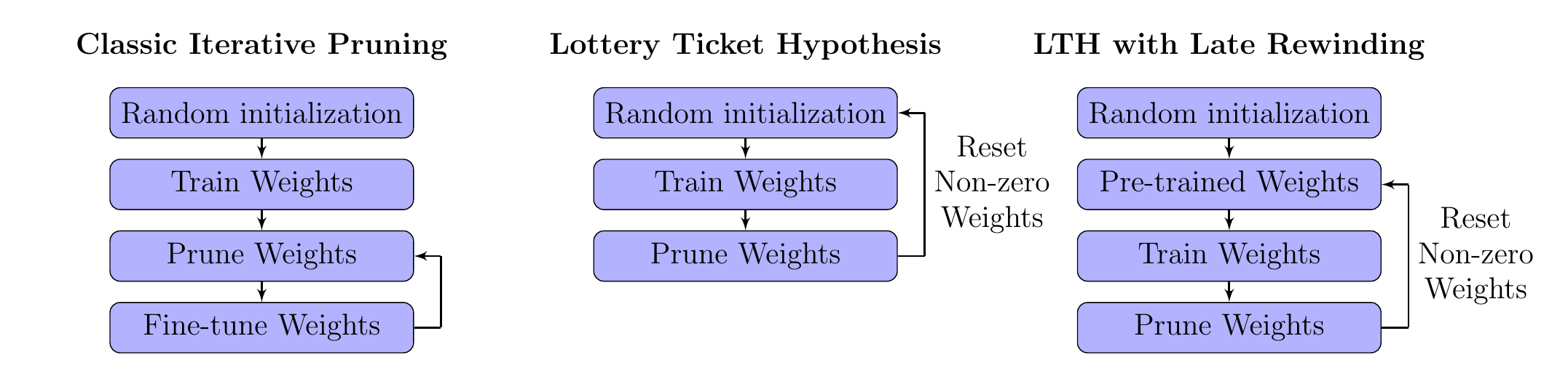}
	\caption{Left: Classical iterative pruning applied to a converged model, like \citet{han_2015}. Middle: Standard Lottery Ticket Hypothesis. Right: LTH with resetting weights to values from an early training step.}
	\label{fig:lth_flow_chart}
\end{figure}
In this Section we discuss the LTH, proposed by \citet{frankle_2018}, in detail. \citet{frankle_2018} show that extremely sparse subnetworks of a randomly initialized network can be found which, after being trained, match or even outperform their densely trained baseline. Furthermore, the sparsely trained network converges at least as fast as standard training but usually faster. Table \ref{tab:lth} summarizes different methods to find LTs. Figure \ref{fig:lth_flow_chart} compares the LTH with classical pruning methods which are applied to converged networks \citep{han_2015}. 

The procedure to determine the sparse networks is as follows: First, the dense network is trained to convergence and the pruning embedding $\psi$ is constructed according to the magnitude criterion, see Section \ref{subsubsec:pruning_criteria}. The un-pruned weights are reset to their values $\Theta^\z_i$. Now, this procedure can be done \emph{one-shot} or \emph{iterative}. In the one-shot case, all weights are pruned after training the dense network at once. Then, the sparse network, the so called \emph{Lottery Ticket} (LT), is trained to convergence. For the iterative procedure, not all coefficients but only $p_0 \cdot 100\%$ (usually $20\%$) of the un-pruned ones are pruned in one iteration. The remaining non-zero weights are reset to their initial value and trained again until convergence. 
This procedure is applied iteratively until the desired pruning rate $p$ is achieved. Finally, the LT is optimized in a last, sparse training.

Iterative LTH has shown to find better performing LTs than the one-shot approach \citep{frankle_2018,frankle_2020a}. But, in order to reach a final pruning rate $p$, the iterative procedure takes
\begin{equation}
\left \lceil \frac{\log ( 1-p )}{\log 4 - \log 5} \right \rceil
\end{equation}
many pre-trainings plus the final, sparse training. For the final pruning rate $p=0.9$, the iterative LTH approach needs in total $12$ trainings of the network. On the other hand, the one-shot approach always requires $2$ trainings for arbitrary pruning rates $p$.

\citet{frankle_2018,gale_2019,frankle_2020a} show that resetting un-pruned weights to their \emph{initial} value only generates well trainable sparse architectures if the networks are not too big. For modern network architectures like ResNets \citep{he_2016}, resetting the weights to their initial value does not lead to similar results as the densely trained baseline. This problem can be overcome by \emph{late rewinding}, \ie resetting the weights to values reached early in the first, dense training \citep{frankle_2020a,frankle_2020b,morcos_2019}. Moreover, pruning coefficients \wrt a dynamic, adaptive basis for $K \times K$ convolutional filters improves late rewinding even further \citep{wimmer_2021b}. 

\citet{lee_2021} use scaled magnitudes to improve results for LTs at high sparsity levels. Their scale approximates the \emph{best choice of layerwise sparsity ratios}. LTs as \emph{equilibria of dynamical systems} are analyzed in \citet{zhang_2021}. They theoretically show that the important dynamics of SGD training are contained in a small $d$ dimensional subspace $W^+ \subset \R^D$, the generator of the so called \emph{inertial manifold}. Only a few training epochs suffice to reliably compute $W^+$. Rewinding un-pruned weights to their values for an early training step and projecting them onto $W^+$ improves results compared to the standard LT with late rewinding. Theoretical guarantees for recovering sparse linear networks via LTs are described in \citet{elesedy_2021}. 

Costs for finding LTs via iterative magnitude pruning can be reduced by using early stopping and low precision training for each pre-training iteration \citep{you_2019}, sharing LTs for different datasets and optimizers \citep{morcos_2019} or iteratively reducing the dataset together with the number of non-zero parameters \citep{zhang_2021b}. Also, the error of a LT from a given family of network configurations (\ie ResNet with varying sparsity, width, depth and number of used training examples) can be well estimated by knowing the performance of only a few trained networks from this network family \citep{rosenfeld_2021}. Despite their first applications on image classification, LTs have also shown to be successful in self-supervised learning \citep{chen_2021b}, natural language processing \citep{yu_2020,chen_2020b}, reinforcement learning tasks \citep{yu_2020,vischer_2022}, transfer learning \citep{soelen_2019} and object recognition tasks like semantic segmentation or object detection \citep{girish_2021}. Moreover, \citet{chen_2021} propose methods to verify the ownership of LTs and hereby protect the rightful owner against intellectual property infringement. 

LTs outperform randomly reinitializing sparse networks in the unstructured pruning case \citep{frankle_2018}. Contrarily, for structured pruning there seems to be no difference between randomly reinitializing the sparse network and resetting the weights to their random initialization \citep{liu_2018}. 

Closely related to the LTH with late rewinding, \citet{renda_2020,le_2021} show that fine-tuning a pruned network with a learning rate schedule rewound to earlier stages in training outperforms classical fine-tuning of sparse networks with small learning rates. \citet{bai_2022} show that \emph{arbitrary} randomly chosen pruning masks can lead to successful sparse models \emph{if} the full network is used during training. For this, they extrude information of weights, which will be pruned eventually, into the sparse network during a pre-training step. Thus, \citet{bai_2022} \emph{is not} a DRT method.
\section{Pruning at initialization}\label{sec:pai}
\begin{table}[tb!]
	\centering
	\addtolength{\leftskip} {-4cm}
	\addtolength{\rightskip}{-4cm}
	\scalebox{0.6}{
		\begin{tabular}{@{}lccccc@{}}
			\toprule
			{Method} & {Criterion} & {One-shot}  & {Pre-train $\psi^\z$} & {Train $\vartheta^\z$} & Initialization of $\vartheta^\z$  \tabularnewline
			\midrule 
			\citet{lee_2018} & $\vert \Theta^\z \odot g^\z \vert$ & \cmark & \xmark & \cmark & standard \\
			\citet{lee_2019} & $\vert \Theta^\z \odot g^\z \vert$ & \cmark & \xmark & \cmark & dynamic isometry \\
			\citet{hayou_2021} & $\mathbb{E} \vert \Theta^\z \odot g^\z \vert^2$ & \cmark & \xmark & \cmark & edge of chaos \\		
			\citet{jorge_2020} &  $\vert \Theta^\z \odot g^\z \vert$ & \xmark & \xmark & \cmark & standard \\
			\citet{wang_2020} & $ - \Theta^\z \odot H^\z g^\z$  & \cmark & \xmark & \cmark & standard \\
			\citet{tanaka_2020} & $ \Theta^\z \odot \nabla_{\Theta^\z} \mathcal{R}$ & \xmark  & \xmark & \cmark & standard \\
			\citet{wimmer_2021} &{\small $\lambda \vert \Theta^\z \odot g^\z \vert + (1-\lambda) ( \Theta^\z \odot H^\z  g^\z)$ } & \cmark & \xmark & \cmark & standard \\
			\citet{su_2020} & random (+ prior knowledge) & \cmark & \xmark & \cmark & standard \\
			\citet{patil_2021} & path norm & \xmark  & \xmark & \cmark & standard \\
			\citet{lubana_2021} & $\vert \Theta^{(0)} \odot g^{(0)} \odot \Theta^{(0)} \vert $ & \cmark  & \xmark & \cmark & standard \\
			\citet{lubana_2021} & $\vert \Theta^\z \odot H^\z g^\z \vert $ & \cmark  & \xmark & \cmark & standard \\
			\citet{zhang_2020} & temporal Jacobian & \cmark  & \xmark & \cmark & standard \\
			\citet{alizadeh_2022} & meta-gradient & \cmark  & \cmark & \cmark & standard \\
			\citet{zhou_2019} & trainable prune probability & \xmark & \cmark & \xmark & standard \\
			\citet{ramanujan_2019} & trainable score $s$ & \xmark & \cmark & \xmark & standard \\
			\citet{diffenderfer_2021} & trainable score $s$ & \xmark & \cmark & \xmark & binarization \\
			\citet{koster_2022} & trainable score $s$ (incl. sign swap)& \xmark & \cmark & \xmark & binarization \\
			\citet{chen_2022} & $ \Theta^\z \odot \nabla_{\Theta^\z} \mathcal{R}$ (+ trained sign swap)& \xmark & \cmark & \xmark & standard \\
			\citet{aladago_2021} & trainable quality score & \xmark & \cmark & \multicolumn{2}{c}{$\vartheta^\z_k$ out of fixed $ \{w^{(1)}_k, \ldots, w^{(n)}_k\}$} \\
			\bottomrule
		\end{tabular}	
	}
	\caption{\label{tab:pai}Comparison of different methods for PaI. Here, $\Theta^\z \in \R^D$ denotes the dense, random initialization of the model, $g^\z$ the gradient and $H^\z$ the Hessian of the loss at beginning of training. The function $\mathcal{R} : \R^D \rightarrow \R$, used for \citet{tanaka_2020}, can be chosen as any almost everywhere differentiable function.}
\end{table}

Overcoming the high cost for the train-prune-reset cycle(s) needed to find LTs is one of the main motivation to use pruning at initialization (PaI) \citep{lee_2018}. With pruning at initialization we mean methods that start with a randomly initialized network and do not perform any pre-training of the network's weights to find the pruning transformation $\psi^\z$. Furthermore for PaI, $\pt = \psi^\z$ holds for all training iterations $t$. Different variants of PaI methods include one-shot pruning \citep{lee_2018,lee_2019,wang_2020,zhang_2020,hayou_2021,wimmer_2021,wimmer_2021b,alizadeh_2022}, iterative pruning \citep{tanaka_2020,jorge_2020,verdenius_2020,patil_2021} and training the pruning transformation \citep{zhou_2019,ramanujan_2019,diffenderfer_2021,aladago_2021}. On top of that, there are methods that train the network after pruning and methods that do \emph{not train} the non-zero parameters at all. Table \ref{tab:pai} summarizes the PaI methods proposed in this Section. 

\subsection{PaI followed by training non-zero weights}
First, we start with comparing different methods that train weights after the pruning step. We can group most of them into \emph{gradient} based approaches and \emph{information flow} based approaches. For a detailed comparison of the three popular PaI methods \citet{lee_2018,tanaka_2020,wang_2020} we refer to \citet{frankle_2021}. Moreover, \citet{fischer_2022} compares them on generated tasks, where known and extremely sparse target networks are \emph{planted} in a randomly initialized model. They show that current PaI methods fail to find those sparse models at extreme sparsity. However, the performance of other pruning methods than PaI is not evaluated and it remains an open question if they are able to find these planted subnetworks. 
  
\paragraph{Gradient based approaches}
Gradient based approaches \citep{lee_2018,lee_2019,verdenius_2020,jorge_2020,hayou_2021} try to construct the sparse network which has the best influence in changing the loss function at the beginning of training, as presented in equations \eqref{eq:grad_taylor}, \eqref{eq:grad_criterion_general} and \eqref{eq:snip_criterion}. However, it was shown that one-shot gradient based approaches have the problem of a vanishing gradient flow, if too many parameters are pruned \citep{lee_2019,wang_2020,wimmer_2020}. Methods to overcome this are given by using an iterative approach \citep{jorge_2020,verdenius_2020} or use an initialization of the network, adjusted to the sparse network \citep{lee_2019,hayou_2021}. These adjusted initialization include so called \emph{dynamic isometric} networks \citep{lee_2019} and networks \emph{at the edge of chaos} \citep{hayou_2021}.

\paragraph{Methods preserving information flow}
Other PaI methods primarily focus on generating sparse networks with a sufficient information flow \citep{wang_2020,tanaka_2020,zhang_2020,patil_2021}. For randomly initialized DNNs, the loss is no better than chance. Therefore, \citet{wang_2020} argue that ``at the beginning of training, it is more important to preserve the training dynamics than the loss itself.'' Consequently, \citet{wang_2020} try to find the sparse network with the highest gradient norm after pruning. They do so, by training the weights with highest $- \Theta^\z_i (H^\z \nabla_{\Theta^\z}\loss )_i$, where $H^\z$ defines the Hessian of the loss function at initialization.  Another approach is given by \citet{tanaka_2020} and \citet{patil_2021}, which try to maximize the path norm in the sparse network, see Section \ref{subsubsec:pruning_criteria} \emph{Conserving information flow} for more details. For RNNs or LSTMs, standard PaI methods do not work well \citep{zhang_2020}. By pruning based on singular values of the temporal Jacobian, \citet{zhang_2020} are able to preserve weights that propagate a high amount of information through the network's temporal depth.

\paragraph{Hybrid and other approaches}
As shown in \citet{wimmer_2021}, only optimizing the sparse network to have the highest information flow possible does not lead to the best sparse architectures -- even for high pruning rates. They conclude that information flow is a \emph{necessary} condition for sparse training but not a \emph{sufficient} one. Therefore, they combine gradient based and information flow based methods to get the best out of both worlds. With their approach, they improve gradient based PaI and PaI based on preserving information flow at one blow. 

Another method guaranteeing a faithful information flow in the network and at the same time improving performance is given by \emph{pruning in the interspace} \citep{wimmer_2021b}. \citet{wimmer_2021b} represent $K \times K$ filters of a convolutional network in the \emph{interspace} -- a linear space spanned by an underlying filter basis. After pruning the filter's coefficients, the filter basis is trained jointly with the non-zero coefficients. By adapting the interspace during training, networks tend to recover from low information flow. Furthermore, using interspace representations has shown to improve not only PaI for gradient based and information flow preserving methods, but also LTH, DST, freezing \emph{and} training dense networks.

An orthogonal approach for guaranteeing high information flow while training only a sparse part of the network is proposed by \citet{price_2021}. Random pruning is combined with a bypassing DCT in each layer. Since DCTs correspond to basis transformations with an orthonormal basis, the information of a layer's input is maintained even if only a tiny fraction of parameters is trained. Therefore, results are improved tremendously for extreme pruning rates. Note, adding DCT improves performance for high $p$ despite using a simple random selection of trained weights. As shown by \citet{price_2021}, adding DCTs to each layer also improves DST methods for high $p$.

\citet{alizadeh_2022} improve \citet{lee_2018} by modeling the effect of pruning on the loss function at training iteration $M > 0$. In contrast, \citet{lee_2018} only analyze the effect of pruning on the loss at initialization. To do so, \citet{alizadeh_2022} compute a \emph{meta-gradient} which is achieved by pre-training the dense network for $M$ epochs. The meta-gradient is computed \wrt pruning mask. After pruning, the non-zero weights are reset to their initialization.

\citet{lubana_2021} theoretically compare \emph{magnitude, gradient based and information flow preserving} PaI methods. Their analysis shows that magnitude based approaches lead to a rapid decrease in the training loss, thus converge fast. Furthermore, gradient based pruning conserves the loss function, removes the slowest changing parameters and preserves the first order dynamics of a model's evolution. They combine magnitude and gradient based pruning to get the best of both worlds which yields the pruning score $\vert \nabla_{\Theta^{(0)}} \loss_{i} \vert \cdot \vert \Theta^{(0)}_i \vert^2$. Finally, information flow preserving pruning by using \citet{wang_2020}'s criterion to get the sparse model with maximal gradient norm removes the weights which maximally increase the loss function. However, this does not preserve the second order dynamics of the model. \emph{Preserving} the gradient norm by training the weights with highest $\vert \Theta^\z_i (H^\z \nabla_{\Theta^{(0)}} \loss)_i \vert$ on the other hand also preserves the second order dynamics and shows improved results compared to \citep{wang_2020}.

\citet{su_2020,frankle_2021} show that the most popular PaI methods \citet{lee_2018,wang_2020,tanaka_2020} do not significantly lose performance if positions of non-zero weights are shuffled randomly in each layer if the sparsity is not too extreme. Consequently, these methods do not appear to find the best sparse architecture, but rather well performing layerwise pruning rates for the given network architecture and global pruning rate. Using the knowledge of well performing PaI methods, \citet{su_2020} show impressive results by randomly pruning weights. Hereby, they use layerwise pruning rates derived by observing the layerwise pruning rates of other well performing PaI methods. \citet{liu_2022} also show that random PaI can reach competitive results to other PaI methods. They experimentally demonstrate that the gap between random PaI and dense training gets smaller if the underlying baseline network becomes bigger. Therefore, random PaI can provide a strong sparse training baseline, especially for large models.  

\subsection{PaI without training non-zero weights}\label{subsec:pai_random}
\begin{figure}
	\centering
	\includegraphics[width=\linewidth]{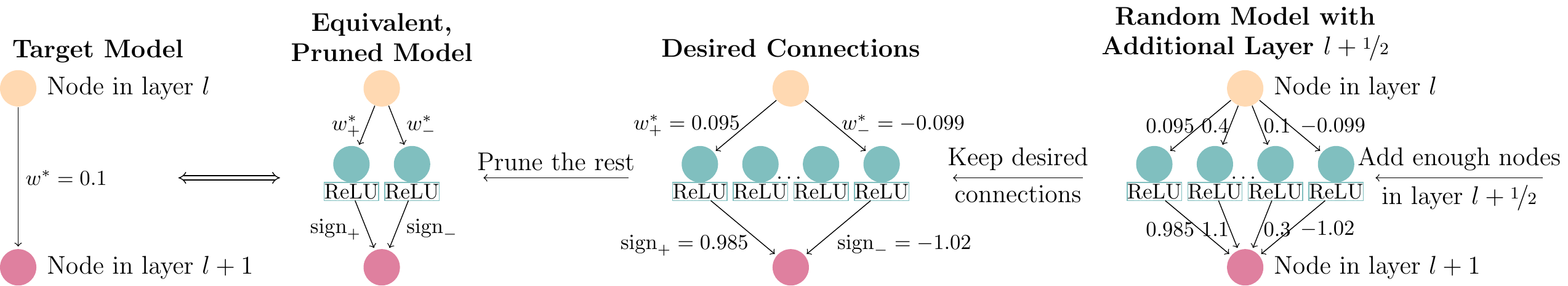}
	\caption{Shows how \citet{malach_2020} approximate a single target weight through random connections by adding a wide enough layer $l+\nicefrac{1}{2}$ between layers $l$ and $l+1$ and afterwards pruning unneeded connections.}\label{fig:pruning_randomized_networks}
\end{figure}
\citet{mallya_2018} show that a pre-trained network can be pruned for a specific task to have good performance even without fine-tuning. Inspired by that, several PaI methods showed that the same holds true for a randomly initialized network. Before we go into detail \emph{how} pruning masks for random weights can be found, we will discuss theoretical works that cover the \emph{universal approximator} ability of pruning a randomly initialized network, also called \emph{strong lottery ticket hypothesis} \citep{malach_2020,pensia_2020}.

\paragraph{Theoretical background}
An intuition that big, randomly initialized networks contain well performing sparse subnetworks is given in \citet{ramanujan_2019}. Let $f^\ast_{\Theta^\ast}: \R^{m} \rightarrow \R^{n}$ be a target network with $\Theta^\ast \in \R^{d^\ast}$. Further, let $f_\Theta : \R^{m} \rightarrow \R^{n}$ be a network with randomly initialized ${\Theta} \in \R^D$ and $\tilde{\Theta}$ be a randomly chosen sparse projection of $\Theta$ with $\Vert \tilde{\Theta}\Vert_0 = d \ll D$ non-zero parameters. Consequently, $f_{\tilde{\Theta}}$ is an arbitrary $d$-sparse subnetwork of $f_{{\Theta}}$. Assuming $D \gg d^\ast \approx d$, \citet{ramanujan_2019} argue that the chance of $f_{\tilde{\Theta}}$ being a well approximator of $f^\ast_{\Theta^\ast}$ is small, but equal to $\delta > 0$. The big, random network has $\binom{D}{d}$ subnetworks with $d$ non-zero parameters. Consequently, the chance of \emph{all} sparse subnetworks $f_{\tilde{\Theta}}$ not being a good approximator for $f^\ast_{\Theta\ast}$ is
\begin{equation}
(1-\delta)^{\binom{D}{d}} \xrightarrow[D \to \infty]{} 0\;.
\end{equation}
Thus, if the randomly initialized network $f_{\Theta}$ is chosen big enough, it will contain, with high chance, a well approximator for $f^\ast_{\Theta^\ast}$ with $d$ non-zero weights.

This intuition is proven in \citet{malach_2020} for MLPs by using a slightly different idea. Here, instead of making each layer in the randomly initialized network arbitrary wide, an intermediate layer $l + \nicefrac{1}{2}$ is added between each two layers $l$ and $l+1$ of the randomly initialized network, see Figure \ref{fig:pruning_randomized_networks} right. These intermediate layers are made wide enough so that each weight $w^\ast$ in the target network can be approximated well enough. It holds
\begin{equation}
 w^\ast \cdot x = \underbrace{(+1)}_{\approx \textnormal{sign}_{+}} \cdot \relu(\underbrace{\vert w^\ast \vert}_{\approx w^\ast_{+}} \cdot x) + \underbrace{(-1)}_{\approx \textnormal{sign}_{-}} \cdot \relu(\underbrace{- \vert w^\ast \vert}_{\approx w^\ast_{-}} \cdot x)\;.
\end{equation}
Therefore, for each weight $w^\ast$ in the target network two paths are constructed. The first one approximates the positive part $\vert w^\ast \vert$ together with sign $+1$ and the other the negative part $-\vert w^\ast \vert$ together with sign $-1$, see middle right of Figure \ref{fig:pruning_randomized_networks}. All remaining weights in the random network are pruned as shown in the middle left of Figure \ref{fig:pruning_randomized_networks}. As a consequence, \citet{malach_2020} show that, with some assumptions on the target network, pruning big randomly initialized networks is an universal approximator. But, for each target weight, a polynomial number of intermediate weights have to be added.
Follow up works \citet{orseau_2020} and \citet{pensia_2020} improve the parameter efficiency by shrinking the big, randomized network. They do so by using more than two paths to approximate a weight in the original network. By allowing the random values to be resampled, the width of the intermediate layer $l + \nicefrac{1}{2}$ can be reduced to $2$ times the size of the original layer $l$ \citep{chijiwa_2021}. \citet{burkholz_2022} allow the randomly initialized network to be deeper than $2$ times the target network. By approximating layers with combinations of univariate and multivariate linear functions, they are also able to approximate convolutional layers. Independent to the approach of \citet{burkholz_2022}, \citet{cunha_2022} extend the result of \citet{pensia_2020} to CNNs by restricting the network's input to be non-negative. 

\paragraph{Methods to train $\psi^\z$}
As already mentioned before, training $\psi^\z \in \{0,1\}^{D\times d}$ requires optimizing in a discrete space. Consequently, different approaches are used to find $\psi^\z$ for a randomly initialized network. In \citet{zhou_2019}, weights are pruned with probabilities modeled by Bernoulli samplers with corresponding trainable parameters. While helping during training, the stochasticity of this procedure may limit the performance at testing time \citep{ramanujan_2019}. The stochasticity is overcome in \citet{ramanujan_2019} by the \texttt{edge-popup} algorithm. For each weight $\Theta^\z_i$ in the network a corresponding pruning score $s_i$ is trained. The weights with top-$d$ scores are kept at their initial value, the remaining ones are pruned. The score $s_i$ is then optimized with the help of the so called \emph{straight through} estimator \citep{bengio_2013} and the pruning transformation is updated in each iteration. \citet{ramanujan_2019} show that a random Wide-ResNet50 \citep{zagoruyko_2016} contains a sub-network with smaller size than a ResNet34 \citep{he_2016} but the same test accuracy on ImageNet \citep{deng_2009}. \citet{koster_2022} and \citet{chen_2022} independently show that allowing the randomly initialized weights to switch signs, improves the performance of \texttt{edge-popup}. \citet{chijiwa_2021} improve \texttt{edge-popup} by allowing re-sampling of the un-trained weights. 
Binarizing the un-pruned, random weights can also be combined with the \texttt{edge-popup} algorithm \citep{diffenderfer_2021}. Finally, \citet{aladago_2021} sample for each weight $\vartheta^\z_i$ in a target network a set of $n$ \emph{possible weights} $w^{(1)}_i, \ldots, w^{(n)}_i$ in a \emph{hallucinated}, $n$ times bigger network. This hallucinated network is fixed. During training, each possible weight $w^{(j)}_i$ has a corresponding \emph{quality score} which is increased if the weight is a good choice for $\vartheta^\z_i$ and decreased otherwise. In the end, the $w^{(j)}_i$ with the highest quality score is set as $\vartheta^\z_i$ whereas the remaining hallucinated weights are discarded, \ie pruned.

\section{Dynamic sparse training}\label{sec:dst}
\begin{table}[tb!]
	\centering
	\addtolength{\leftskip} {-4cm}
	\addtolength{\rightskip}{-4cm}
	\scalebox{0.64}{
		\begin{tabular}{@{}lcccccc@{}}
			\toprule
			\multirow{2}{*}[-0.5\dimexpr \aboverulesep + \belowrulesep + \cmidrulewidth]{Method} & \multirow{2}{*}[-0.5\dimexpr \aboverulesep + \belowrulesep + \cmidrulewidth]{\shortstack{Sparse \\ initialization}} & \multicolumn{2}{c}{Sparsity} &\multirow{2}{*}[-0.5\dimexpr \aboverulesep + \belowrulesep + \cmidrulewidth]{\shortstack{Pruning \\ criterion}} & \multirow{2}{*}[-0.5\dimexpr \aboverulesep + \belowrulesep + \cmidrulewidth]{\shortstack{Regrow \\ criterion}} & \multirow{2}{*}[-0.5\dimexpr \aboverulesep + \belowrulesep + \cmidrulewidth]{$F_p$} \\
			\cmidrule(l){3-3}\cmidrule(r){4-4}
			&  & {Global} & {Adaptive} &  & &    \tabularnewline
			\midrule 
			\citet{bellec_2018} & random & \xmark & \xmark & sign change & random & triggered by pruning \\
			\citet{mocanu_2018} & random & \xmark & \xmark & magnitude & random & hyperparameter \\
			\citet{mostafa_2019} & random & \cmark & \cmark & magnitude & random & hyperparameter \\
			\citet{dettmers_2019} & magnitude & \xmark & \cmark & magnitude & gradient momentum & 1 epoch \\
			\citet{evci_2020} & random & \xmark & \xmark & magnitude & gradient & hyperparameter \\
			\bottomrule
		\end{tabular}	
	}
	\caption{\label{tab:dst}Comparison of different methods for dynamic sparse training.}
\end{table}
For high pruning rates, PaI is not able to perform equally well as classical pruning methods. One possible explanation for the performance gap is that a randomly initialized network does not contain enough information to find a suitable sparse subnetwork that trains well. Adapting the pruning transformation during training or using network pre-training to find $\psi^\z$ overcomes this lack of information. As shown in Section \ref{sec:lth}, finding LTs is expensive. Furthermore, it is not answered if resetting weights to their initial value can match late rewinding for big scale datasets. Dynamic sparse training performs well for high pruning rates while only needing one, fully sparse training. This is achieved by redistributing sparsity over the network in the course of training. By always fixing a given rate of parameters at zero, the number of trained weights is kept constant during training. Different DST methods are collected in Table \ref{tab:dst}.

\subsection{Dynamic sparse training methods}
Inspired by the rewiring of synaptic connectivity during the learning process in the human brain \citep{chambers_2017}, DEEP-R \citep{bellec_2018} trains sparse DNNs while allowing the non-zero connections to rewire during training. To do so, pruning and rewiring is modeled as stochastic sampling of network configurations from a posterior. However, this procedure is computationally costly and challenging to apply to big networks and datasets \citep{dettmers_2019}.

In \citet{mocanu_2018}, a sparse subnetwork with $d$ non-zero parameters is chosen randomly at the beginning of training. Here, an initial pruning rate has to be defined separately for every layer. After each epoch, trained weights with the smallest magnitude are pruned layerwise with rate $q$ and the same number of weights is regrown at random positions in this layer. Note, the pruning rate $q$ used for \emph{updating} $\pt$ is different from the \emph{initial} pruning rate $p$. This procedure is done until training converges. Thus, in each epoch the same number of parameters is pruned and regrown which makes training hard to converge. Furthermore, the number of non-zero parameters needs to be defined for each layer before training and can not be adapted during training. 

\emph{Dynamic sparse reparameterization} \citep{mostafa_2019} overcomes this problem by using magnitude pruning with an adaptive, global threshold. Furthermore, the number of regrown weights in each layer is adapted proportionally to the number of non-zero weights in that layer. 

Regrowing weights not randomly, but based on their gradient's momentum was proposed by \citet{dettmers_2019}. Here, not only the regrown weights are determined by their momentum, but also their number in each layer is. 

RigL \citep{evci_2020} bypasses the need of \citet{dettmers_2019} to compute the dense gradient in each training iteration, and only regrows weights based on their actual gradient in the updating iteration of $\psi^\t$, \ie $t \in \{F_p, 2 F_p, \ldots\}$. Also, the number of pruned and regrown parameters is reduced by a cosine schedule to accelerate and improve convergence. 

Overparameterized, densely trained DNNs in combination with SGD have shown good generalization abilities \citep{du_2019,li_2018b,brutzkus_2018}. \citet{liu_2021b} showed that the good generalization ability of DST models can be explained by the so called \emph{in-time-over-parameterization}. Training of sparse DNNs can be viewed in the \emph{space-time} manifold. To overparameterize DST models in this manifold, three properties must be fulfilled:
\begin{enumerate}
	\item The dense baseline network has to be big enough.
	\item Exploration of trained weights has to be guaranteed during training.
	\item The training time has to be long enough so that the network can test enough sparse architectures in training.
\end{enumerate}
If in-time-overparameterization is guaranteed for DST methods by these three criteria, \citet{liu_2021b} show that sparse networks are able to outperform the dense, standard overparameterized ones. In-time-overparametrization can be achieved by either increasing the number of training epochs, or by reducing the batch size while keeping $F_p$ constant. The latter leads to more updates of $\pt$ while not increasing the number of training epochs.

\subsection{Closely related methods}
We want to highlight that there exists more methods which are called \emph{dynamic sparse training} in literature which do not fulfill our definition of it. We do not present these methods here in detail since they allow to update \emph{all} weights of the network and only mask them out in the forward pass. Examples for such methods are \citet{guo_2016,ding_2019,kusupati_2020,sanh_2020,liu_2020b}. Other dynamic training methods \citep{jayakumar_2020,schwarz_2021,zhou_2021b,zhou_2021c} sample different subnetworks for each training iteration, update this subnetwork while keeping all un-trained parameters fixed at their previous position. In the next training iteration, a new subnetwork is sampled. Therefore, they can not store the network's parameters $\Theta^{(t)}$ in its reduced form $\vartheta^{(t)}$ for all $t \in \{0, 1, \ldots, T\}$. \citet{schwarz_2021} additionally exchange the standard weights $\Theta^{(t)}$ through a reparametrization using the power  $\Theta^{(t)} = \phi^{(t)} \vert \phi^{(t)} \vert^{(\alpha - 1)}$ with $\alpha > 1$. By doing so, weights close to $0$ are unlikely to grow. As a consequence, the parameters will form a heavy tailed distribution at convergence. This improves results since the parameters are pruned based on their magnitude and heavy tailed distributions are highly compressible via magnitude pruning \citep{barsbey_2021}. Using this reparametrization might also improve other DRT methods proposed in this work. 

\section{Freezing parts of a network}\label{sec:freezing}
\begin{table}[tb!]
	\centering
	\addtolength{\leftskip} {-4cm}
	\addtolength{\rightskip}{-4cm}
	\scalebox{0.655}{
		\begin{tabular}{@{}lcccc@{}}
			\toprule
			{Method} & {frozen part} & structured & {criterion} & {additional projection} \tabularnewline
			\midrule 
			ELMs \citep{huang_2004} & all except classifier & \cmark & --- & \xmark  \\
			\citet{hoffer_2018} & only classifier & \cmark & --- & \xmark \\
			\citet{rosenfeld_2019} & all except batch normalization & \cmark & --- & \xmark \\
			\citet{li_2018} & unstructured & \xmark & random & orthogonal projection \\
			\citet{zhou_2019} & unstructured & \xmark & iterative magnitude (LTH) & \xmark \\
			\citet{wimmer_2020} & unstructured & \xmark & $\vert \Theta^\z \odot g^\z \vert$ & \xmark \\
			\citet{sung_2021} & unstructured & \xmark & $\vert \nabla_{\Theta^{(0)}} \log f_{\Theta^{(0)}}\vert^2$ & \xmark \\
			\citet{rosenfeld_2019} & structured & \cmark & random & \xmark \\ 
			\bottomrule
		\end{tabular}	
	}
	\caption{\label{tab:freeze}Comparison of different methods for freezing parts of a neural network. Here, $\Theta^\z$ denotes the dense, random initialization of the network and $g^\z$ the gradient of the loss function at beginning of training.}
\end{table}
Contrarily to pruning, \emph{freezing} parts of a randomly initialized network has attracted less interest in research in recent years. The main reason for this is that frozen, non-zero weights must be accounted for in the forward propagation. Thus, no (theoretical) speed up for inference can be obtained. In addition, frozen weights must be stored even after training, while zeros only require a small memory footprint. But, by using pseudorandom number generators for initializing the neural network, frozen weights can be recovered with a single 32bit integer on top of the memory cost for a pruned network \citep{wimmer_2020}. The proposed methods to freeze a network are summarized in Table \ref{tab:freeze}. 

\subsection{Theoretical background for freezing}
\citet{saxe_2011} show that convolutional-pooling architectures can be inherently frequency selective while using random weights. In \citet{giryes_2016}, euclidean distances and angles between input data points are analyzed while propagating through a $\relu$ network with random i.i.d. Gaussian weights. With $\relu$ activation functions, each layer of the network shrinks euclidean distances between points inversely proportional to their euclidean angle. This means that points with a small initial angle migrate closer towards each other the deeper the network is. Assuming the data \emph{behaves well}, meaning data points in the same class have a small angle and points from different classes have a bigger angle between another, random Gaussian networks can be seen as an \emph{universal system that separates any data}. On the other hand, if the data is not perfect, training might be needed to overcome big intra-class angles or small inter-class angles to achieve good generalization.

Freezing parameters at their initial value was used in \citet{li_2018} in order to measure the \emph{intrinsic dimension} of the objective landscape. First, the dense network is optimized to generate the best possible solution. Then, the number of frozen parameters is gradually decreased, starting by freezing all parameters. The network's parameters are computed according to \eqref{eq:freezing_trafo}, with $\psi^\t = \psi^\z$ equaling a random, orthogonal projection. If a similar performance as the dense solution is reached, the number of non-frozen parameters determines the intrinsic dimension of the objective landscape. While using frozen weights only as a tool to measure the dimension of a problem, \citet{li_2018} showed that freezing parameters can lead to competitive results. 

\subsection{Freezing methods}
We want to start with the so called \emph{extreme learning machines} (ELMs) \citep{huang_2004,huang_2011,qing_2020}. ELMs are MLPs, usually with one hidden layer. The parameters of the hidden node are frozen and usually initialized randomly. However, the classification layer is trained by a closed form solution of the least-square regression \citep{huang_2004}. ELMs are universal approximators if the number of hidden nodes is chosen big enough \citep{huang_2011}. In SOTA networks, ELMs can be used to substitute the classification layer \citep{qing_2020}. Closely related to ELMs are \emph{random vector functional link neural networks} \citep{pao_1992,pao_1994} which, in addition, allow links between the input and output layer. 

A completely orthogonal approach to ELMs is proposed in \citet{hoffer_2018}, where the classification layer is substituted by a random orthogonal matrix. For the classification layer, only a temperature parameter $T$ is learnt. Experiments show that the random orthogonal layer with optimized $T$ yields comparable results to a trainable classifier for modern architectures on CIFAR-10/100 \citep{krizhevsky_2012} and ImageNet \citep{deng_2009}.

In \citet{zhou_2019}, pruning weights and freezing them at their initial values is compared in the LTH framework. They show that freezing usually performs better for a low number of trained parameters whereas pruning has better results if more parameters are trained. Finally, they show that a combination of pruning and freezing -- depending whether a weight moved towards zero or away from zero during dense training -- reaches the best results. 

Freezing at initialization is used in \citet{wimmer_2020} to overcome the problem of vanishing gradients for PaI. Similar to \citet{zhou_2019} they show that freezing outperforms pruning if only few parameters are trained. For high freezing rates, freezing still guarantees a sufficient information flow in the sparsely trained network. On the other hand, if a higher number of parameters is trained, pruning also performs better than freezing in this setting. But, by using weight decay on the frozen parameters, \citet{wimmer_2020} are able to get the best from both worlds, frozen weights at the beginning of training to ensure faithful information flow and sparse networks in the end of it, while improving both of them at the same time.

Inspired by transfer learning, \citet{sung_2021} freeze large parts of the model to reduce the size of the newly learned part of the network. Frozen weights are chosen according to their importance for changing the network's output, measured by the Fisher information matrix. They show that freezing parameters helps to reduce communication costs in distributed training as well as memory requirements for checkpointing networks during training. Unsurprisingly, they reach the best results if the freezing mask is allowed to change during training compared to keeping the freezing mask fixed. 

\citet{rosenfeld_2019} freeze parameters in a structured way. They mainly focus on training only a fraction of filters (\ie output channels) and determine the frozen parts randomly. In this setting, freezing outperforms pruning for almost all numbers of trained weights.

Furthermore, it is shown in \citet{rosenfeld_2019,frankle_2020} that freezing \emph{all} weights except the trainable batch normalization \citep{ioffe_2015} parameters leads to non-trivial performance. 


\section{Comparing and discussing different dimensionality reduced training methods}\label{sec:discussion}
In this Section we discuss the different DRT approaches introduced in Sections \ref{sec:lth}, \ref{sec:pai}, \ref{sec:dst} and \ref{sec:freezing}. Figure \ref{fig:structure} shows a high-level comparison between LTH/PaI, DST and freezing. Leaning on this structural comparison, we will now discuss different aspects of the methods.

\subsection{Performance.}
\begin{table}[tb!]
	\centering
	\addtolength{\leftskip} {-4cm}
	\addtolength{\rightskip}{-4cm}
	\scalebox{0.735}{
		\begin{tabular}{@{}lccccccc@{}}
			\toprule
			\multirow{2}{*}[-0.5\dimexpr \aboverulesep + \belowrulesep + \cmidrulewidth]{Method} &
			\multirow{2}{*}[-0.5\dimexpr \aboverulesep + \belowrulesep + \cmidrulewidth]{Category} & \multirow{2}{*}[-0.5\dimexpr \aboverulesep + \belowrulesep + \cmidrulewidth]{Top-1-Acc} & 
			\multirow{2}{*}[-0.5\dimexpr \aboverulesep + \belowrulesep + \cmidrulewidth]{\shortstack{Training \\ FLOPs}} &
			\multirow{2}{*}[-0.5\dimexpr \aboverulesep + \belowrulesep + \cmidrulewidth]{\shortstack{Testing \\ FLOPs}} &
			\multirow{2}{*}[-0.5\dimexpr \aboverulesep + \belowrulesep + \cmidrulewidth]{Top-1-Acc} & 
			\multirow{2}{*}[-0.5\dimexpr \aboverulesep + \belowrulesep + \cmidrulewidth]{\shortstack{Training \\ FLOPs}} &
			\multirow{2}{*}[-0.5\dimexpr \aboverulesep + \belowrulesep + \cmidrulewidth]{\shortstack{Testing \\ FLOPs}}  \\
			& & & & & & & \\
			\midrule
			Dense & --- &$76.8 \pm 0.1$ & $1 \times$ & $1 \times$ & $76.8 \pm 0.1$ & $1 \times$ & $1 \times$ \\
			\cmidrule(lr){3-8}
			& & \multicolumn{3}{c}{pruning rate $p=0.8$} & \multicolumn{3}{c}{pruning rate $p=0.9$} \\
			\cmidrule(lr){3-5} \cmidrule(lr){6-8}
			Random & PaI & $70.6 \pm 0.1$ &  $0.23 \times$ & $0.23 \times$ & $65.8 \pm 0.0$ & $0.10 \times$ & $0.10 \times$ \\
			\citet{lee_2018} & PaI & $72.0 \pm 0.1$ & $0.23 \times$ & $0.23 \times$ & $67.2 \pm 0.1$ & $0.10 \times$ & $0.10 \times$ \\
			\midrule
			\citet{frankle_2020a} & LTH &  $75.8 \pm 0.1$  \\
			\midrule
			\citet{evci_2020} & DST&  $74.6 \pm 0.1$ & $0.23 \times$ & $0.23 \times$ & $72.0 \pm 0.1$ & $0.10 \times$ & $0.10 \times$ \\
			+ \citet{wimmer_2021b} & DST&  $76.0 \pm 0.1$ & & & $74.3 \pm 0.1$ &  &  \\
			\citet{evci_2020} $\times 5$ & DST&  $76.6 \pm 0.1$ & $1.14 \times$ & $0.23 \times$ & $75.7 \pm 0.1$ & $0.52 \times$ & $0.10 \times$ \\
			\citet{mocanu_2018} & DST &  $72.9 \pm 0.4$ & $0.23 \times$ & $0.23 \times$ & $69.6 \pm 0.2$ & $0.10 \times$ & $0.10 \times$ \\
			\citet{dettmers_2019} & DST &  $75.2 \pm 0.1$ & $0.61 \times$ & $0.42 \times$ & $72.9 \pm 0.1$ & $0.50 \times$ & $0.24 \times$ \\					
			\bottomrule
		\end{tabular}
	}	
	\caption{\label{tab:performance}Performance of different methods for a ResNet$50$ trained on ImageNet. Results, except for the LTH experiment and \citet{wimmer_2021b} are derived from \citet{evci_2020}, Figure 2. \citet{wimmer_2021b} uses the pruning method from \citep{evci_2020} combined with a trainable basis. The result for \citet{frankle_2020a} is reported from \citet{evci_2020b} which is LTH with late rewinding.} 
\end{table}
We start with the most important criterion, the \emph{performance} of the methods. With performance we mean, test accuracy (or different metrics for tasks other than image classification) after training. This survey covers methods that reduce training cost. Therefore, performance needs to be brought in the context with the cost for the method. As baseline for comparing different methods, we will use our main cost measure -- the number of trained (or equivalently stored) parameters.

For the performance evaluation, we use LTH with late rewinding. Note, this is \emph{not} a real DRT method since training starts with a subset of $\Theta^\t$ for a small $t > 0$ here. However, literature only reports results for modern networks and big scale tasks like ImageNet for LTH with late rewinding. As already mentioned in Section \ref{sec:lth}, resetting weights to their initialization shows worse results than late rewinding which should be kept in mind.

It was shown and discussed in many works that LTs have better results than PaI for high pruning rates \citep{wang_2020,frankle_2021}. Also, DST improves results compared to PaI \citep{dettmers_2019,evci_2020}. As an example, Table \ref{tab:performance} compares PaI, LTH and DST for a ResNet50 \citep{He2015} on ImageNet \citep{deng_2009}. Table \ref{tab:performance} shows that LTs approximately reach the same performance than DST. Both of them outperform PaI. But, if the training time spent on DST methods is doubled, DST exceeds LT \citep{liu_2021b}, see also discussion in Section \ref{sec:dst}. The doubled training time for DST is a fair comparison, since LTs need at least twice the training time than a standard DST method. On the other hand, the reported LTH result from \citet{evci_2020b} does \emph{not} use the standard way to find LTs \citep{frankle_2020a}, but \emph{gradual magnitude pruning} \citep{gale_2019} to find the sparse subnetwork. Thus, by using the approach to iteratively train-prune-rewind the weights  \citep{frankle_2020a}, LTs performance could be improved further. Moreover, Table \ref{tab:performance} shows that pruning coefficients \wrt adaptive bases \citep{wimmer_2021b} for $K \times K$ filters improves performance of SOTA methods further.

\begin{wrapfigure}{L}{.5\textwidth}
	\centering
	\includegraphics[width=.98\linewidth]{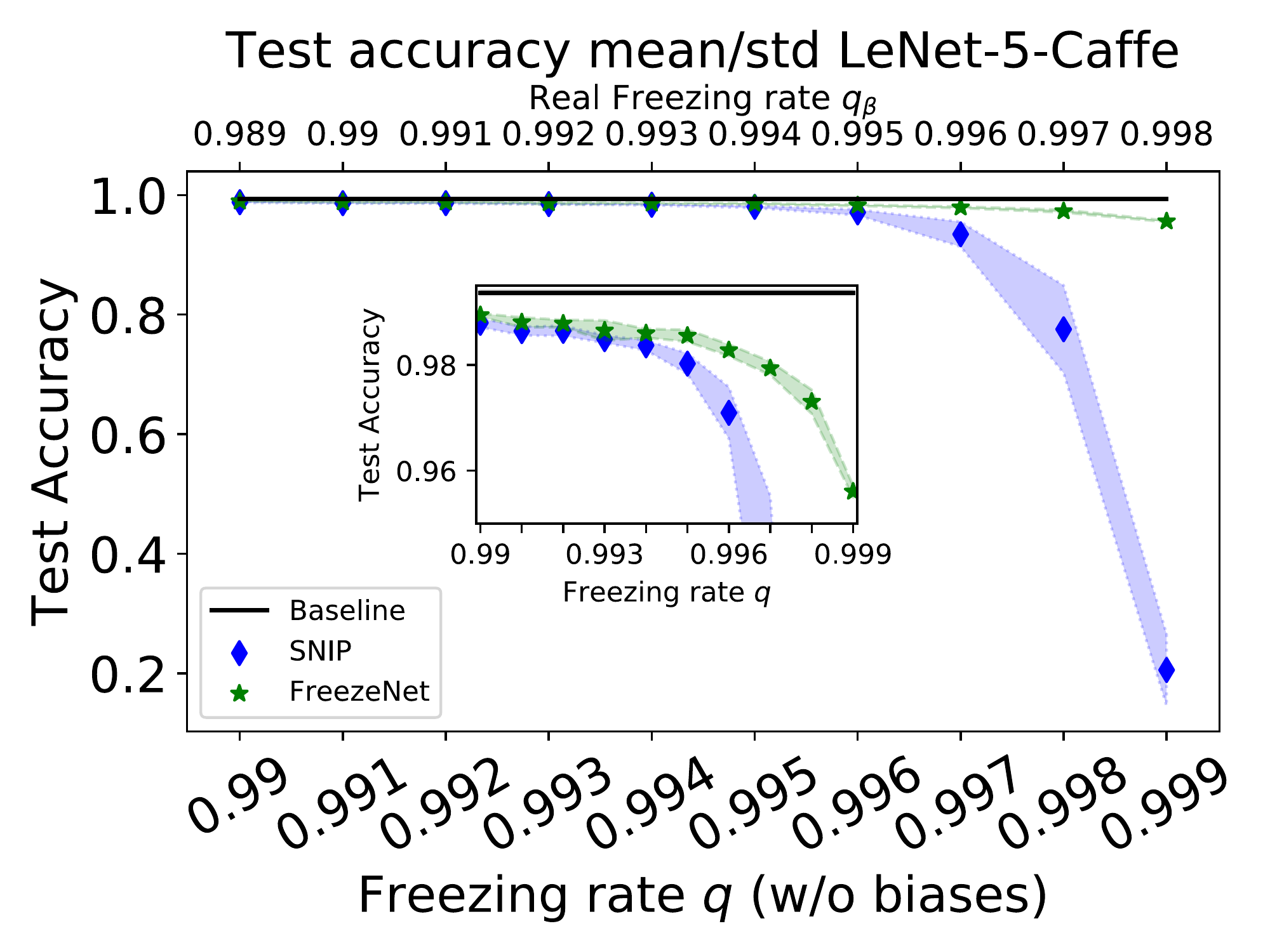}
	\caption{(Figure 1 in \citet{wimmer_2020}) Comparing the PaI method SNIP \citep{lee_2018} and the model with the same trained parameters but the un-trained parameters frozen at their initial value,
		FreezeNet \citep{wimmer_2020} for a LeNet-$5$-Caffe \citep{lecun_1998} on MNIST \citep{lecun_1998}.}
	\label{fig:freeze_wimmer}
\end{wrapfigure}

PaI's inferior performance is not a surprise since PaI methods are clearly the less sophisticated ones. As we will see in the following discussions, this simplicity will reduce costs at other levels, like \emph{training time} or \emph{hyperparameter tuning}. Our first observation therefore is, that pre-training the network to find a well trainable sparse architecture (LT) or adapt the sparse architecture during training (DST) improves results compared to finding a sparse architecture \emph{ad-hoc} at the beginning of training.

Freezing the parameters is compared with pruning them in the PaI setting \citep{wimmer_2020} and the LTH setup \citep{zhou_2019}. Usually, freezing leads to slightly worse results than pruning for a higher number of trained parameters and better results for fewer trained parameters. Figure \ref{fig:freeze_wimmer} shows a comparison between pruning and freezing weights for a low number of trained weights in the PaI setting. \citet{price_2021} show that the same holds true if the random dense layer is replaced by a dense DCT. Using DCTs has the advantage that they are cheap to compute. 

\subsection{Storage cost}
As introduced in Section \ref{subsec:storing}, techniques like the compressed sparse row format \citep{tinney_1967} are used to store sparse/frozen networks after training. By using an initialization derived from a pseudorandom number generator,
\begin{align}
\memory(\texttt{freeze}) & = \memory(\texttt{prune}) + 32\text{bit} \\ & \approx \memory(\texttt{prune}) 
\end{align} 
holds. Consequently, pruning and freezing have the same memory cost in this case. But, if no pseudorandom number generator can be used, pruning has much lower memory requirements than freezing.

Also, Sections \ref{sec:lth} - \ref{sec:freezing} propose methods that do not use a simple pruning/freezing transformation but also an additional linear transformation to embed the small, trainable parameters $\vartheta \in \R^d$ into the bigger space $\R^D$. Examples for this is pruning coupled with an additional basis transformation with shared diagonal blocks \citep{wimmer_2021b} or a random, orthogonal projection \citep{li_2018}. The cost for storing these embeddings is neglectable for \citet{wimmer_2021b}. The same holds for \citet{li_2018} if the random transformation is created with a pseudorandom number generator.

\subsection{Training cost}\label{subsec:training_costs}
In the following, we will compare the \emph{training costs} for different methods which mainly are measured by the \emph{training time} for one sparse model and the number of tunable \emph{hyperparameters} for training which implicitly determines the number of training runs needed overall. On top of that, some methods induce additional gradient computations before or during training which will be discussed in the end. 
\subsubsection{Training time}
We assume that all methods, DST, LTs, PaI and freezing train the {final} sparse model for $T$ iterations. Here, $T$ is the number of training iterations used for the dense network to (i) converge and (ii) have good generalization ability. Consequently, DST, PaI and freezing have approximately the same training time for one model. LTs on the other hand need, as discussed in Section \ref{sec:lth}, at least twice the number of training iterations than a standard training. By using iterative train-prune-reset cycles, the number of training iterations can easily reach $10 - 20\times$ the standard number. Thus, LTs need a massive amount of pre-training for the pruning transformation $\psi^\z$. 

As shown in \citet{liu_2021b} and discussed in Section \ref{sec:dst}, DST massively profit from increasing the training time. 

\subsubsection{Hyperparameters}
All proposed methods induce at least one hyperparameter more than training the corresponding dense model. This hyperparameter is given by the \emph{number of trainable parameters}.\footnote{In pruning literature, many methods determine the number of trainable parameters not directly but indirect by choosing an associated hyperparameter as for example weighting the $\ell_1$ regularization. But, all proposed methods in this work choose the number of trained parameters directly.} 

\paragraph{Determining the number of trained parameters}
PaI, freezing and LT determine \emph{one} global pruning/freezing rate. DST methods on the other hand often need to specify the rate of trained parameters for each layer separately. If the layerwise pruning rates can be adapted dynamically during training \citep{dettmers_2019}, their initial choice has shown to be not too important and can be set constant for all layers. Other works like \citet{mocanu_2018,evci_2020,liu_2021b} heuristically determine layerwise pruning rates by an \emph{Erd\H{o}s-R\' enyi model} \citep{erdoes_1959}.

\paragraph{Additional hyperparameters for PaI}
Besides the pruning rate, PaI methods introduce hyperparameters for computing the pruning transformation $\psi^\z$. For \citet{lee_2018,wang_2020,verdenius_2020,jorge_2020} this is the number of data batches, needed to compute the \emph{pruning score} of each parameter.  The iterative approaches \citet{tanaka_2020,verdenius_2020,jorge_2020} need to determine the number of conducted pruning iterations. However, by using a high number of data batches and many pruning iterations, these parameters are not required to be fine-tuned further \citep{wang_2020,tanaka_2020}.
A special case in this setting is \citet{wimmer_2021}, interpolating between the two pruning methods \citet{lee_2018} and \citet{wang_2020}. Consequently, they need to tune one hyperparameter to balance between these two methods.

\paragraph{Additional hyperparameters for LTs}
As shown in \citet{frankle_2020a}, resetting weights not to their initial value but a value early on in training leads to better results. Consequently, a well performing \emph{rewinding} iteration has to be found. A proper experimental analysis for various datasets and models is given in \citet{frankle_2020a}. Furthermore, if iterative pruning is used, the number of pruned weights in each iteration has to be determined. Usually, $20\%$ of the non-zero weights are removed in each iteration \citep{frankle_2018}.

\paragraph{Additional hyperparameters for freezing}
Freezing is closely related to either PaI \citep{wimmer_2020}, LTs \citep{zhou_2019} or random pruning \citep{rosenfeld_2019}. Since freezing does not introduce additional hyperparameters, the same number of additional hyperparameters as for the corresponding pruning methods are needed. Note, to achieve good results with randomly frozen weights, layerwise freezing rates have to be determined which need to be fine-tuned accordingly.

\paragraph{Additional hyperparameters for DST and costs for updating $\psi^\t$}
Compared to PaI, DST has the advantage that the pruning transformation is not fixed at $\psi^\z$. This leads to better performance after training while using the same number of trained parameters and training iterations. But this also results in costs for determining $\psi^\t$. Besides computing $\psi^\t$, the \emph{update frequency} $F_p$ and the \emph{update pruning rate} $q_l^\t$ have to found. The update pruning rate $q_l^\t$ is defined as the ratio of formerly non-zero parameters which are newly pruned in layer $l$ if $\psi^\t$ is updated in iteration $t$. Note, the update pruning rate might vary between different update iterations of the transformation and also between layers. Also, the \emph{regrowing} rate needs to be determined since it can be different from the update pruning rate \citep{dettmers_2019}. 

Finally, there are the actual costs for updating $\psi^\t$ itself. In all considered methods in Section \ref{sec:dst}, weights are dried according to having the smallest magnitude -- magnitude based pruning. For this, the $d$ un-pruned coefficients have to be sorted. After setting some weights to zero, the same number of weights has to be flagged as trainable again. The costs for regrowing weights can almost be for free by using random regrowing of weights \citep{mocanu_2018,mostafa_2019}. On the other hand, \citet{evci_2020} require the gradient of the \emph{whole} network at the update iterations $t \in \{F_p, 2F_p,\ldots\}$ to determine $\psi^\t$. \citet{dettmers_2019} even need these gradients in \emph{each} training iteration in order to update a momentum parameter which also requires an additional weighting hyperparameter. A temperature parameter is needed in \citet{bellec_2018} to model a \emph{random walk} in the parameter space to explore new weights.

\subsubsection{Gradient computations and backward pass.}\label{subsubsec:gradients}
All proposed methods update only sparse parts of the weights via backpropagation. But, some of them additionally require the computation of the dense gradient for pre-training, determining $\psi^\z$ or to update $\psi^\t$.

First of all, LTs need to \emph{train} the dense network in order to find $\psi^\z$. This can limit the size of the biggest underlying dense network that can be used. For iterative LTH, networks with sparsity $p_k = 1 - 0.8^k$ have to be trained additionally for all $k$ with $p_k > p$.

PaI methods need the computation of a dense gradient before training \citep{lee_2018,tanaka_2020,verdenius_2020,jorge_2020}. Even a Hessian-vector product has to be calculated for the method \citet{wang_2020}. But after having found $\psi^\z$, PaI methods are trained with a fixed, sparse architecture without the need to compute a dense gradient anymore.

DST methods might not need a dense gradient at all, if regrown weights are found by a random selection \citep{mocanu_2018,bellec_2018,mostafa_2019}. As mentioned above, \citet{evci_2020} compute the dense gradient at each update step of the pruning transformation $t \in \{F_p, 2  F_p , \ldots\}$. The most extreme case is given for \citet{dettmers_2019} which need to compute the dense gradient in each iteration. 

Finally there is a difference between gradient computations for pruned and frozen models. Pruned networks compute gradients of activation maps with sparse weight tensors. Frozen networks on the other hand compute gradients of activation maps with the help of the frozen weights, \ie require dense computations, see Figure \ref{fig:freezing_overview} right side. However, frozen weights do not need to be updated. Therefore, it is enough to compute only a sparse part of the weights' gradients. In summary, freezing also reduces computations in the backward pass, but not as much as pruning does.

\subsection{Forward Pass.}
In Section \ref{subsubsec:gradients}, gradient computations and the backward pass for the different methods are discussed. Gradients only have to be computed during training whereas the forward pass is needed for both, training and inference. Therefore, we discuss the forward pass in a standalone Section. Still, this discussion should also be seen as a part of the training costs, Section \ref{subsec:training_costs}.

During and after the sparse training, LTH, DST and PaI all have the same cost for inference. This statement is not completely true, since different methods might create varying sparsity distributions for the same global pruning rate. This can result in different FLOP costs for inferring the sparse networks, see for example Table \ref{tab:performance}. However, analyzing sparsity distributions obtained by different pruning methods is beyond the scope of this work. 


If parts of a network are frozen during training, \emph{all} parameters participate in the forward pass. Of course, this also holds true for inference time. Consequently, the computational cost for inference can not be reduced and is equal to the densely trained network.

\subsection{Summary}
In summary, we see that {freezing} parameters instead of pruning them leads to the same memory requirements and better results for training a low number of parameters. However, if more parameters are trained, pruning leads to equal or better performance. For pruning, the sparsity of the parameters can reduce the number of required computations for evaluating the network if the used soft- and hardware supports sparse computations. Thus, freezing seems to be a good option if the model size is a bottleneck, \ie only a small part of the network is trained, whereas pruning is preferably used otherwise. Furthermore, the computational overhead induced by freezing layers can be reduced by using dense DCTs instead of dense frozen layers.

For pruning, there is a trade-off between simplicity of the method and performance after training. As shown, DST and LTs have approximately the same generalization ability and are superior to PaI for the same number of non-zero parameters. However, PaI guarantees a fixed sparse architecture throughout training, determined by only a few dense gradient computations. Furthermore, PaI does not require extensive and costly hyperparameter tuning. Finding LTs requires expensive pre-training of the dense network. Also, LTs at initialization often show unsatisfactory results for modern network architectures and big scale datasets. Resetting weights to an early phase in training is required to achieve good performance. As shown, DST methods need more hyperparameters than PaI and LTH. Further, DST methods might need to regularly compute the full gradient of the network and increase the training time in order to reach their best performance.

As mentioned, presented results for LTs use a small pre-training step for the sparse initialization. Therefore, we conclude that DST methods achieve the best overall performance for completely sparse training. Moreover, using adaptive bases for the $K \times K$ filters of convolutional layers further improves the proposed pruning and freezing approaches.

\section{Conclusions}\label{sec:conclusions}
In this work we have introduced a general framework to describe the training of DNNs with reduced dimensionality (DRT). The proposed methods are dominated by pruning neural networks, but we also discussed freezing parts of the network at its random initialization. The methods are categorized in pruning at initialization (PaI), lottery tickets (LTs), dynamic sparse training (DST) and freezing. SOTA methods for each criterion are presented. Furthermore, the different approaches are compared afterwards. 

We first discussed that pruning leads to better results than freezing if more parameters are optimized whereas for a low number of trained weights, freezing performs better than pruning. Furthermore, LTs and DST perform better than PaI, while PaI contains the easiest and least expensive methods. For LTs, many trainings, including training the dense model, are needed to find the sparse architecture. Also, LTs achieve their best results if weights are reset to an early phase in training which does not yield a complete sparse training. DST methods usually need more hyperparameters to be tuned than LTs and PaI. All proposed DRT methods can be further enhanced by representing convolutional filters with an adaptive representation instead of the standard, spatial one. 

Altogether, finding the best DRT method for a specific setup is a trade-off between available training time and performance of the final network. The training time is mostly influenced by the number of trainings required to find the sparse architecture, ranging from 0 (PaI, DST and freezing) to more than $20 \times$ (LTs), and the number of individual trainings required to tune hyperparameters. Also, the need to compute dense gradients (for some DST and PaI methods) or to pre-train the dense network (LTs) has to be considered if a DRT method is chosen. As discussed, frozen models have similar computational costs for inference as densely trained networks. Consequently, if the sparsity together with the used soft- and hardware can actually reduce computations, pruning should be preferred to freezing, or frozen parameters should be replaced by information preserving transformations which are cheap to compute, as for example the DCT.

\bibliography{ref_survey}

\begin{thebibliography}{195}
\expandafter\ifx\csname natexlab\endcsname\relax\def\natexlab#1{#1}\fi
\providecommand{\url}[1]{\texttt{#1}}
\providecommand{\href}[2]{#2}
\providecommand{\path}[1]{#1}
\providecommand{\DOIprefix}{doi:}
\providecommand{\ArXivprefix}{arXiv:}
\providecommand{\URLprefix}{URL: }
\providecommand{\Pubmedprefix}{pmid:}
\providecommand{\doi}[1]{\href{http://dx.doi.org/#1}{\path{#1}}}
\providecommand{\Pubmed}[1]{\href{pmid:#1}{\path{#1}}}
\providecommand{\bibinfo}[2]{#2}
\ifx\xfnm\relax \def\xfnm[#1]{\unskip,\space#1}\fi
\bibitem[{Pham et~al.(2021)Pham, Dai, Xie, Luong, and Le}]{pham_2021}
\bibinfo{author}{H.~Pham}, \bibinfo{author}{Z.~Dai}, \bibinfo{author}{Q.~Xie},
  \bibinfo{author}{M.-T. Luong}, \bibinfo{author}{Q.~V. Le},
\newblock \bibinfo{title}{Meta pseudo labels},
\newblock in: \bibinfo{booktitle}{Proceedings of the IEEE Conference on
  Computer Vision and Pattern Recognition}, \bibinfo{year}{2021}.
\bibitem[{Park et~al.(2020)Park, Zhang, Chiu, Chen, Li, Chan, Le, and
  Wu}]{park_2020}
\bibinfo{author}{D.~S. Park}, \bibinfo{author}{Y.~Zhang},
  \bibinfo{author}{C.~Chiu}, \bibinfo{author}{Y.~Chen},
  \bibinfo{author}{B.~Li}, \bibinfo{author}{W.~Chan}, \bibinfo{author}{Q.~V.
  Le}, \bibinfo{author}{Y.~Wu},
\newblock \bibinfo{title}{Specaugment on large scale datasets},
\newblock in: \bibinfo{booktitle}{{IEEE} International Conference on Acoustics,
  Speech and Signal Processing}, \bibinfo{year}{2020}.
\bibitem[{Wang et~al.(2021)Wang, Bochkovskiy, and Liao}]{wang_2021}
\bibinfo{author}{C.-Y. Wang}, \bibinfo{author}{A.~Bochkovskiy},
  \bibinfo{author}{H.-Y.~M. Liao},
\newblock \bibinfo{title}{Scaled-yolov4: Scaling cross stage partial network},
\newblock in: \bibinfo{booktitle}{Proceedings of the IEEE/CVF Conference on
  Computer Vision and Pattern Recognition}, \bibinfo{year}{2021}.
\bibitem[{Yang et~al.(2019)Yang, Dai, Yang, Carbonell, Salakhutdinov, and
  Le}]{yang_2019}
\bibinfo{author}{Z.~Yang}, \bibinfo{author}{Z.~Dai}, \bibinfo{author}{Y.~Yang},
  \bibinfo{author}{J.~Carbonell}, \bibinfo{author}{R.~R. Salakhutdinov},
  \bibinfo{author}{Q.~V. Le},
\newblock \bibinfo{title}{Xlnet: Generalized autoregressive pretraining for
  language understanding},
\newblock in: \bibinfo{booktitle}{Advances in Neural Information Processing
  Systems 32}, \bibinfo{year}{2019}.
\bibitem[{Schwartz et~al.(2020)Schwartz, Dodge, Smith, and
  Etzioni}]{schwartz_2019}
\bibinfo{author}{R.~Schwartz}, \bibinfo{author}{J.~Dodge},
  \bibinfo{author}{N.~A. Smith}, \bibinfo{author}{O.~Etzioni},
\newblock \bibinfo{title}{Green {AI}},
\newblock \bibinfo{journal}{Communications of the ACM} \bibinfo{volume}{63}
  (\bibinfo{year}{2020}) \bibinfo{pages}{54--63}.
\bibitem[{Peters et~al.(2018)Peters, Neumann, Iyyer, Gardner, Clark, Lee, and
  Zettlemoyer}]{peters_2018}
\bibinfo{author}{M.~E. Peters}, \bibinfo{author}{M.~Neumann},
  \bibinfo{author}{M.~Iyyer}, \bibinfo{author}{M.~Gardner},
  \bibinfo{author}{C.~Clark}, \bibinfo{author}{K.~Lee},
  \bibinfo{author}{L.~Zettlemoyer},
\newblock \bibinfo{title}{Deep contextualized word representations},
\newblock in: \bibinfo{booktitle}{Proceedings of the 2018 Conference of the
  North {A}merican Chapter of the Association for Computational Linguistics:
  Human Language Technologies}, \bibinfo{year}{2018}.
\bibitem[{Mahajan et~al.(2018)Mahajan, Girshick, Ramanathan, He, Paluri, Li,
  Bharambe, and van~der Maaten}]{mahajan_2018}
\bibinfo{author}{D.~Mahajan}, \bibinfo{author}{R.~Girshick},
  \bibinfo{author}{V.~Ramanathan}, \bibinfo{author}{K.~He},
  \bibinfo{author}{M.~Paluri}, \bibinfo{author}{Y.~Li},
  \bibinfo{author}{A.~Bharambe}, \bibinfo{author}{L.~van~der Maaten},
\newblock \bibinfo{title}{Exploring the limits of weakly supervised
  pretraining},
\newblock in: \bibinfo{booktitle}{Proceedings of the European Conference on
  Computer Vision}, \bibinfo{year}{2018}.
\bibitem[{Kolesnikov et~al.(2020)Kolesnikov, Beyer, Zhai, Puigcerver, Yung,
  Gelly, and Houlsby}]{kolesnikov_2020}
\bibinfo{author}{A.~Kolesnikov}, \bibinfo{author}{L.~Beyer},
  \bibinfo{author}{X.~Zhai}, \bibinfo{author}{J.~Puigcerver},
  \bibinfo{author}{J.~Yung}, \bibinfo{author}{S.~Gelly},
  \bibinfo{author}{N.~Houlsby},
\newblock \bibinfo{title}{Big transfer (bit): General visual representation
  learning},
\newblock in: \bibinfo{booktitle}{Proceedings of the European Conference on
  Computer Vision}, \bibinfo{year}{2020}.
\bibitem[{Dosovitskiy et~al.(2021)Dosovitskiy, Beyer, Kolesnikov, Weissenborn,
  Zhai, Unterthiner, Dehghani, Minderer, Heigold, Gelly, Uszkoreit, and
  Houlsby}]{dosovitskiy_2021}
\bibinfo{author}{A.~Dosovitskiy}, \bibinfo{author}{L.~Beyer},
  \bibinfo{author}{A.~Kolesnikov}, \bibinfo{author}{D.~Weissenborn},
  \bibinfo{author}{X.~Zhai}, \bibinfo{author}{T.~Unterthiner},
  \bibinfo{author}{M.~Dehghani}, \bibinfo{author}{M.~Minderer},
  \bibinfo{author}{G.~Heigold}, \bibinfo{author}{S.~Gelly},
  \bibinfo{author}{J.~Uszkoreit}, \bibinfo{author}{N.~Houlsby},
\newblock \bibinfo{title}{An image is worth 16x16 words: Transformers for image
  recognition at scale},
\newblock in: \bibinfo{booktitle}{9th International Conference on Learning
  Representations}, \bibinfo{year}{2021}.
\bibitem[{Strubell et~al.(2019)Strubell, Ganesh, and McCallum}]{strubell_2019}
\bibinfo{author}{E.~Strubell}, \bibinfo{author}{A.~Ganesh},
  \bibinfo{author}{A.~McCallum},
\newblock \bibinfo{title}{Energy and policy considerations for deep learning in
  {NLP}},
\newblock in: \bibinfo{booktitle}{Proceedings of the 57th Conference of the
  Association for Computational Linguistics}, \bibinfo{year}{2019}.
\bibitem[{Amodei et~al.(line)Amodei, Hernandez, Sastry, Clark, Brockman, and
  Sutskever}]{amodei_2018}
\bibinfo{author}{D.~Amodei}, \bibinfo{author}{D.~Hernandez},
  \bibinfo{author}{G.~Sastry}, \bibinfo{author}{J.~Clark},
  \bibinfo{author}{G.~Brockman}, \bibinfo{author}{I.~Sutskever},
  \bibinfo{title}{Ai and compute}, \bibinfo{howpublished}{OpenAI Blog},
  \bibinfo{year}{2018 [Online]}. \URLprefix
  \url{https://openai.com/blog/ai-and-compute/}, \bibinfo{note}{last accessed:
  05/16/2022}.
\bibitem[{Gustafson(2011)}]{gustafson_2011}
\bibinfo{author}{J.~L. Gustafson},
\newblock \bibinfo{title}{Moore's law},
\newblock in: \bibinfo{booktitle}{Encyclopedia of Parallel Computing},
  \bibinfo{year}{2011}, pp. \bibinfo{pages}{1177--1184}.
\bibitem[{Strubell et~al.(2020)Strubell, Ganesh, and McCallum}]{strubell_2020}
\bibinfo{author}{E.~Strubell}, \bibinfo{author}{A.~Ganesh},
  \bibinfo{author}{A.~McCallum},
\newblock \bibinfo{title}{Energy and policy considerations for modern deep
  learning research},
\newblock in: \bibinfo{booktitle}{Proceedings of the AAAI Conference on
  Artificial Intelligence}, \bibinfo{year}{2020}.
\bibitem[{Han et~al.(2015)Han, Pool, Tran, and Dally}]{han_2015}
\bibinfo{author}{S.~Han}, \bibinfo{author}{J.~Pool}, \bibinfo{author}{J.~Tran},
  \bibinfo{author}{W.~Dally},
\newblock \bibinfo{title}{Learning both weights and connections for efficient
  neural network},
\newblock in: \bibinfo{booktitle}{Advances in Neural Information Processing
  Systems 28}, \bibinfo{year}{2015}.
\bibitem[{Courbariaux et~al.(2015)Courbariaux, Bengio, and
  David}]{courbariaux_2015}
\bibinfo{author}{M.~Courbariaux}, \bibinfo{author}{Y.~Bengio},
  \bibinfo{author}{J.-P. David},
\newblock \bibinfo{title}{Binaryconnect: Training deep neural networks with
  binary weights during propagations},
\newblock in: \bibinfo{booktitle}{Advances in Neural Information Processing
  Systems 28}, \bibinfo{year}{2015}.
\bibitem[{Wu et~al.(2016)Wu, Leng, Wang, Hu, and Cheng}]{wu_2016}
\bibinfo{author}{J.~Wu}, \bibinfo{author}{C.~Leng}, \bibinfo{author}{Y.~Wang},
  \bibinfo{author}{Q.~Hu}, \bibinfo{author}{J.~Cheng},
\newblock \bibinfo{title}{Quantized convolutional neural networks for mobile
  devices},
\newblock in: \bibinfo{booktitle}{Proceedings of the IEEE Conference on
  Computer Vision and Pattern Recognition}, \bibinfo{year}{2016}.
\bibitem[{Zhou et~al.(2017)Zhou, Yao, Guo, Xu, and Chen}]{zhou_2017}
\bibinfo{author}{A.~Zhou}, \bibinfo{author}{A.~Yao}, \bibinfo{author}{Y.~Guo},
  \bibinfo{author}{L.~Xu}, \bibinfo{author}{Y.~Chen},
\newblock \bibinfo{title}{Incremental network quantization: Towards lossless
  cnns with low-precision weights},
\newblock in: \bibinfo{booktitle}{5th International Conference on Learning
  Representations}, \bibinfo{year}{2017}.
\bibitem[{Zhang et~al.(2018)Zhang, Yang, Ye, and Hua}]{zhang_2018}
\bibinfo{author}{D.~Zhang}, \bibinfo{author}{J.~Yang}, \bibinfo{author}{D.~Ye},
  \bibinfo{author}{G.~Hua},
\newblock \bibinfo{title}{Lq-nets: Learned quantization for highly accurate and
  compact deep neural networks},
\newblock in: \bibinfo{booktitle}{Proceedings of the European Conference on
  Computer Vision}, \bibinfo{year}{2018}.
\bibitem[{Jacob et~al.(2018)Jacob, Kligys, Chen, Zhu, Tang, Howard, Adam, and
  Kalenichenko}]{jacob_2018}
\bibinfo{author}{B.~Jacob}, \bibinfo{author}{S.~Kligys},
  \bibinfo{author}{B.~Chen}, \bibinfo{author}{M.~Zhu},
  \bibinfo{author}{M.~Tang}, \bibinfo{author}{A.~Howard},
  \bibinfo{author}{H.~Adam}, \bibinfo{author}{D.~Kalenichenko},
\newblock \bibinfo{title}{Quantization and training of neural networks for
  efficient integer-arithmetic-only inference},
\newblock in: \bibinfo{booktitle}{Proceedings of the IEEE Conference on
  Computer Vision and Pattern Recognition}, \bibinfo{year}{2018}.
\bibitem[{Li et~al.(2019)Li, Wang, Liang, Qin, Yan, and Fan}]{li_2019b}
\bibinfo{author}{R.~Li}, \bibinfo{author}{Y.~Wang}, \bibinfo{author}{F.~Liang},
  \bibinfo{author}{H.~Qin}, \bibinfo{author}{J.~Yan}, \bibinfo{author}{R.~Fan},
\newblock \bibinfo{title}{Fully quantized network for object detection},
\newblock in: \bibinfo{booktitle}{Proceedings of the IEEE/CVF Conference on
  Computer Vision and Pattern Recognition}, \bibinfo{year}{2019}.
\bibitem[{Nowlan and Hinton(1992)}]{nowlan_1992}
\bibinfo{author}{S.~J. Nowlan}, \bibinfo{author}{G.~E. Hinton},
\newblock \bibinfo{title}{Simplifying neural networks by soft weight-sharing},
\newblock \bibinfo{journal}{Neural Computation} \bibinfo{volume}{4}
  (\bibinfo{year}{1992}) \bibinfo{pages}{473--493}.
\bibitem[{Chen et~al.(2015)Chen, Wilson, Tyree, Weinberger, and
  Chen}]{chen_2015}
\bibinfo{author}{W.~Chen}, \bibinfo{author}{J.~Wilson},
  \bibinfo{author}{S.~Tyree}, \bibinfo{author}{K.~Weinberger},
  \bibinfo{author}{Y.~Chen},
\newblock \bibinfo{title}{Compressing neural networks with the hashing trick},
\newblock in: \bibinfo{booktitle}{Proceedings of the 32nd International
  Conference on Machine Learning}, \bibinfo{year}{2015}.
\bibitem[{Ullrich et~al.(2017)Ullrich, Meeds, and Welling}]{ullrich_2017}
\bibinfo{author}{K.~Ullrich}, \bibinfo{author}{E.~Meeds},
  \bibinfo{author}{M.~Welling},
\newblock \bibinfo{title}{Soft weight-sharing for neural network compression},
\newblock in: \bibinfo{booktitle}{5th International Conference on Learning
  Representations}, \bibinfo{year}{2017}.
\bibitem[{Xue et~al.(2013)Xue, Li, and Gong}]{xue_2013}
\bibinfo{author}{J.~Xue}, \bibinfo{author}{J.~Li}, \bibinfo{author}{Y.~Gong},
\newblock \bibinfo{title}{Restructuring of deep neural network acoustic models
  with singular value decomposition},
\newblock in: \bibinfo{booktitle}{Interspeech}, \bibinfo{year}{2013}.
\bibitem[{Lebedev et~al.(2015)Lebedev, Ganin, Rakhuba, Oseledets, and
  Lempitsky}]{lebedev_2015}
\bibinfo{author}{V.~Lebedev}, \bibinfo{author}{Y.~Ganin},
  \bibinfo{author}{M.~Rakhuba}, \bibinfo{author}{I.~V. Oseledets},
  \bibinfo{author}{V.~S. Lempitsky},
\newblock \bibinfo{title}{Speeding-up convolutional neural networks using
  fine-tuned cp-decomposition},
\newblock in: \bibinfo{booktitle}{3rd International Conference on Learning
  Representations}, \bibinfo{year}{2015}.
\bibitem[{Novikov et~al.(2015)Novikov, Podoprikhin, Osokin, and
  Vetrov}]{novikov_2015}
\bibinfo{author}{A.~Novikov}, \bibinfo{author}{D.~Podoprikhin},
  \bibinfo{author}{A.~Osokin}, \bibinfo{author}{D.~P. Vetrov},
\newblock \bibinfo{title}{Tensorizing neural networks},
\newblock in: \bibinfo{booktitle}{Advances in Neural Information Processing
  Systems 28}, \bibinfo{year}{2015}.
\bibitem[{Sainath et~al.(2013)Sainath, Kingsbury, Sindhwani, Arisoy, and
  Ramabhadran}]{sainath_2013}
\bibinfo{author}{T.~Sainath}, \bibinfo{author}{B.~Kingsbury},
  \bibinfo{author}{V.~Sindhwani}, \bibinfo{author}{E.~Arisoy},
  \bibinfo{author}{B.~Ramabhadran},
\newblock \bibinfo{title}{Low-rank matrix factorization for deep neural network
  training with high-dimensional output targets},
\newblock in: \bibinfo{booktitle}{IEEE International Conference on Acoustics,
  Speech and Signal Processing}, \bibinfo{year}{2013}.
\bibitem[{Denton et~al.(2014)Denton, Zaremba, Bruna, LeCun, and
  Fergus}]{denton_2014}
\bibinfo{author}{E.~Denton}, \bibinfo{author}{W.~Zaremba},
  \bibinfo{author}{J.~Bruna}, \bibinfo{author}{Y.~LeCun},
  \bibinfo{author}{R.~Fergus},
\newblock \bibinfo{title}{Exploiting linear structure within convolutional
  networks for efficient evaluation},
\newblock in: \bibinfo{booktitle}{Advances in Neural Information Processing
  Systems 27}, \bibinfo{year}{2014}.
\bibitem[{Liu et~al.(2015)Liu, Wang, Foroosh, Tappen, and Pensky}]{liu_2015}
\bibinfo{author}{B.~Liu}, \bibinfo{author}{M.~Wang},
  \bibinfo{author}{H.~Foroosh}, \bibinfo{author}{M.~Tappen},
  \bibinfo{author}{M.~Pensky},
\newblock \bibinfo{title}{Sparse convolutional neural networks.},
\newblock in: \bibinfo{booktitle}{Proceedings of the IEEE Conference on
  Computer Vision and Pattern Recognition}, \bibinfo{year}{2015}.
\bibitem[{Janowsky(1989)}]{janowsky_1989}
\bibinfo{author}{S.~A. Janowsky},
\newblock \bibinfo{title}{Pruning versus clipping in neural networks},
\newblock \bibinfo{journal}{Physical Review A} \bibinfo{volume}{39}
  (\bibinfo{year}{1989}) \bibinfo{pages}{6600--6603}.
\bibitem[{Mozer and Smolensky(1989)}]{mozer_1989}
\bibinfo{author}{M.~C. Mozer}, \bibinfo{author}{P.~Smolensky},
\newblock \bibinfo{title}{Skeletonization: A technique for trimming the fat
  from a network via relevance assessment},
\newblock in: \bibinfo{booktitle}{Advances in Neural Information Processing
  Systems 1}, \bibinfo{year}{1989}.
\bibitem[{{Karnin}(1990)}]{karnin_1990}
\bibinfo{author}{E.~D. {Karnin}},
\newblock \bibinfo{title}{A simple procedure for pruning back-propagation
  trained neural networks},
\newblock \bibinfo{journal}{IEEE Transactions on Neural Networks}
  \bibinfo{volume}{1} (\bibinfo{year}{1990}) \bibinfo{pages}{239--242}.
\bibitem[{LeCun et~al.(1990)LeCun, Denker, and Solla}]{lecun_1990}
\bibinfo{author}{Y.~LeCun}, \bibinfo{author}{J.~S. Denker},
  \bibinfo{author}{S.~A. Solla},
\newblock \bibinfo{title}{Optimal brain damage},
\newblock in: \bibinfo{booktitle}{Advances in Neural Information Processing
  Systems 2}, \bibinfo{year}{1990}.
\bibitem[{Guo et~al.(2016)Guo, Yao, and Chen}]{guo_2016}
\bibinfo{author}{Y.~Guo}, \bibinfo{author}{A.~Yao}, \bibinfo{author}{Y.~Chen},
\newblock \bibinfo{title}{Dynamic network surgery for efficient dnns},
\newblock in: \bibinfo{booktitle}{Advances in Neural Information Processing
  Systems 29}, \bibinfo{year}{2016}.
\bibitem[{Qian and Klabjan(2021)}]{qian_2021}
\bibinfo{author}{X.~Qian}, \bibinfo{author}{D.~Klabjan},
\newblock \bibinfo{title}{A probabilistic approach to neural network pruning},
\newblock in: \bibinfo{booktitle}{Proceedings of the 38th International
  Conference on Machine Learning}, \bibinfo{year}{2021}.
\bibitem[{Blalock et~al.(2020)Blalock, Ortiz, Frankle, and
  Guttag}]{blalock_2020}
\bibinfo{author}{D.~W. Blalock}, \bibinfo{author}{J.~J.~G. Ortiz},
  \bibinfo{author}{J.~Frankle}, \bibinfo{author}{J.~V. Guttag},
\newblock \bibinfo{title}{What is the state of neural network pruning?},
\newblock in: \bibinfo{booktitle}{Proceedings of Machine Learning and Systems
  2}, \bibinfo{year}{2020}.
\bibitem[{Arora et~al.(2018)Arora, Ge, Neyshabur, and Zhang}]{arora_2018}
\bibinfo{author}{S.~Arora}, \bibinfo{author}{R.~Ge},
  \bibinfo{author}{B.~Neyshabur}, \bibinfo{author}{Y.~Zhang},
\newblock \bibinfo{title}{Stronger generalization bounds for deep nets via a
  compression approach},
\newblock in: \bibinfo{booktitle}{Proceedings of the 35th International
  Conference on Machine Learning}, \bibinfo{year}{2018}.
\bibitem[{Bartoldson et~al.(2020)Bartoldson, Morcos, Barbu, and
  Erlebacher}]{bartoldson_2019}
\bibinfo{author}{B.~Bartoldson}, \bibinfo{author}{A.~Morcos},
  \bibinfo{author}{A.~Barbu}, \bibinfo{author}{G.~Erlebacher},
\newblock \bibinfo{title}{The generalization-stability tradeoff in neural
  network pruning},
\newblock in: \bibinfo{booktitle}{Advances in Neural Information Processing
  Systems 33}, \bibinfo{year}{2020}.
\bibitem[{Barsbey et~al.(2021)Barsbey, Sefidgaran, Erdogdu, Richard, and
  Simsekli}]{barsbey_2021}
\bibinfo{author}{M.~Barsbey}, \bibinfo{author}{M.~Sefidgaran},
  \bibinfo{author}{M.~A. Erdogdu}, \bibinfo{author}{G.~Richard},
  \bibinfo{author}{U.~Simsekli},
\newblock \bibinfo{title}{Heavy tails in {SGD} and compressibility of
  overparametrized neural networks},
\newblock in: \bibinfo{booktitle}{Advances in Neural Information Processing
  Systems 34}, \bibinfo{year}{2021}.
\bibitem[{Anwar et~al.(2017)Anwar, Hwang, and Sung}]{anwar_2017}
\bibinfo{author}{S.~Anwar}, \bibinfo{author}{K.~Hwang},
  \bibinfo{author}{W.~Sung},
\newblock \bibinfo{title}{Structured pruning of deep convolutional neural
  networks},
\newblock \bibinfo{journal}{ACM Journal on Emerging Technologies in Computing
  Systems} \bibinfo{volume}{13} (\bibinfo{year}{2017}) \bibinfo{pages}{1--18}.
\bibitem[{Chen et~al.(2020)Chen, Chen, and Pan}]{chen_2020}
\bibinfo{author}{J.~Chen}, \bibinfo{author}{S.~Chen}, \bibinfo{author}{S.~J.
  Pan},
\newblock \bibinfo{title}{Storage efficient and dynamic flexible runtime
  channel pruning via deep reinforcement learning},
\newblock in: \bibinfo{booktitle}{Advances in Neural Information Processing
  Systems 33}, \bibinfo{year}{2020}.
\bibitem[{Huang and Wang(2018)}]{huang_2018}
\bibinfo{author}{Z.~Huang}, \bibinfo{author}{N.~Wang},
\newblock \bibinfo{title}{Data-driven sparse structure selection for deep
  neural networks},
\newblock in: \bibinfo{booktitle}{Proceedings of the European Conference on
  Computer Vision}, \bibinfo{year}{2018}.
\bibitem[{Zhuang et~al.(2018)Zhuang, Tan, Zhuang, Liu, Guo, Wu, Huang, and
  Zhu}]{zhuang_2018}
\bibinfo{author}{Z.~Zhuang}, \bibinfo{author}{M.~Tan},
  \bibinfo{author}{B.~Zhuang}, \bibinfo{author}{J.~Liu},
  \bibinfo{author}{Y.~Guo}, \bibinfo{author}{Q.~Wu},
  \bibinfo{author}{J.~Huang}, \bibinfo{author}{J.~Zhu},
\newblock \bibinfo{title}{Discrimination-aware channel pruning for deep neural
  networks},
\newblock in: \bibinfo{booktitle}{Advances in Neural Information Processing
  Systems 31}, \bibinfo{year}{2018}.
\bibitem[{Zhuang et~al.(2020)Zhuang, Zhang, Huang, Zeng, Shuang, and
  Li}]{zhuang_2020}
\bibinfo{author}{T.~Zhuang}, \bibinfo{author}{Z.~Zhang},
  \bibinfo{author}{Y.~Huang}, \bibinfo{author}{X.~Zeng},
  \bibinfo{author}{K.~Shuang}, \bibinfo{author}{X.~Li},
\newblock \bibinfo{title}{Neuron-level structured pruning using polarization
  regularizer},
\newblock in: \bibinfo{booktitle}{Advances in Neural Information Processing
  Systems 33}, \bibinfo{year}{2020}.
\bibitem[{Joo et~al.(2021)Joo, Yi, Baek, and Kim}]{joo_2021}
\bibinfo{author}{D.~Joo}, \bibinfo{author}{E.~Yi}, \bibinfo{author}{S.~Baek},
  \bibinfo{author}{J.~Kim},
\newblock \bibinfo{title}{Linearly replaceable filters for deep network channel
  pruning},
\newblock in: \bibinfo{booktitle}{Proceedings of the AAAI Conference on
  Artificial Intelligence}, \bibinfo{year}{2021}.
\bibitem[{Wang et~al.(2020)Wang, Zhang, Hu, Zhang, and Su}]{wang_2020d}
\bibinfo{author}{Y.~Wang}, \bibinfo{author}{X.~Zhang}, \bibinfo{author}{X.~Hu},
  \bibinfo{author}{B.~Zhang}, \bibinfo{author}{H.~Su},
\newblock \bibinfo{title}{Dynamic network pruning with interpretable layerwise
  channel selection},
\newblock in: \bibinfo{booktitle}{Proceedings of the AAAI Conference on
  Artificial Intelligence}, \bibinfo{year}{2020}.
\bibitem[{Frankle and Carbin(2018)}]{frankle_2018}
\bibinfo{author}{J.~Frankle}, \bibinfo{author}{M.~Carbin},
\newblock \bibinfo{title}{The lottery ticket hypothesis: Finding sparse,
  trainable neural networks},
\newblock in: \bibinfo{booktitle}{6th International Conference on Learning
  Representations}, \bibinfo{year}{2018}.
\bibitem[{Lee et~al.(2019)Lee, Ajanthan, and Torr}]{lee_2018}
\bibinfo{author}{N.~Lee}, \bibinfo{author}{T.~Ajanthan}, \bibinfo{author}{P.~H.
  Torr},
\newblock \bibinfo{title}{{SNIP}: Single-shot network pruning based on
  connection sensitivity},
\newblock in: \bibinfo{booktitle}{7th International Conference on Learning
  Representations}, \bibinfo{year}{2019}.
\bibitem[{Wang et~al.(2020)Wang, Zhang, and Grosse}]{wang_2020}
\bibinfo{author}{C.~Wang}, \bibinfo{author}{G.~Zhang},
  \bibinfo{author}{R.~Grosse},
\newblock \bibinfo{title}{Picking winning tickets before training by preserving
  gradient flow},
\newblock in: \bibinfo{booktitle}{8th International Conference on Learning
  Representations}, \bibinfo{year}{2020}.
\bibitem[{Tanaka et~al.(2020)Tanaka, Kunin, Yamins, and Ganguli}]{tanaka_2020}
\bibinfo{author}{H.~Tanaka}, \bibinfo{author}{D.~Kunin}, \bibinfo{author}{D.~L.
  Yamins}, \bibinfo{author}{S.~Ganguli},
\newblock \bibinfo{title}{Pruning neural networks without any data by
  iteratively conserving synaptic flow},
\newblock in: \bibinfo{booktitle}{Advances in Neural Information Processing
  Systems 33}, \bibinfo{year}{2020}.
\bibitem[{Li et~al.(2017)Li, Kadav, Durdanovic, Samet, and Graf}]{li_2016}
\bibinfo{author}{H.~Li}, \bibinfo{author}{A.~Kadav},
  \bibinfo{author}{I.~Durdanovic}, \bibinfo{author}{H.~Samet},
  \bibinfo{author}{H.~P. Graf},
\newblock \bibinfo{title}{Pruning filters for efficient convnets},
\newblock in: \bibinfo{booktitle}{5th International Conference on Learning
  Representations}, \bibinfo{year}{2017}.
\bibitem[{{Mao} et~al.(2017){Mao}, {Han}, {Pool}, {Li}, {Liu}, {Wang}, and
  {Dally}}]{mao_2017}
\bibinfo{author}{H.~{Mao}}, \bibinfo{author}{S.~{Han}},
  \bibinfo{author}{J.~{Pool}}, \bibinfo{author}{W.~{Li}},
  \bibinfo{author}{X.~{Liu}}, \bibinfo{author}{Y.~{Wang}},
  \bibinfo{author}{W.~J. {Dally}},
\newblock \bibinfo{title}{Exploring the granularity of sparsity in
  convolutional neural networks},
\newblock in: \bibinfo{booktitle}{Proceedings of the IEEE Conference on
  Computer Vision and Pattern Recognition Workshops}, \bibinfo{year}{2017}.
\bibitem[{Han et~al.(2016)Han, Liu, Mao, Pu, Pedram, Horowitz, and
  Dally}]{han_2016}
\bibinfo{author}{S.~Han}, \bibinfo{author}{X.~Liu}, \bibinfo{author}{H.~Mao},
  \bibinfo{author}{J.~Pu}, \bibinfo{author}{A.~Pedram}, \bibinfo{author}{M.~A.
  Horowitz}, \bibinfo{author}{W.~J. Dally},
\newblock \bibinfo{title}{Eie: Efficient inference engine on compressed deep
  neural network},
\newblock \bibinfo{journal}{ACM SIGARCH Computer Architecture News}
  \bibinfo{volume}{44} (\bibinfo{year}{2016}) \bibinfo{pages}{243--254}.
\bibitem[{Parashar et~al.(2017)Parashar, Rhu, Mukkara, Puglielli, Venkatesan,
  Khailany, Emer, Keckler, and Dally}]{parashar_2017}
\bibinfo{author}{A.~Parashar}, \bibinfo{author}{M.~Rhu},
  \bibinfo{author}{A.~Mukkara}, \bibinfo{author}{A.~Puglielli},
  \bibinfo{author}{R.~Venkatesan}, \bibinfo{author}{B.~Khailany},
  \bibinfo{author}{J.~Emer}, \bibinfo{author}{S.~W. Keckler},
  \bibinfo{author}{W.~J. Dally},
\newblock \bibinfo{title}{Scnn},
\newblock in: \bibinfo{booktitle}{Proceedings of the 44th Annual International
  Symposium on Computer Architecture}, \bibinfo{publisher}{ACM},
  \bibinfo{year}{2017}.
\bibitem[{Huang et~al.(2004)Huang, Zhu, and Siew}]{huang_2004}
\bibinfo{author}{G.-B. Huang}, \bibinfo{author}{Q.-Y. Zhu},
  \bibinfo{author}{C.-K. Siew},
\newblock \bibinfo{title}{Extreme learning machine: a new learning scheme of
  feedforward neural networks},
\newblock in: \bibinfo{booktitle}{IEEE International Joint Conference on Neural
  Networks 2}, \bibinfo{year}{2004}.
\bibitem[{Hoffer et~al.(2018)Hoffer, Hubara, and Soudry}]{hoffer_2018}
\bibinfo{author}{E.~Hoffer}, \bibinfo{author}{I.~Hubara},
  \bibinfo{author}{D.~Soudry},
\newblock \bibinfo{title}{Fix your classifier: the marginal value of training
  the last weight layer},
\newblock in: \bibinfo{booktitle}{6th International Conference on Learning
  Representations}, \bibinfo{year}{2018}.
\bibitem[{Rosenfeld and Tsotsos(2019)}]{rosenfeld_2019}
\bibinfo{author}{A.~Rosenfeld}, \bibinfo{author}{J.~K. Tsotsos},
\newblock \bibinfo{title}{Intriguing properties of randomly weighted networks:
  Generalizing while learning next to nothing},
\newblock in: \bibinfo{booktitle}{Conference on Computer and Robot Vision},
  \bibinfo{year}{2019}.
\bibitem[{Wimmer et~al.(2020)Wimmer, Mehnert, and Condurache}]{wimmer_2020}
\bibinfo{author}{P.~Wimmer}, \bibinfo{author}{J.~Mehnert},
  \bibinfo{author}{A.~P. Condurache},
\newblock \bibinfo{title}{{FreezeNet}: {Full} performance by reduced storage
  costs},
\newblock in: \bibinfo{booktitle}{Proceedings of the Asian Conference on
  Computer Vision}, \bibinfo{year}{2020}.
\bibitem[{Sung et~al.(2021)Sung, Nair, and Raffel}]{sung_2021}
\bibinfo{author}{Y.-L. Sung}, \bibinfo{author}{V.~Nair},
  \bibinfo{author}{C.~Raffel},
\newblock \bibinfo{title}{Training neural networks with fixed sparse masks},
\newblock in: \bibinfo{booktitle}{Advances in Neural Information Processing
  Systems 34}, \bibinfo{year}{2021}.
\bibitem[{Bellec et~al.(2018)Bellec, Kappel, Maass, and
  Legenstein}]{bellec_2018}
\bibinfo{author}{G.~Bellec}, \bibinfo{author}{D.~Kappel},
  \bibinfo{author}{W.~Maass}, \bibinfo{author}{R.~Legenstein},
\newblock \bibinfo{title}{Deep rewiring: Training very sparse deep networks},
\newblock in: \bibinfo{booktitle}{6th International Conference on Learning
  Representations}, \bibinfo{year}{2018}.
\bibitem[{Wimmer et~al.(2021)Wimmer, Mehnert, and Condurache}]{wimmer_2021}
\bibinfo{author}{P.~Wimmer}, \bibinfo{author}{J.~Mehnert},
  \bibinfo{author}{A.~P. Condurache},
\newblock \bibinfo{title}{{COPS}: {Controlled} pruning before training starts},
\newblock in: \bibinfo{booktitle}{International Joint Conference on Neural
  Networks}, \bibinfo{year}{2021}.
\bibitem[{Wimmer et~al.(2022)Wimmer, Mehnert, and Condurache}]{wimmer_2021b}
\bibinfo{author}{P.~Wimmer}, \bibinfo{author}{J.~Mehnert},
  \bibinfo{author}{A.~P. Condurache},
\newblock \bibinfo{title}{Interspace pruning: Using adaptive filter
  representations to improve training of sparse {CNN}s},
\newblock \bibinfo{journal}{CoRR} \bibinfo{volume}{abs/2203.07808}
  (\bibinfo{year}{2022}). \URLprefix \url{https://arxiv.org/abs/2203.07808},
  \bibinfo{note}{last accessed: 05/16/2022}.
\bibitem[{Diffenderfer and Kailkhura(2021)}]{diffenderfer_2021}
\bibinfo{author}{J.~Diffenderfer}, \bibinfo{author}{B.~Kailkhura},
\newblock \bibinfo{title}{Multi-prize lottery ticket hypothesis: Finding
  accurate binary neural networks by pruning a randomly weighted network},
\newblock in: \bibinfo{booktitle}{9th International Conference on Learning
  Representations}, \bibinfo{year}{2021}.
\bibitem[{Wang et~al.(2021)Wang, Qin, Zhang, and Fu}]{wang_2021c}
\bibinfo{author}{H.~Wang}, \bibinfo{author}{C.~Qin},
  \bibinfo{author}{Y.~Zhang}, \bibinfo{author}{Y.~Fu},
\newblock \bibinfo{title}{Emerging paradigms of neural network pruning},
\newblock \bibinfo{journal}{CoRR} \bibinfo{volume}{abs/2103.06460v2}
  (\bibinfo{year}{2021}). \URLprefix \url{https://arxiv.org/abs/2103.06460v2},
  \bibinfo{note}{last accessed: 05/05/2022}.
\bibitem[{{He} et~al.(2015){He}, {Zhang}, {Ren}, and {Sun}}]{He2015}
\bibinfo{author}{K.~{He}}, \bibinfo{author}{X.~{Zhang}},
  \bibinfo{author}{S.~{Ren}}, \bibinfo{author}{J.~{Sun}},
\newblock \bibinfo{title}{Delving deep into rectifiers: Surpassing human-level
  performance on imagenet classification},
\newblock in: \bibinfo{booktitle}{IEEE International Conference on Computer
  Vision}, \bibinfo{year}{2015}.
\bibitem[{Glorot and Bengio(2010)}]{xavier_2010}
\bibinfo{author}{X.~Glorot}, \bibinfo{author}{Y.~Bengio},
\newblock \bibinfo{title}{Understanding the difficulty of training deep
  feedforward neural networks},
\newblock in: \bibinfo{booktitle}{Proceedings of the 13th International
  Conference on Artificial Intelligence and Statistics}, \bibinfo{year}{2010}.
\bibitem[{Saxe et~al.(2014)Saxe, McClelland, and Ganguli}]{saxe_2014}
\bibinfo{author}{A.~M. Saxe}, \bibinfo{author}{J.~L. McClelland},
  \bibinfo{author}{S.~Ganguli},
\newblock \bibinfo{title}{Exact solutions to the nonlinear dynamics of learning
  in deep linear neural networks},
\newblock in: \bibinfo{booktitle}{2nd International Conference on Learning
  Representations}, \bibinfo{year}{2014}.
\bibitem[{Martens(2010)}]{martens_2010}
\bibinfo{author}{J.~Martens},
\newblock \bibinfo{title}{Deep learning via hessian-free optimization},
\newblock in: \bibinfo{booktitle}{Proceedings of the 27th International
  Conference on Machine Learning}, \bibinfo{year}{2010}.
\bibitem[{Sutskever et~al.(2013)Sutskever, Martens, Dahl, and
  Hinton}]{sutskever_2013}
\bibinfo{author}{I.~Sutskever}, \bibinfo{author}{J.~Martens},
  \bibinfo{author}{G.~Dahl}, \bibinfo{author}{G.~Hinton},
\newblock \bibinfo{title}{On the importance of initialization and momentum in
  deep learning},
\newblock in: \bibinfo{booktitle}{Proceedings of the 30th International
  Conference on Machine Learning}, \bibinfo{year}{2013}.
\bibitem[{Hanin and Rolnick(2018)}]{hanin_2018}
\bibinfo{author}{B.~Hanin}, \bibinfo{author}{D.~Rolnick},
\newblock \bibinfo{title}{How to start training: The effect of initialization
  and architecture},
\newblock in: \bibinfo{booktitle}{Advances in Neural Information Processing
  Systems 31}, \bibinfo{year}{2018}.
\bibitem[{Robbins and Monro(1951)}]{robbins_2007}
\bibinfo{author}{H.~Robbins}, \bibinfo{author}{S.~Monro},
\newblock \bibinfo{title}{A stochastic approximation method},
\newblock \bibinfo{journal}{The Annals of Mathematical Statistics}
  \bibinfo{volume}{22} (\bibinfo{year}{1951}) \bibinfo{pages}{400--407}.
\bibitem[{Duchi et~al.(2011)Duchi, Hazan, and Singer}]{duchi_2011}
\bibinfo{author}{J.~Duchi}, \bibinfo{author}{E.~Hazan},
  \bibinfo{author}{Y.~Singer},
\newblock \bibinfo{title}{Adaptive subgradient methods for online learning and
  stochastic optimization},
\newblock \bibinfo{journal}{Journal of Machine Learning Research}
  \bibinfo{volume}{12} (\bibinfo{year}{2011}) \bibinfo{pages}{2121--2159}.
\bibitem[{Kingma and Ba(2015)}]{kingma_2014}
\bibinfo{author}{D.~P. Kingma}, \bibinfo{author}{J.~Ba},
\newblock \bibinfo{title}{Adam: {A} method for stochastic optimization},
\newblock in: \bibinfo{booktitle}{3rd International Conference on Learning
  Representations}, \bibinfo{year}{2015}.
\bibitem[{Loshchilov and Hutter(2019)}]{loshchilov_2018d}
\bibinfo{author}{I.~Loshchilov}, \bibinfo{author}{F.~Hutter},
\newblock \bibinfo{title}{Decoupled weight decay regularization},
\newblock in: \bibinfo{booktitle}{7th International Conference on Learning
  Representations}, \bibinfo{year}{2019}.
\bibitem[{Krogh and Hertz(1992)}]{krogh_1991}
\bibinfo{author}{A.~Krogh}, \bibinfo{author}{J.~A. Hertz},
\newblock \bibinfo{title}{A simple weight decay can improve generalization},
\newblock in: \bibinfo{booktitle}{Advances in Neural Information Processing
  Systems 4}, \bibinfo{year}{1992}.
\bibitem[{Hinton et~al.(2012)Hinton, Srivastava, Krizhevsky, Sutskever, and
  Salakhutdinov}]{hinton_2012}
\bibinfo{author}{G.~E. Hinton}, \bibinfo{author}{N.~Srivastava},
  \bibinfo{author}{A.~Krizhevsky}, \bibinfo{author}{I.~Sutskever},
  \bibinfo{author}{R.~Salakhutdinov},
\newblock \bibinfo{title}{Improving neural networks by preventing co-adaptation
  of feature detectors},
\newblock \bibinfo{journal}{CoRR} \bibinfo{volume}{abs/1207.0580}
  (\bibinfo{year}{2012}). \URLprefix \url{http://arxiv.org/abs/1207.0580},
  \bibinfo{note}{last accessed: 03/04/2022}.
\bibitem[{Ba and Frey(2013)}]{ba_2013}
\bibinfo{author}{J.~Ba}, \bibinfo{author}{B.~Frey},
\newblock \bibinfo{title}{Adaptive dropout for training deep neural networks},
\newblock in: \bibinfo{booktitle}{Advances in Neural Information Processing
  Systems 26}, \bibinfo{year}{2013}.
\bibitem[{Kingma et~al.(2015)Kingma, Salimans, and Welling}]{kingma_2015}
\bibinfo{author}{D.~P. Kingma}, \bibinfo{author}{T.~Salimans},
  \bibinfo{author}{M.~Welling},
\newblock \bibinfo{title}{Variational dropout and the local reparameterization
  trick},
\newblock in: \bibinfo{booktitle}{Advances in Neural Information Processing
  Systems 28}, \bibinfo{year}{2015}.
\bibitem[{Ioffe and Szegedy(2015)}]{ioffe_2015}
\bibinfo{author}{S.~Ioffe}, \bibinfo{author}{C.~Szegedy},
\newblock \bibinfo{title}{Batch normalization: Accelerating deep network
  training by reducing internal covariate shift},
\newblock in: \bibinfo{booktitle}{Proceedings of the 32nd International
  Conference on Machine Learning}, \bibinfo{year}{2015}.
\bibitem[{Yao et~al.(2007)Yao, Rosasco, and Caponnetto}]{yao_2007}
\bibinfo{author}{Y.~Yao}, \bibinfo{author}{L.~Rosasco},
  \bibinfo{author}{A.~Caponnetto},
\newblock \bibinfo{title}{On early stopping in gradient descent learning},
\newblock \bibinfo{journal}{Constructive Approximation} \bibinfo{volume}{26}
  (\bibinfo{year}{2007}) \bibinfo{pages}{289--315}.
\bibitem[{Finnoff et~al.(1993)Finnoff, Hergert, and Zimmermann}]{finnoff_1993}
\bibinfo{author}{W.~Finnoff}, \bibinfo{author}{F.~Hergert},
  \bibinfo{author}{H.~G. Zimmermann},
\newblock \bibinfo{title}{Improving model selection by nonconvergent methods},
\newblock \bibinfo{journal}{Neural Networks} \bibinfo{volume}{6}
  (\bibinfo{year}{1993}) \bibinfo{pages}{771--783}.
\bibitem[{Cohen and Welling(2016)}]{cohen_2016}
\bibinfo{author}{T.~S. Cohen}, \bibinfo{author}{M.~Welling},
\newblock \bibinfo{title}{Group equivariant convolutional networks},
\newblock in: \bibinfo{booktitle}{Proceedings of the 33rd International
  Conference on Machine Learning}, \bibinfo{year}{2016}.
\bibitem[{Jaderberg et~al.(2015)Jaderberg, Simonyan, Zisserman, and
  Kavukcuoglu}]{jaderberg_2015}
\bibinfo{author}{M.~Jaderberg}, \bibinfo{author}{K.~Simonyan},
  \bibinfo{author}{A.~Zisserman}, \bibinfo{author}{K.~Kavukcuoglu},
\newblock \bibinfo{title}{Spatial transformer networks},
\newblock in: \bibinfo{booktitle}{Advances in Neural Information Processing
  Systems 28}, \bibinfo{year}{2015}.
\bibitem[{Rath and Condurache(2020)}]{rath_2020}
\bibinfo{author}{M.~Rath}, \bibinfo{author}{A.~P. Condurache},
\newblock \bibinfo{title}{Invariant integration in deep convolutional feature
  space},
\newblock in: \bibinfo{booktitle}{28th European Symposium on Artificial Neural
  Networks, Computational Intelligence and Machine Learning},
  \bibinfo{year}{2020}.
\bibitem[{Rath and Condurache(2022)}]{rath_2022}
\bibinfo{author}{M.~Rath}, \bibinfo{author}{A.~P. Condurache},
\newblock \bibinfo{title}{Improving the sample-complexity of deep
  classification networks with invariant integration},
\newblock in: \bibinfo{booktitle}{Proceedings of the 17th International Joint
  Conference on Computer Vision, Imaging and Computer Graphics Theory and
  Applications}, \bibinfo{year}{2022}.
\bibitem[{Coors et~al.(2018)Coors, Condurache, and Geiger}]{coors_2018}
\bibinfo{author}{B.~Coors}, \bibinfo{author}{A.~P. Condurache},
  \bibinfo{author}{A.~Geiger},
\newblock \bibinfo{title}{Spherenet: Learning spherical representations for
  detection and classification in omnidirectional images},
\newblock in: \bibinfo{booktitle}{Proceedings of the European Conference on
  Computer Vision}, \bibinfo{year}{2018}.
\bibitem[{Lust and Condurache(2020)}]{lust_2020b}
\bibinfo{author}{J.~Lust}, \bibinfo{author}{A.~P. Condurache},
\newblock \bibinfo{title}{Gran: An efficient gradient-norm based detector for
  adversarial and misclassified examples},
\newblock in: \bibinfo{booktitle}{28th European Symposium on Artificial Neural
  Networks, Computational Intelligence and Machine Learning},
  \bibinfo{year}{2020}.
\bibitem[{Lust and Condurache(2022)}]{lust_2022}
\bibinfo{author}{J.~Lust}, \bibinfo{author}{A.~P. Condurache},
\newblock \bibinfo{title}{Efficient detection of adversarial,
  out-of-distribution and other misclassified samples},
\newblock \bibinfo{journal}{Neurocomputing} \bibinfo{volume}{470}
  (\bibinfo{year}{2022}) \bibinfo{pages}{335--343}.
\bibitem[{Ren et~al.(2019)Ren, Liu, Fertig, Snoek, Poplin, Depristo, Dillon,
  and Lakshminarayanan}]{ren_2019}
\bibinfo{author}{J.~Ren}, \bibinfo{author}{P.~J. Liu},
  \bibinfo{author}{E.~Fertig}, \bibinfo{author}{J.~Snoek},
  \bibinfo{author}{R.~Poplin}, \bibinfo{author}{M.~Depristo},
  \bibinfo{author}{J.~Dillon}, \bibinfo{author}{B.~Lakshminarayanan},
\newblock \bibinfo{title}{Likelihood ratios for out-of-distribution detection},
\newblock in: \bibinfo{booktitle}{Advances in Neural Information Processing
  Systems 32}, \bibinfo{year}{2019}.
\bibitem[{Serr\`a et~al.(2020)Serr\`a, Álvarez, Gómez, Slizovskaia, Núñez,
  and Luque}]{serra_2020}
\bibinfo{author}{J.~Serr\`a}, \bibinfo{author}{D.~Álvarez},
  \bibinfo{author}{V.~Gómez}, \bibinfo{author}{O.~Slizovskaia},
  \bibinfo{author}{J.~F. Núñez}, \bibinfo{author}{J.~Luque},
\newblock \bibinfo{title}{Input complexity and out-of-distribution detection
  with likelihood-based generative models},
\newblock in: \bibinfo{booktitle}{8th International Conference on Learning
  Representations}, \bibinfo{year}{2020}.
\bibitem[{Hassibi and Stork(1992)}]{hassibi_1993}
\bibinfo{author}{B.~Hassibi}, \bibinfo{author}{D.~Stork},
\newblock \bibinfo{title}{Second order derivatives for network pruning: Optimal
  brain surgeon},
\newblock in: \bibinfo{booktitle}{Advances in Neural Information Processing
  Systems 4}, \bibinfo{year}{1992}.
\bibitem[{Mostafa and Wang(2019)}]{mostafa_2019}
\bibinfo{author}{H.~Mostafa}, \bibinfo{author}{X.~Wang},
\newblock \bibinfo{title}{Parameter efficient training of deep convolutional
  neural networks by dynamic sparse reparameterization},
\newblock in: \bibinfo{booktitle}{Proceedings of the 36th International
  Conference on Machine Learning}, \bibinfo{year}{2019}.
\bibitem[{Price and Tanner(2021)}]{price_2021}
\bibinfo{author}{I.~Price}, \bibinfo{author}{J.~Tanner},
\newblock \bibinfo{title}{Dense for the price of sparse: Improved performance
  of sparsely initialized networks via a subspace offset},
\newblock in: \bibinfo{booktitle}{Proceedings of the 38th International
  Conference on Machine Learning}, \bibinfo{year}{2021}.
\bibitem[{Huang et~al.(2011)Huang, Wang, and Lan}]{huang_2011}
\bibinfo{author}{G.-B. Huang}, \bibinfo{author}{D.~H. Wang},
  \bibinfo{author}{Y.~Lan},
\newblock \bibinfo{title}{Extreme learning machines: a survey},
\newblock \bibinfo{journal}{International journal of machine learning and
  cybernetics} \bibinfo{volume}{2} (\bibinfo{year}{2011})
  \bibinfo{pages}{107--122}.
\bibitem[{Qing et~al.(2020)Qing, Zeng, Li, and Huang}]{qing_2020}
\bibinfo{author}{Y.~Qing}, \bibinfo{author}{Y.~Zeng}, \bibinfo{author}{Y.~Li},
  \bibinfo{author}{G.-B. Huang},
\newblock \bibinfo{title}{Deep and wide feature based extreme learning machine
  for image classification},
\newblock \bibinfo{journal}{Neurocomputing} \bibinfo{volume}{412}
  (\bibinfo{year}{2020}) \bibinfo{pages}{426--436}.
\bibitem[{Saxe et~al.(2011)Saxe, Koh, Chen, Bhand, Suresh, and Ng}]{saxe_2011}
\bibinfo{author}{A.~Saxe}, \bibinfo{author}{P.~W. Koh},
  \bibinfo{author}{Z.~Chen}, \bibinfo{author}{M.~Bhand},
  \bibinfo{author}{B.~Suresh}, \bibinfo{author}{A.~Ng},
\newblock \bibinfo{title}{On random weights and unsupervised feature learning},
\newblock in: \bibinfo{booktitle}{Proceedings of the 28th International
  Conference on Machine Learning}, \bibinfo{year}{2011}.
\bibitem[{Giryes et~al.(2016)Giryes, Sapiro, and Bronstein}]{giryes_2016}
\bibinfo{author}{R.~Giryes}, \bibinfo{author}{G.~Sapiro},
  \bibinfo{author}{A.~M. Bronstein},
\newblock \bibinfo{title}{Deep neural networks with random gaussian weights: A
  universal classification strategy?},
\newblock \bibinfo{journal}{IEEE Trans. Signal Process.} \bibinfo{volume}{64}
  (\bibinfo{year}{2016}) \bibinfo{pages}{3444--3457}.
\bibitem[{Li et~al.(2018)Li, Farkhoor, Liu, and Yosinski}]{li_2018}
\bibinfo{author}{C.~Li}, \bibinfo{author}{H.~Farkhoor},
  \bibinfo{author}{R.~Liu}, \bibinfo{author}{J.~Yosinski},
\newblock \bibinfo{title}{Measuring the intrinsic dimension of objective
  landscapes},
\newblock in: \bibinfo{booktitle}{6th International Conference on Learning
  Representations}, \bibinfo{year}{2018}.
\bibitem[{Tinney and Walker(1967)}]{tinney_1967}
\bibinfo{author}{W.~Tinney}, \bibinfo{author}{J.~Walker},
\newblock \bibinfo{title}{Direct solutions of sparse network equations by
  optimally ordered triangular factorization},
\newblock \bibinfo{journal}{Proceedings of the IEEE} \bibinfo{volume}{55}
  (\bibinfo{year}{1967}) \bibinfo{pages}{1801--1809}.
\bibitem[{Liu et~al.(2019)Liu, Sun, Zhou, Huang, and Darrell}]{liu_2018}
\bibinfo{author}{Z.~Liu}, \bibinfo{author}{M.~Sun}, \bibinfo{author}{T.~Zhou},
  \bibinfo{author}{G.~Huang}, \bibinfo{author}{T.~Darrell},
\newblock \bibinfo{title}{Rethinking the value of network pruning},
\newblock in: \bibinfo{booktitle}{7th International Conference on Learning
  Representations}, \bibinfo{year}{2019}.
\bibitem[{Wang et~al.(2020)Wang, Zhang, Xie, Zhou, Su, Zhang, and
  Hu}]{wang_2020b}
\bibinfo{author}{Y.~Wang}, \bibinfo{author}{X.~Zhang},
  \bibinfo{author}{L.~Xie}, \bibinfo{author}{J.~Zhou}, \bibinfo{author}{H.~Su},
  \bibinfo{author}{B.~Zhang}, \bibinfo{author}{X.~Hu},
\newblock \bibinfo{title}{Pruning from scratch},
\newblock in: \bibinfo{booktitle}{Proceedings of the AAAI Conference on
  Artificial Intelligence}, \bibinfo{year}{2020}.
\bibitem[{Verdenius et~al.(2020)Verdenius, Stol, and
  Forr{\'{e}}}]{verdenius_2020}
\bibinfo{author}{S.~Verdenius}, \bibinfo{author}{M.~Stol},
  \bibinfo{author}{P.~Forr{\'{e}}},
\newblock \bibinfo{title}{Pruning via iterative ranking of sensitivity
  statistics},
\newblock \bibinfo{journal}{CoRR} \bibinfo{volume}{abs/2006.00896v2}
  (\bibinfo{year}{2020}). \URLprefix \url{https://arxiv.org/abs/2006.00896v2},
  \bibinfo{note}{last accessed: 04/19/2022}.
\bibitem[{Park et~al.(2017)Park, Li, Wen, Tang, Li, Chen, and
  Dubey}]{park_2017}
\bibinfo{author}{J.~Park}, \bibinfo{author}{S.~R. Li},
  \bibinfo{author}{W.~Wen}, \bibinfo{author}{P.~T.~P. Tang},
  \bibinfo{author}{H.~Li}, \bibinfo{author}{Y.~Chen},
  \bibinfo{author}{P.~Dubey},
\newblock \bibinfo{title}{Faster cnns with direct sparse convolutions and
  guided pruning},
\newblock in: \bibinfo{booktitle}{5th International Conference on Learning
  Representations}, \bibinfo{year}{2017}.
\bibitem[{Liu et~al.(2021)Liu, Mocanu, Matavalam, Pei, and
  Pechenizkiy}]{liu_2021}
\bibinfo{author}{S.~Liu}, \bibinfo{author}{D.~C. Mocanu},
  \bibinfo{author}{A.~R.~R. Matavalam}, \bibinfo{author}{Y.~Pei},
  \bibinfo{author}{M.~Pechenizkiy},
\newblock \bibinfo{title}{Sparse evolutionary deep learning with over one
  million artificial neurons on commodity hardware},
\newblock \bibinfo{journal}{Neural Comput. Appl.} \bibinfo{volume}{33}
  (\bibinfo{year}{2021}) \bibinfo{pages}{2589--2604}.
\bibitem[{Elsen et~al.(2020)Elsen, Dukhan, Gale, and Simonyan}]{elsen_2020}
\bibinfo{author}{E.~Elsen}, \bibinfo{author}{M.~Dukhan},
  \bibinfo{author}{T.~Gale}, \bibinfo{author}{K.~Simonyan},
\newblock \bibinfo{title}{Fast sparse convnets},
\newblock in: \bibinfo{booktitle}{Proceedings of the IEEE/CVF Conference on
  Computer Vision and Pattern Recognition}, \bibinfo{year}{2020}.
\bibitem[{Gale et~al.(2020)Gale, Zaharia, Young, and Elsen}]{gale_2020}
\bibinfo{author}{T.~Gale}, \bibinfo{author}{M.~Zaharia},
  \bibinfo{author}{C.~Young}, \bibinfo{author}{E.~Elsen},
\newblock \bibinfo{title}{Sparse gpu kernels for deep learning},
\newblock in: \bibinfo{booktitle}{Proceedings of the International Conference
  for High Performance Computing, Networking, Storage and Analysis},
  \bibinfo{year}{2020}.
\bibitem[{Wang(2020)}]{wang_2020c}
\bibinfo{author}{Z.~Wang},
\newblock \bibinfo{title}{Sparsert: Accelerating unstructured sparsity on gpus
  for deep learning inference},
\newblock in: \bibinfo{booktitle}{Proceedings of the ACM International
  Conference on Parallel Architectures and Compilation Techniques},
  \bibinfo{year}{2020}.
\bibitem[{Hubara et~al.(2021)Hubara, Chmiel, Island, Banner, Naor, and
  Soudry}]{hubara_2021}
\bibinfo{author}{I.~Hubara}, \bibinfo{author}{B.~Chmiel},
  \bibinfo{author}{M.~Island}, \bibinfo{author}{R.~Banner},
  \bibinfo{author}{J.~Naor}, \bibinfo{author}{D.~Soudry},
\newblock \bibinfo{title}{Accelerated sparse neural training: A provable and
  efficient method to find n:m transposable masks},
\newblock in: \bibinfo{booktitle}{Advances in Neural Information Processing
  Systems 34}, \bibinfo{year}{2021}.
\bibitem[{Zhou et~al.(2021)Zhou, Ma, Zhu, Liu, Zhang, Yuan, Sun, and
  Li}]{zhou_2021}
\bibinfo{author}{A.~Zhou}, \bibinfo{author}{Y.~Ma}, \bibinfo{author}{J.~Zhu},
  \bibinfo{author}{J.~Liu}, \bibinfo{author}{Z.~Zhang},
  \bibinfo{author}{K.~Yuan}, \bibinfo{author}{W.~Sun}, \bibinfo{author}{H.~Li},
\newblock \bibinfo{title}{Learning n:m fine-grained structured sparse neural
  networks from scratch},
\newblock in: \bibinfo{booktitle}{9th International Conference on Learning
  Representations}, \bibinfo{year}{2021}.
\bibitem[{Sun et~al.(2021)Sun, Zhou, Stuijk, Wijnhoven, Nelson, Li, and
  Corporaal}]{sun_2021}
\bibinfo{author}{W.~Sun}, \bibinfo{author}{A.~Zhou},
  \bibinfo{author}{S.~Stuijk}, \bibinfo{author}{R.~G.~J. Wijnhoven},
  \bibinfo{author}{A.~Nelson}, \bibinfo{author}{H.~Li},
  \bibinfo{author}{H.~Corporaal},
\newblock \bibinfo{title}{Dominosearch: Find layer-wise fine-grained n:m sparse
  schemes from dense neural networks},
\newblock in: \bibinfo{booktitle}{Advances in Neural Information Processing
  Systems 34}, \bibinfo{year}{2021}.
\bibitem[{Pool and Yu(2021)}]{pool_2021}
\bibinfo{author}{J.~Pool}, \bibinfo{author}{C.~Yu},
\newblock \bibinfo{title}{Channel permutations for n:m sparsity},
\newblock in: \bibinfo{booktitle}{Advances in Neural Information Processing
  Systems 34}, \bibinfo{year}{2021}.
\bibitem[{Holmes et~al.(2021)Holmes, Zhang, He, and Wu}]{holmes_2021}
\bibinfo{author}{C.~Holmes}, \bibinfo{author}{M.~Zhang},
  \bibinfo{author}{Y.~He}, \bibinfo{author}{B.~Wu},
\newblock \bibinfo{title}{Nx{MT}ransformer: Semi-structured sparsification for
  natural language understanding via {ADMM}},
\newblock in: \bibinfo{booktitle}{Advances in Neural Information Processing
  Systems 34}, \bibinfo{year}{2021}.
\bibitem[{NVIDIA(2020)}]{nvidia_2020}
\bibinfo{author}{NVIDIA}, \bibinfo{title}{Nvidia a100 tensor core gpu
  architecture}, \bibinfo{year}{2020}. \bibinfo{note}{{URL}:
  \url{https://images.nvidia.com/aem-dam/en-zz/Solutions/data-center/nvidia-ampere-architecture-whitepaper.pdf},
  last accessed: 03/30/2022}.
\bibitem[{Sanh et~al.(2020)Sanh, Wolf, and Rush}]{sanh_2020}
\bibinfo{author}{V.~Sanh}, \bibinfo{author}{T.~Wolf}, \bibinfo{author}{A.~M.
  Rush},
\newblock \bibinfo{title}{Movement pruning: Adaptive sparsity by fine-tuning},
\newblock in: \bibinfo{booktitle}{Advances in Neural Information Processing
  Systems 33}, \bibinfo{year}{2020}.
\bibitem[{Jayakumar et~al.(2020)Jayakumar, Pascanu, Rae, Osindero, and
  Elsen}]{jayakumar_2020}
\bibinfo{author}{S.~Jayakumar}, \bibinfo{author}{R.~Pascanu},
  \bibinfo{author}{J.~Rae}, \bibinfo{author}{S.~Osindero},
  \bibinfo{author}{E.~Elsen},
\newblock \bibinfo{title}{Top-kast: Top-k always sparse training},
\newblock in: \bibinfo{booktitle}{Advances in Neural Information Processing
  Systems 33}, \bibinfo{year}{2020}.
\bibitem[{Mocanu et~al.(2018)Mocanu, Mocanu, Stone, Nguyen, Gibescu, and
  Liotta}]{mocanu_2018}
\bibinfo{author}{D.~Mocanu}, \bibinfo{author}{E.~Mocanu},
  \bibinfo{author}{P.~Stone}, \bibinfo{author}{P.~Nguyen},
  \bibinfo{author}{M.~Gibescu}, \bibinfo{author}{A.~Liotta},
\newblock \bibinfo{title}{Scalable training of artificial neural networks with
  adaptive sparse connectivity inspired by network science},
\newblock \bibinfo{journal}{Nature Communications} \bibinfo{volume}{9}
  (\bibinfo{year}{2018}) \bibinfo{pages}{2383}.
\bibitem[{Evci et~al.(2020)Evci, Gale, Menick, Castro, and Elsen}]{evci_2020}
\bibinfo{author}{U.~Evci}, \bibinfo{author}{T.~Gale},
  \bibinfo{author}{J.~Menick}, \bibinfo{author}{P.~S. Castro},
  \bibinfo{author}{E.~Elsen},
\newblock \bibinfo{title}{Rigging the lottery: Making all tickets winners},
\newblock in: \bibinfo{booktitle}{Proceedings of the 37th International
  Conference on Machine Learning}, \bibinfo{year}{2020}.
\bibitem[{Dettmers and Zettlemoyer(2019)}]{dettmers_2019}
\bibinfo{author}{T.~Dettmers}, \bibinfo{author}{L.~Zettlemoyer},
\newblock \bibinfo{title}{Sparse networks from scratch: Faster training without
  losing performance},
\newblock \bibinfo{journal}{CoRR} \bibinfo{volume}{abs/1907.04840v2}
  (\bibinfo{year}{2019}). \URLprefix \url{http://arxiv.org/abs/1907.04840v2},
  \bibinfo{note}{last accessed: 03/31/2022}.
\bibitem[{Zhou et~al.(2019)Zhou, Lan, Liu, and Yosinski}]{zhou_2019}
\bibinfo{author}{H.~Zhou}, \bibinfo{author}{J.~Lan}, \bibinfo{author}{R.~Liu},
  \bibinfo{author}{J.~Yosinski},
\newblock \bibinfo{title}{Deconstructing lottery tickets: Zeros, signs, and the
  supermask},
\newblock in: \bibinfo{booktitle}{Advances in Neural Information Processing
  Systems 32}, \bibinfo{year}{2019}.
\bibitem[{Frankle et~al.(2020)Frankle, Dziugaite, Roy, and
  Carbin}]{frankle_2020a}
\bibinfo{author}{J.~Frankle}, \bibinfo{author}{G.~K. Dziugaite},
  \bibinfo{author}{D.~Roy}, \bibinfo{author}{M.~Carbin},
\newblock \bibinfo{title}{Linear mode connectivity and the lottery ticket
  hypothesis},
\newblock in: \bibinfo{booktitle}{Proceedings of the 37th International
  Conference on Machine Learning}, \bibinfo{year}{2020}.
\bibitem[{Redman et~al.(2022)Redman, FONOBEROVA, Mohr, Kevrekidis, and
  Mezic}]{redman_2022}
\bibinfo{author}{W.~T. Redman}, \bibinfo{author}{M.~FONOBEROVA},
  \bibinfo{author}{R.~Mohr}, \bibinfo{author}{Y.~Kevrekidis},
  \bibinfo{author}{I.~Mezic},
\newblock \bibinfo{title}{An operator theoretic view on pruning deep neural
  networks},
\newblock in: \bibinfo{booktitle}{10th International Conference on Learning
  Representations}, \bibinfo{year}{2022}.
\bibitem[{Mezi{\'{c}}(2005)}]{mezic_2005}
\bibinfo{author}{I.~Mezi{\'{c}}},
\newblock \bibinfo{title}{Spectral properties of dynamical systems, model
  reduction and decompositions},
\newblock \bibinfo{journal}{Nonlinear Dynamics} \bibinfo{volume}{41}
  (\bibinfo{year}{2005}) \bibinfo{pages}{309--325}.
\bibitem[{Lee et~al.(2021)Lee, Park, Mo, Ahn, and Shin}]{lee_2021}
\bibinfo{author}{J.~Lee}, \bibinfo{author}{S.~Park}, \bibinfo{author}{S.~Mo},
  \bibinfo{author}{S.~Ahn}, \bibinfo{author}{J.~Shin},
\newblock \bibinfo{title}{Layer-adaptive sparsity for the magnitude-based
  pruning},
\newblock in: \bibinfo{booktitle}{9th International Conference on Learning
  Representations}, \bibinfo{year}{2021}.
\bibitem[{Frankle et~al.(2021)Frankle, Dziugaite, Roy, and
  Carbin}]{frankle_2021}
\bibinfo{author}{J.~Frankle}, \bibinfo{author}{G.~K. Dziugaite},
  \bibinfo{author}{D.~Roy}, \bibinfo{author}{M.~Carbin},
\newblock \bibinfo{title}{Pruning neural networks at initialization: Why are we
  missing the mark?},
\newblock in: \bibinfo{booktitle}{9th International Conference on Learning
  Representations}, \bibinfo{year}{2021}.
\bibitem[{Lee et~al.(2020)Lee, Ajanthan, Gould, and Torr}]{lee_2019}
\bibinfo{author}{N.~Lee}, \bibinfo{author}{T.~Ajanthan},
  \bibinfo{author}{S.~Gould}, \bibinfo{author}{P.~H.~S. Torr},
\newblock \bibinfo{title}{A signal propagation perspective for pruning neural
  networks at initialization},
\newblock in: \bibinfo{booktitle}{8th International Conference on Learning
  Representations}, \bibinfo{year}{2020}.
\bibitem[{de~Jorge et~al.(2021)de~Jorge, Sanyal, Behl, Torr, Rogez, and
  Dokania}]{jorge_2020}
\bibinfo{author}{P.~de~Jorge}, \bibinfo{author}{A.~Sanyal},
  \bibinfo{author}{H.~Behl}, \bibinfo{author}{P.~Torr},
  \bibinfo{author}{G.~Rogez}, \bibinfo{author}{P.~K. Dokania},
\newblock \bibinfo{title}{Progressive skeletonization: Trimming more fat from a
  network at initialization},
\newblock in: \bibinfo{booktitle}{9th International Conference on Learning
  Representations}, \bibinfo{year}{2021}.
\bibitem[{Hayou et~al.(2021)Hayou, Ton, Doucet, and Teh}]{hayou_2021}
\bibinfo{author}{S.~Hayou}, \bibinfo{author}{J.-F. Ton},
  \bibinfo{author}{A.~Doucet}, \bibinfo{author}{Y.~W. Teh},
\newblock \bibinfo{title}{Robust pruning at initialization},
\newblock in: \bibinfo{booktitle}{9th International Conference on Learning
  Representations}, \bibinfo{year}{2021}.
\bibitem[{Poole et~al.(2016)Poole, Lahiri, Raghu, Sohl-Dickstein, and
  Ganguli}]{poole_2016}
\bibinfo{author}{B.~Poole}, \bibinfo{author}{S.~Lahiri},
  \bibinfo{author}{M.~Raghu}, \bibinfo{author}{J.~Sohl-Dickstein},
  \bibinfo{author}{S.~Ganguli},
\newblock \bibinfo{title}{Exponential expressivity in deep neural networks
  through transient chaos},
\newblock in: \bibinfo{booktitle}{Advances in Neural Information Processing
  Systems 29}, \bibinfo{year}{2016}.
\bibitem[{Schoenholz et~al.(2017)Schoenholz, Gilmer, Ganguli, and
  Sohl{-}Dickstein}]{schoenholz_2017}
\bibinfo{author}{S.~S. Schoenholz}, \bibinfo{author}{J.~Gilmer},
  \bibinfo{author}{S.~Ganguli}, \bibinfo{author}{J.~Sohl{-}Dickstein},
\newblock \bibinfo{title}{Deep information propagation},
\newblock in: \bibinfo{booktitle}{5th International Conference on Learning
  Representations}, \bibinfo{year}{2017}.
\bibitem[{Xiao et~al.(2018)Xiao, Bahri, Sohl{-}Dickstein, Schoenholz, and
  Pennington}]{xiao_2018}
\bibinfo{author}{L.~Xiao}, \bibinfo{author}{Y.~Bahri},
  \bibinfo{author}{J.~Sohl{-}Dickstein}, \bibinfo{author}{S.~S. Schoenholz},
  \bibinfo{author}{J.~Pennington},
\newblock \bibinfo{title}{Dynamical isometry and a mean field theory of cnns:
  How to train 10, 000-layer vanilla convolutional neural networks},
\newblock in: \bibinfo{booktitle}{Proceedings of the 35th International
  Conference on Machine Learning}, \bibinfo{year}{2018}.
\bibitem[{Neyshabur et~al.(2015{\natexlab{a}})Neyshabur, Salakhutdinov, and
  Srebro}]{neyshabur_2015}
\bibinfo{author}{B.~Neyshabur}, \bibinfo{author}{R.~Salakhutdinov},
  \bibinfo{author}{N.~Srebro},
\newblock \bibinfo{title}{Path-sgd: Path-normalized optimization in deep neural
  networks},
\newblock in: \bibinfo{booktitle}{Advances in Neural Information Processing
  Systems 28}, \bibinfo{year}{2015}{\natexlab{a}}.
\bibitem[{Neyshabur et~al.(2015{\natexlab{b}})Neyshabur, Tomioka, and
  Srebro}]{neyshabur_2015b}
\bibinfo{author}{B.~Neyshabur}, \bibinfo{author}{R.~Tomioka},
  \bibinfo{author}{N.~Srebro},
\newblock \bibinfo{title}{Norm-based capacity control in neural networks},
\newblock in: \bibinfo{booktitle}{Proceedings of The 28th Conference on
  Learning Theory}, \bibinfo{year}{2015}{\natexlab{b}}.
\bibitem[{Gebhart et~al.(2021)Gebhart, Saxena, and Schrater}]{gebhart_2021}
\bibinfo{author}{T.~Gebhart}, \bibinfo{author}{U.~Saxena},
  \bibinfo{author}{P.~Schrater},
\newblock \bibinfo{title}{A unified paths perspective for pruning at
  initialization},
\newblock \bibinfo{journal}{CoRR} \bibinfo{volume}{abs/2101.10552}
  (\bibinfo{year}{2021}). \URLprefix \url{https://arxiv.org/abs/2101.10552},
  \bibinfo{note}{last accessed: 04/19/2022}.
\bibitem[{Patil and Dovrolis(2021)}]{patil_2021}
\bibinfo{author}{S.~M. Patil}, \bibinfo{author}{C.~Dovrolis},
\newblock \bibinfo{title}{{PHEW}: Constructing sparse networks that learn fast
  and generalize well without training data},
\newblock in: \bibinfo{booktitle}{Proceedings of the 38th International
  Conference on Machine Learning}, \bibinfo{year}{2021}.
\bibitem[{Ramanujan et~al.(2020)Ramanujan, Wortsman, Kembhavi, Farhadi, and
  Rastegari}]{ramanujan_2019}
\bibinfo{author}{V.~Ramanujan}, \bibinfo{author}{M.~Wortsman},
  \bibinfo{author}{A.~Kembhavi}, \bibinfo{author}{A.~Farhadi},
  \bibinfo{author}{M.~Rastegari},
\newblock \bibinfo{title}{What's hidden in a randomly weighted neural
  network?},
\newblock in: \bibinfo{booktitle}{Proceedings of the IEEE/CVF Conference on
  Computer Vision and Pattern Recognition}, \bibinfo{year}{2020}.
\bibitem[{Aladago and Torresani(2021)}]{aladago_2021}
\bibinfo{author}{M.~M. Aladago}, \bibinfo{author}{L.~Torresani},
\newblock \bibinfo{title}{Slot machines: Discovering winning combinations of
  random weights in neural networks},
\newblock in: \bibinfo{booktitle}{Proceedings of the 38th International
  Conference on Machine Learning}, \bibinfo{year}{2021}.
\bibitem[{Mallya et~al.(2018)Mallya, Davis, and Lazebnik}]{mallya_2018}
\bibinfo{author}{A.~Mallya}, \bibinfo{author}{D.~Davis},
  \bibinfo{author}{S.~Lazebnik},
\newblock \bibinfo{title}{Piggyback: Adapting a single network to multiple
  tasks by learning to mask weights},
\newblock in: \bibinfo{booktitle}{Proceedings of the European Conference on
  Computer Vision}, \bibinfo{year}{2018}.
\bibitem[{Zhang et~al.(2021)Zhang, Lin, Chao, Wang, Wu, Huang, Xu, Tian, and
  Ji}]{zhang_2021}
\bibinfo{author}{Y.~Zhang}, \bibinfo{author}{M.~Lin},
  \bibinfo{author}{F.~Chao}, \bibinfo{author}{Y.~Wang},
  \bibinfo{author}{Y.~Wu}, \bibinfo{author}{F.~Huang}, \bibinfo{author}{M.~Xu},
  \bibinfo{author}{Y.~Tian}, \bibinfo{author}{R.~Ji},
\newblock \bibinfo{title}{Lottery jackpots exist in pre-trained models},
\newblock \bibinfo{journal}{CoRR} \bibinfo{volume}{abs/2104.08700v5}
  (\bibinfo{year}{2021}). \URLprefix \url{https://arxiv.org/abs/2104.08700v5},
  \bibinfo{note}{last accessed: 05/02/2022}.
\bibitem[{Bengio et~al.(2013)Bengio, L{\'{e}}onard, and
  Courville}]{bengio_2013}
\bibinfo{author}{Y.~Bengio}, \bibinfo{author}{N.~L{\'{e}}onard},
  \bibinfo{author}{A.~C. Courville},
\newblock \bibinfo{title}{Estimating or propagating gradients through
  stochastic neurons for conditional computation},
\newblock \bibinfo{journal}{CoRR} \bibinfo{volume}{abs/1308.3432}
  (\bibinfo{year}{2013}). \URLprefix \url{http://arxiv.org/abs/1308.3432},
  \bibinfo{note}{last accessed: 03/31/2022}.
\bibitem[{Liu and Zenke(2020)}]{liu_2020}
\bibinfo{author}{T.~Liu}, \bibinfo{author}{F.~Zenke},
\newblock \bibinfo{title}{Finding trainable sparse networks through neural
  tangent transfer},
\newblock in: \bibinfo{booktitle}{Proceedings of the 37th International
  Conference on Machine Learning}, \bibinfo{year}{2020}.
\bibitem[{Jacot et~al.(2018)Jacot, Hongler, and Gabriel}]{jacot_2018}
\bibinfo{author}{A.~Jacot}, \bibinfo{author}{C.~Hongler},
  \bibinfo{author}{F.~Gabriel},
\newblock \bibinfo{title}{Neural tangent kernel: Convergence and generalization
  in neural networks},
\newblock in: \bibinfo{booktitle}{Advances in Neural Information Processing
  Systems 31}, \bibinfo{year}{2018}.
\bibitem[{Arora et~al.(2019)Arora, Du, Hu, Li, Salakhutdinov, and
  Wang}]{arora_2019}
\bibinfo{author}{S.~Arora}, \bibinfo{author}{S.~S. Du},
  \bibinfo{author}{W.~Hu}, \bibinfo{author}{Z.~Li}, \bibinfo{author}{R.~R.
  Salakhutdinov}, \bibinfo{author}{R.~Wang},
\newblock \bibinfo{title}{On exact computation with an infinitely wide neural
  net},
\newblock in: \bibinfo{booktitle}{Advances in Neural Information Processing
  Systems 32}, \bibinfo{year}{2019}.
\bibitem[{Lee et~al.(2019)Lee, Xiao, Schoenholz, Bahri, Novak, Sohl-Dickstein,
  and Pennington}]{lee_2019b}
\bibinfo{author}{J.~Lee}, \bibinfo{author}{L.~Xiao},
  \bibinfo{author}{S.~Schoenholz}, \bibinfo{author}{Y.~Bahri},
  \bibinfo{author}{R.~Novak}, \bibinfo{author}{J.~Sohl-Dickstein},
  \bibinfo{author}{J.~Pennington},
\newblock \bibinfo{title}{Wide neural networks of any depth evolve as linear
  models under gradient descent},
\newblock in: \bibinfo{booktitle}{Advances in Neural Information Processing
  Systems 32}, \bibinfo{year}{2019}.
\bibitem[{Morcos et~al.(2019)Morcos, Yu, Paganini, and Tian}]{morcos_2019}
\bibinfo{author}{A.~Morcos}, \bibinfo{author}{H.~Yu},
  \bibinfo{author}{M.~Paganini}, \bibinfo{author}{Y.~Tian},
\newblock \bibinfo{title}{One ticket to win them all: generalizing lottery
  ticket initializations across datasets and optimizers},
\newblock in: \bibinfo{booktitle}{Advances in Neural Information Processing
  Systems 32}, \bibinfo{year}{2019}.
\bibitem[{Zhang et~al.(2021)Zhang, Chen, Chen, and Wang}]{zhang_2021b}
\bibinfo{author}{Z.~Zhang}, \bibinfo{author}{X.~Chen},
  \bibinfo{author}{T.~Chen}, \bibinfo{author}{Z.~Wang},
\newblock \bibinfo{title}{Efficient lottery ticket finding: Less data is more},
\newblock in: \bibinfo{booktitle}{Proceedings of the 38th International
  Conference on Machine Learning}, \bibinfo{year}{2021}.
\bibitem[{You et~al.(2020)You, Li, Xu, Fu, Wang, Chen, Baraniuk, Wang, and
  Lin}]{you_2019}
\bibinfo{author}{H.~You}, \bibinfo{author}{C.~Li}, \bibinfo{author}{P.~Xu},
  \bibinfo{author}{Y.~Fu}, \bibinfo{author}{Y.~Wang},
  \bibinfo{author}{X.~Chen}, \bibinfo{author}{R.~G. Baraniuk},
  \bibinfo{author}{Z.~Wang}, \bibinfo{author}{Y.~Lin},
\newblock \bibinfo{title}{Drawing early-bird tickets: Toward more efficient
  training of deep networks},
\newblock in: \bibinfo{booktitle}{8th International Conference on Learning
  Representations}, \bibinfo{year}{2020}.
\bibitem[{Gale et~al.(2019)Gale, Elsen, and Hooker}]{gale_2019}
\bibinfo{author}{T.~Gale}, \bibinfo{author}{E.~Elsen},
  \bibinfo{author}{S.~Hooker},
\newblock \bibinfo{title}{The state of sparsity in deep neural networks},
\newblock in: \bibinfo{booktitle}{36th International Conference on Machine
  Learning Joint Workshop on On-Device Machine Learning \& Compact Deep Neural
  Network Representations (ODML-CDNNR)}, \bibinfo{year}{2019}.
\bibitem[{He et~al.(2016)He, Zhang, Ren, and Sun}]{he_2016}
\bibinfo{author}{K.~He}, \bibinfo{author}{X.~Zhang}, \bibinfo{author}{S.~Ren},
  \bibinfo{author}{J.~Sun},
\newblock \bibinfo{title}{Deep residual learning for image recognition},
\newblock in: \bibinfo{booktitle}{Proceedings of the IEEE Conference on
  Computer Vision and Pattern Recognition}, \bibinfo{year}{2016}.
\bibitem[{Frankle et~al.(2020)Frankle, Schwab, and Morcos}]{frankle_2020b}
\bibinfo{author}{J.~Frankle}, \bibinfo{author}{D.~J. Schwab},
  \bibinfo{author}{A.~S. Morcos},
\newblock \bibinfo{title}{The early phase of neural network training},
\newblock in: \bibinfo{booktitle}{8th International Conference on Learning
  Representations}, \bibinfo{year}{2020}.
\bibitem[{Elesedy et~al.(2021)Elesedy, Kanade, and Teh}]{elesedy_2021}
\bibinfo{author}{B.~Elesedy}, \bibinfo{author}{V.~Kanade},
  \bibinfo{author}{Y.~W. Teh},
\newblock \bibinfo{title}{Lottery tickets in linear models: An analysis of
  iterative magnitude pruning},
\newblock in: \bibinfo{booktitle}{Sparsity in Neural Networks Workshop},
  \bibinfo{year}{2021}.
\bibitem[{Rosenfeld et~al.(2021)Rosenfeld, Frankle, Carbin, and
  Shavit}]{rosenfeld_2021}
\bibinfo{author}{J.~S. Rosenfeld}, \bibinfo{author}{J.~Frankle},
  \bibinfo{author}{M.~Carbin}, \bibinfo{author}{N.~Shavit},
\newblock \bibinfo{title}{On the predictability of pruning across scales},
\newblock in: \bibinfo{booktitle}{Proceedings of the 38th International
  Conference on Machine Learning}, \bibinfo{year}{2021}.
\bibitem[{Chen et~al.(2021)Chen, Frankle, Chang, Liu, Zhang, Carbin, and
  Wang}]{chen_2021b}
\bibinfo{author}{T.~Chen}, \bibinfo{author}{J.~Frankle},
  \bibinfo{author}{S.~Chang}, \bibinfo{author}{S.~Liu},
  \bibinfo{author}{Y.~Zhang}, \bibinfo{author}{M.~Carbin},
  \bibinfo{author}{Z.~Wang},
\newblock \bibinfo{title}{The lottery tickets hypothesis for supervised and
  self-supervised pre-training in computer vision models},
\newblock in: \bibinfo{booktitle}{Proceedings of the IEEE/CVF Conference on
  Computer Vision and Pattern Recognition}, \bibinfo{year}{2021}.
\bibitem[{Yu et~al.(2020)Yu, Edunov, Tian, and Morcos}]{yu_2020}
\bibinfo{author}{H.~Yu}, \bibinfo{author}{S.~Edunov},
  \bibinfo{author}{Y.~Tian}, \bibinfo{author}{A.~S. Morcos},
\newblock \bibinfo{title}{Playing the lottery with rewards and multiple
  languages: lottery tickets in rl and nlp},
\newblock in: \bibinfo{booktitle}{8th International Conference on Learning
  Representations}, \bibinfo{year}{2020}.
\bibitem[{Chen et~al.(2020)Chen, Frankle, Chang, Liu, Zhang, Wang, and
  Carbin}]{chen_2020b}
\bibinfo{author}{T.~Chen}, \bibinfo{author}{J.~Frankle},
  \bibinfo{author}{S.~Chang}, \bibinfo{author}{S.~Liu},
  \bibinfo{author}{Y.~Zhang}, \bibinfo{author}{Z.~Wang},
  \bibinfo{author}{M.~Carbin},
\newblock \bibinfo{title}{The lottery ticket hypothesis for pre-trained bert
  networks},
\newblock in: \bibinfo{booktitle}{Advances in Neural Information Processing
  Systems 33}, \bibinfo{year}{2020}.
\bibitem[{Vischer et~al.(2022)Vischer, Lange, and Sprekeler}]{vischer_2022}
\bibinfo{author}{M.~Vischer}, \bibinfo{author}{R.~T. Lange},
  \bibinfo{author}{H.~Sprekeler},
\newblock \bibinfo{title}{On lottery tickets and minimal task representations
  in deep reinforcement learning},
\newblock in: \bibinfo{booktitle}{10th International Conference on Learning
  Representations}, \bibinfo{year}{2022}.
\bibitem[{Soelen and Sheppard(2019)}]{soelen_2019}
\bibinfo{author}{R.~V. Soelen}, \bibinfo{author}{J.~W. Sheppard},
\newblock \bibinfo{title}{Using winning lottery tickets in transfer learning
  for convolutional neural networks},
\newblock in: \bibinfo{booktitle}{International Joint Conference on Neural
  Networks}, \bibinfo{year}{2019}.
\bibitem[{Girish et~al.(2021)Girish, Maiya, Gupta, Chen, Davis, and
  Shrivastava}]{girish_2021}
\bibinfo{author}{S.~Girish}, \bibinfo{author}{S.~R. Maiya},
  \bibinfo{author}{K.~Gupta}, \bibinfo{author}{H.~Chen}, \bibinfo{author}{L.~S.
  Davis}, \bibinfo{author}{A.~Shrivastava},
\newblock \bibinfo{title}{The lottery ticket hypothesis for object
  recognition},
\newblock in: \bibinfo{booktitle}{Proceedings of the IEEE/CVF Conference on
  Computer Vision and Pattern Recognition}, \bibinfo{year}{2021}.
\bibitem[{Chen et~al.(2021)Chen, Cheng, Wang, Gan, Liu, and Wang}]{chen_2021}
\bibinfo{author}{X.~Chen}, \bibinfo{author}{Y.~Cheng},
  \bibinfo{author}{S.~Wang}, \bibinfo{author}{Z.~Gan},
  \bibinfo{author}{J.~Liu}, \bibinfo{author}{Z.~Wang},
\newblock \bibinfo{title}{The elastic lottery ticket hypothesis},
\newblock in: \bibinfo{booktitle}{Advances in Neural Information Processing
  Systems 34}, \bibinfo{year}{2021}.
\bibitem[{Renda et~al.(2020)Renda, Frankle, and Carbin}]{renda_2020}
\bibinfo{author}{A.~Renda}, \bibinfo{author}{J.~Frankle},
  \bibinfo{author}{M.~Carbin},
\newblock \bibinfo{title}{Comparing rewinding and fine-tuning in neural network
  pruning},
\newblock in: \bibinfo{booktitle}{8th International Conference on Learning
  Representations}, \bibinfo{year}{2020}.
\bibitem[{Le and Hua(2021)}]{le_2021}
\bibinfo{author}{D.~H. Le}, \bibinfo{author}{B.-S. Hua},
\newblock \bibinfo{title}{Network pruning that matters: A case study on
  retraining variants},
\newblock in: \bibinfo{booktitle}{9th International Conference on Learning
  Representations}, \bibinfo{year}{2021}.
\bibitem[{Bai et~al.(2022)Bai, Wang, TAO, Li, and Fu}]{bai_2022}
\bibinfo{author}{Y.~Bai}, \bibinfo{author}{H.~Wang}, \bibinfo{author}{Z.~TAO},
  \bibinfo{author}{K.~Li}, \bibinfo{author}{Y.~Fu},
\newblock \bibinfo{title}{Dual lottery ticket hypothesis},
\newblock in: \bibinfo{booktitle}{10th International Conference on Learning
  Representations}, \bibinfo{year}{2022}.
\bibitem[{Su et~al.(2020)Su, Chen, Cai, Wu, Gao, Wang, and Lee}]{su_2020}
\bibinfo{author}{J.~Su}, \bibinfo{author}{Y.~Chen}, \bibinfo{author}{T.~Cai},
  \bibinfo{author}{T.~Wu}, \bibinfo{author}{R.~Gao}, \bibinfo{author}{L.~Wang},
  \bibinfo{author}{J.~D. Lee},
\newblock \bibinfo{title}{Sanity-checking pruning methods: Random tickets can
  win the jackpot},
\newblock in: \bibinfo{booktitle}{Advances in Neural Information Processing
  Systems 33}, \bibinfo{year}{2020}.
\bibitem[{Lubana and Dick(2021)}]{lubana_2021}
\bibinfo{author}{E.~S. Lubana}, \bibinfo{author}{R.~Dick},
\newblock \bibinfo{title}{A gradient flow framework for analyzing network
  pruning},
\newblock in: \bibinfo{booktitle}{9th International Conference on Learning
  Representations}, \bibinfo{year}{2021}.
\bibitem[{Zhang and Stadie(2020)}]{zhang_2020}
\bibinfo{author}{S.~Zhang}, \bibinfo{author}{B.~C. Stadie},
\newblock \bibinfo{title}{One-shot pruning of recurrent neural networks by
  jacobian spectrum evaluation},
\newblock in: \bibinfo{booktitle}{8th International Conference on Learning
  Representations}, \bibinfo{year}{2020}.
\bibitem[{Alizadeh et~al.(2022)Alizadeh, Tailor, Zintgraf, van Amersfoort,
  Farquhar, Lane, and Gal}]{alizadeh_2022}
\bibinfo{author}{M.~Alizadeh}, \bibinfo{author}{S.~A. Tailor},
  \bibinfo{author}{L.~M. Zintgraf}, \bibinfo{author}{J.~van Amersfoort},
  \bibinfo{author}{S.~Farquhar}, \bibinfo{author}{N.~D. Lane},
  \bibinfo{author}{Y.~Gal},
\newblock \bibinfo{title}{Prospect pruning: Finding trainable weights at
  initialization using meta-gradients},
\newblock in: \bibinfo{booktitle}{10th International Conference on Learning
  Representations}, \bibinfo{year}{2022}.
\bibitem[{Koster et~al.(2022)Koster, Grothe, and Rettinger}]{koster_2022}
\bibinfo{author}{N.~Koster}, \bibinfo{author}{O.~Grothe},
  \bibinfo{author}{A.~Rettinger},
\newblock \bibinfo{title}{Signing the supermask: Keep, hide, invert},
\newblock in: \bibinfo{booktitle}{10th International Conference on Learning
  Representations}, \bibinfo{year}{2022}.
\bibitem[{Chen et~al.(2022)Chen, Zhang, and Wang}]{chen_2022}
\bibinfo{author}{X.~Chen}, \bibinfo{author}{J.~Zhang},
  \bibinfo{author}{Z.~Wang},
\newblock \bibinfo{title}{Peek-a-boo: What (more) is disguised in a randomly
  weighted neural network, and how to find it efficiently},
\newblock in: \bibinfo{booktitle}{10th International Conference on Learning
  Representations}, \bibinfo{year}{2022}.
\bibitem[{Fischer and Burkholz(2022)}]{fischer_2022}
\bibinfo{author}{J.~Fischer}, \bibinfo{author}{R.~Burkholz},
\newblock \bibinfo{title}{Plant 'n' seek: Can you find the winning ticket?},
\newblock in: \bibinfo{booktitle}{10th International Conference on Learning
  Representations}, \bibinfo{year}{2022}.
\bibitem[{Liu et~al.(2022)Liu, Chen, Chen, Shen, Mocanu, Wang, and
  Pechenizkiy}]{liu_2022}
\bibinfo{author}{S.~Liu}, \bibinfo{author}{T.~Chen}, \bibinfo{author}{X.~Chen},
  \bibinfo{author}{L.~Shen}, \bibinfo{author}{D.~C. Mocanu},
  \bibinfo{author}{Z.~Wang}, \bibinfo{author}{M.~Pechenizkiy},
\newblock \bibinfo{title}{The unreasonable effectiveness of random pruning:
  Return of the most naive baseline for sparse training},
\newblock in: \bibinfo{booktitle}{10th International Conference on Learning
  Representations}, \bibinfo{year}{2022}.
\bibitem[{Malach et~al.(2020)Malach, Yehudai, Shalev-Schwartz, and
  Shamir}]{malach_2020}
\bibinfo{author}{E.~Malach}, \bibinfo{author}{G.~Yehudai},
  \bibinfo{author}{S.~Shalev-Schwartz}, \bibinfo{author}{O.~Shamir},
\newblock \bibinfo{title}{Proving the lottery ticket hypothesis: Pruning is all
  you need},
\newblock in: \bibinfo{booktitle}{Proceedings of the 37th International
  Conference on Machine Learning}, \bibinfo{year}{2020}.
\bibitem[{Pensia et~al.(2020)Pensia, Rajput, Nagle, Vishwakarma, and
  Papailiopoulos}]{pensia_2020}
\bibinfo{author}{A.~Pensia}, \bibinfo{author}{S.~Rajput},
  \bibinfo{author}{A.~Nagle}, \bibinfo{author}{H.~Vishwakarma},
  \bibinfo{author}{D.~Papailiopoulos},
\newblock \bibinfo{title}{Optimal lottery tickets via subset sum: Logarithmic
  over-parameterization is sufficient},
\newblock in: \bibinfo{booktitle}{Advances in Neural Information Processing
  Systems 33}, \bibinfo{year}{2020}.
\bibitem[{Orseau et~al.(2020)Orseau, Hutter, and Rivasplata}]{orseau_2020}
\bibinfo{author}{L.~Orseau}, \bibinfo{author}{M.~Hutter},
  \bibinfo{author}{O.~Rivasplata},
\newblock \bibinfo{title}{Logarithmic pruning is all you need},
\newblock in: \bibinfo{booktitle}{Advances in Neural Information Processing
  Systems 33}, \bibinfo{year}{2020}.
\bibitem[{Chijiwa et~al.(2021)Chijiwa, Yamaguchi, Ida, Umakoshi, and
  Inoue}]{chijiwa_2021}
\bibinfo{author}{D.~Chijiwa}, \bibinfo{author}{S.~Yamaguchi},
  \bibinfo{author}{Y.~Ida}, \bibinfo{author}{K.~Umakoshi},
  \bibinfo{author}{T.~Inoue},
\newblock \bibinfo{title}{Pruning randomly initialized neural networks with
  iterative randomization},
\newblock in: \bibinfo{booktitle}{Advances in Neural Information Processing
  Systems 34}, \bibinfo{year}{2021}.
\bibitem[{Burkholz et~al.(2022)Burkholz, Laha, Mukherjee, and
  Gotovos}]{burkholz_2022}
\bibinfo{author}{R.~Burkholz}, \bibinfo{author}{N.~Laha},
  \bibinfo{author}{R.~Mukherjee}, \bibinfo{author}{A.~Gotovos},
\newblock \bibinfo{title}{On the existence of universal lottery tickets},
\newblock in: \bibinfo{booktitle}{10th International Conference on Learning
  Representations}, \bibinfo{year}{2022}.
\bibitem[{da~Cunha et~al.(2022)da~Cunha, Natale, and Viennot}]{cunha_2022}
\bibinfo{author}{A.~da~Cunha}, \bibinfo{author}{E.~Natale},
  \bibinfo{author}{L.~Viennot},
\newblock \bibinfo{title}{Proving the lottery ticket hypothesis for
  convolutional neural networks},
\newblock in: \bibinfo{booktitle}{10th International Conference on Learning
  Representations}, \bibinfo{year}{2022}.
\bibitem[{Zagoruyko and Komodakis(2016)}]{zagoruyko_2016}
\bibinfo{author}{S.~Zagoruyko}, \bibinfo{author}{N.~Komodakis},
\newblock \bibinfo{title}{Wide residual networks},
\newblock in: \bibinfo{booktitle}{Proceedings of the British Machine Vision
  Conference}, \bibinfo{year}{2016}.
\bibitem[{{Deng} et~al.(2009){Deng}, {Dong}, {Socher}, {Li}, {Li}, and
  {Fei-Fei}}]{deng_2009}
\bibinfo{author}{J.~{Deng}}, \bibinfo{author}{W.~{Dong}},
  \bibinfo{author}{R.~{Socher}}, \bibinfo{author}{L.~{Li}},
  \bibinfo{author}{K.~{Li}}, \bibinfo{author}{L.~{Fei-Fei}},
\newblock \bibinfo{title}{Imagenet: A large-scale hierarchical image database},
\newblock in: \bibinfo{booktitle}{Proceedings of the IEEE Conference on
  Computer Vision and Pattern Recognition}, \bibinfo{year}{2009}.
\bibitem[{Chambers and Rumpel(2017)}]{chambers_2017}
\bibinfo{author}{A.~R. Chambers}, \bibinfo{author}{S.~Rumpel},
\newblock \bibinfo{title}{A stable brain from unstable components: Emerging
  concepts and implications for neural computation},
\newblock \bibinfo{journal}{Neuroscience} \bibinfo{volume}{357}
  (\bibinfo{year}{2017}) \bibinfo{pages}{172--184}.
\bibitem[{Du et~al.(2019)Du, Zhai, Poczos, and Singh}]{du_2019}
\bibinfo{author}{S.~S. Du}, \bibinfo{author}{X.~Zhai},
  \bibinfo{author}{B.~Poczos}, \bibinfo{author}{A.~Singh},
\newblock \bibinfo{title}{Gradient descent provably optimizes
  over-parameterized neural networks},
\newblock in: \bibinfo{booktitle}{7th International Conference on Learning
  Representations}, \bibinfo{year}{2019}.
\bibitem[{Li and Liang(2018)}]{li_2018b}
\bibinfo{author}{Y.~Li}, \bibinfo{author}{Y.~Liang},
\newblock \bibinfo{title}{Learning overparameterized neural networks via
  stochastic gradient descent on structured data},
\newblock in: \bibinfo{booktitle}{Advances in Neural Information Processing
  Systems 31}, \bibinfo{year}{2018}.
\bibitem[{Brutzkus et~al.(2018)Brutzkus, Globerson, Malach, and
  Shalev-Shwartz}]{brutzkus_2018}
\bibinfo{author}{A.~Brutzkus}, \bibinfo{author}{A.~Globerson},
  \bibinfo{author}{E.~Malach}, \bibinfo{author}{S.~Shalev-Shwartz},
\newblock \bibinfo{title}{{SGD} learns over-parameterized networks that
  provably generalize on linearly separable data},
\newblock in: \bibinfo{booktitle}{6th International Conference on Learning
  Representations}, \bibinfo{year}{2018}.
\bibitem[{Liu et~al.(2021)Liu, Yin, Mocanu, and Pechenizkiy}]{liu_2021b}
\bibinfo{author}{S.~Liu}, \bibinfo{author}{L.~Yin}, \bibinfo{author}{D.~C.
  Mocanu}, \bibinfo{author}{M.~Pechenizkiy},
\newblock \bibinfo{title}{Do we actually need dense over-parameterization?
  {In}-time over-parameterization in sparse training},
\newblock in: \bibinfo{booktitle}{Proceedings of the 38th International
  Conference on Machine Learning}, \bibinfo{year}{2021}.
\bibitem[{Ding et~al.(2019)Ding, Ding, Zhou, Guo, Han, and Liu}]{ding_2019}
\bibinfo{author}{X.~Ding}, \bibinfo{author}{G.~Ding},
  \bibinfo{author}{X.~Zhou}, \bibinfo{author}{Y.~Guo},
  \bibinfo{author}{J.~Han}, \bibinfo{author}{J.~Liu},
\newblock \bibinfo{title}{Global sparse momentum sgd for pruning very deep
  neural networks},
\newblock in: \bibinfo{booktitle}{Advances in Neural Information Processing
  Systems 32}, \bibinfo{year}{2019}.
\bibitem[{Kusupati et~al.(2020)Kusupati, Ramanujan, Somani, Wortsman, Jain,
  Kakade, and Farhadi}]{kusupati_2020}
\bibinfo{author}{A.~Kusupati}, \bibinfo{author}{V.~Ramanujan},
  \bibinfo{author}{R.~Somani}, \bibinfo{author}{M.~Wortsman},
  \bibinfo{author}{P.~Jain}, \bibinfo{author}{S.~Kakade},
  \bibinfo{author}{A.~Farhadi},
\newblock \bibinfo{title}{Soft threshold weight reparameterization for
  learnable sparsity},
\newblock in: \bibinfo{booktitle}{Proceedings of the 37th International
  Conference on Machine Learning}, \bibinfo{year}{2020}.
\bibitem[{Liu et~al.(2020)Liu, Xu, Shi, Cheung, and So}]{liu_2020b}
\bibinfo{author}{J.~Liu}, \bibinfo{author}{Z.~Xu}, \bibinfo{author}{R.~Shi},
  \bibinfo{author}{R.~C.~C. Cheung}, \bibinfo{author}{H.~K. So},
\newblock \bibinfo{title}{Dynamic sparse training: Find efficient sparse
  network from scratch with trainable masked layers},
\newblock in: \bibinfo{booktitle}{8th International Conference on Learning
  Representations}, \bibinfo{year}{2020}.
\bibitem[{Schwarz et~al.(2021)Schwarz, Jayakumar, Pascanu, Latham, and
  Teh}]{schwarz_2021}
\bibinfo{author}{J.~Schwarz}, \bibinfo{author}{S.~Jayakumar},
  \bibinfo{author}{R.~Pascanu}, \bibinfo{author}{P.~E. Latham},
  \bibinfo{author}{Y.~W. Teh},
\newblock \bibinfo{title}{Powerpropagation: A sparsity inducing weight
  reparameterisation},
\newblock in: \bibinfo{booktitle}{Advances in Neural Information Processing
  Systems 34}, \bibinfo{year}{2021}.
\bibitem[{Zhou et~al.(2021{\natexlab{a}})Zhou, Zhang, Xu, and
  Zhang}]{zhou_2021b}
\bibinfo{author}{X.~Zhou}, \bibinfo{author}{W.~Zhang}, \bibinfo{author}{H.~Xu},
  \bibinfo{author}{T.~Zhang},
\newblock \bibinfo{title}{Effective sparsification of neural networks with
  global sparsity constraint},
\newblock in: \bibinfo{booktitle}{Proceedings of the IEEE/CVF Conference on
  Computer Vision and Pattern Recognition}, \bibinfo{year}{2021}{\natexlab{a}}.
\bibitem[{Zhou et~al.(2021{\natexlab{b}})Zhou, Zhang, Chen, Diao, and
  Zhang}]{zhou_2021c}
\bibinfo{author}{X.~Zhou}, \bibinfo{author}{W.~Zhang},
  \bibinfo{author}{Z.~Chen}, \bibinfo{author}{S.~Diao},
  \bibinfo{author}{T.~Zhang},
\newblock \bibinfo{title}{Efficient neural network training via forward and
  backward propagation sparsification},
\newblock in: \bibinfo{booktitle}{Advances in Neural Information Processing
  Systems 34}, \bibinfo{year}{2021}{\natexlab{b}}.
\bibitem[{Pao and Takefuji(1992)}]{pao_1992}
\bibinfo{author}{Y.-H. Pao}, \bibinfo{author}{Y.~Takefuji},
\newblock \bibinfo{title}{Functional-link net computing: theory, system
  architecture, and functionalities},
\newblock \bibinfo{journal}{Computer} \bibinfo{volume}{25}
  (\bibinfo{year}{1992}) \bibinfo{pages}{76--79}.
\bibitem[{Pao et~al.(1994)Pao, Park, and Sobajic}]{pao_1994}
\bibinfo{author}{Y.-H. Pao}, \bibinfo{author}{G.-H. Park},
  \bibinfo{author}{D.~J. Sobajic},
\newblock \bibinfo{title}{Learning and generalization characteristics of the
  random vector functional-link net},
\newblock \bibinfo{journal}{Neurocomputing} \bibinfo{volume}{6}
  (\bibinfo{year}{1994}) \bibinfo{pages}{163--180}.
\bibitem[{Krizhevsky(2012)}]{krizhevsky_2012}
\bibinfo{author}{A.~Krizhevsky},
\newblock \bibinfo{title}{Learning multiple layers of features from tiny
  images},
\newblock \bibinfo{journal}{University of Toronto}  (\bibinfo{year}{2012}).
  \bibinfo{note}{{URL}: \url{http://www.cs.toronto.edu/~kriz/cifar.html}, last
  accessed: 05/13/2022}.
\bibitem[{Frankle et~al.(2021)Frankle, Schwab, and Morcos}]{frankle_2020}
\bibinfo{author}{J.~Frankle}, \bibinfo{author}{D.~J. Schwab},
  \bibinfo{author}{A.~S. Morcos},
\newblock \bibinfo{title}{Training batchnorm and only batchnorm: On the
  expressive power of random features in {CNN}s},
\newblock in: \bibinfo{booktitle}{9th International Conference on Learning
  Representations}, \bibinfo{year}{2021}.
\bibitem[{Evci et~al.(2022)Evci, Dauphin, Ioannou, and Keskin}]{evci_2020b}
\bibinfo{author}{U.~Evci}, \bibinfo{author}{Y.~Dauphin},
  \bibinfo{author}{Y.~Ioannou}, \bibinfo{author}{C.~Keskin},
\newblock \bibinfo{title}{Gradient flow in sparse neural networks and how
  lottery tickets win},
\newblock in: \bibinfo{booktitle}{Proceedings of the AAAI Conference on
  Artificial Intelligence}, \bibinfo{year}{2022}.
\bibitem[{{LeCun} et~al.(1998){LeCun}, {Bottou}, {Bengio}, and
  {Haffner}}]{lecun_1998}
\bibinfo{author}{Y.~{LeCun}}, \bibinfo{author}{L.~{Bottou}},
  \bibinfo{author}{Y.~{Bengio}}, \bibinfo{author}{P.~{Haffner}},
\newblock \bibinfo{title}{Gradient-based learning applied to document
  recognition},
\newblock \bibinfo{journal}{Proceedings of the IEEE} \bibinfo{volume}{86}
  (\bibinfo{year}{1998}) \bibinfo{pages}{2278--2324}.
\bibitem[{Erd\H{o}s and R\'enyi(1959)}]{erdoes_1959}
\bibinfo{author}{P.~Erd\H{o}s}, \bibinfo{author}{A.~R\'enyi},
\newblock \bibinfo{title}{On random graphs {I}},
\newblock \bibinfo{journal}{Publicationes Mathematicae Debrecen}
  \bibinfo{volume}{6} (\bibinfo{year}{1959}) \bibinfo{pages}{290--297}.

\end{thebibliography}
	
\end{document}